\documentclass[final,3p,times,authoryear]{elsarticle}
\usepackage{amssymb}
\usepackage{amsmath}
\usepackage{cite}
\usepackage{amsmath,amssymb,amsfonts}
\usepackage{algorithmic}
\usepackage{graphicx}
\usepackage{textcomp}
\usepackage{bm}
\usepackage{multirow}
\usepackage{threeparttable}
\usepackage{graphicx}
\usepackage{subcaption}
\usepackage{url}
\usepackage{booktabs} % For professional table formatting
\usepackage{siunitx} % For consistent number formatting in tables
\usepackage{xcolor}
\usepackage{booktabs} % For professional-quality tables
\usepackage{multirow} % For multirow cells
\usepackage{tabularx} % For adjustable column widths
\usepackage{siunitx}
\usepackage{hyperref}
\usepackage{xcolor} % Needed for \color
\usepackage[linesnumbered,ruled,vlined]{algorithm2e}
\usepackage{mathtools}

\DeclareSIUnit{\inch}{''}

\usepackage{booktabs}
\journal{Nuclear Physics B}

\begin{document}

\begin{frontmatter}

%% Title, authors and addresses

%% use the tnoteref command within \title for footnotes;
%% use the tnotetext command for theassociated footnote;
%% use the fnref command within \author or \affiliation for footnotes;
%% use the fntext command for theassociated footnote;
%% use the corref command within \author for corresponding author footnotes;
%% use the cortext command for theassociated footnote;
%% use the ead command for the email address,
%% and the form \ead[url] for the home page:
%% \title{Title\tnoteref{label1}}
%% \tnotetext[label1]{}
%% \author{Name\corref{cor1}\fnref{label2}}
%% \ead{email address}
%% \ead[url]{home page}
%% \fntext[label2]{}
%% \cortext[cor1]{}
%% \affiliation{organization={},
%%            addressline={}, 
%%            city={},
%%            postcode={}, 
%%            state={},
%%            country={}}
%% \fntext[label3]{}

\title{Graph-Based Fault Diagnosis for Rotating Machinery: Adaptive Segmentation and Structural Feature Integration} %% Article title

%%% use optional labels to link authors explicitly to addresses:
 \author[label1]{Moirangthem Tiken Singh}
 \affiliation[label1]{organization={Department of Computer Science and Engineering, DUIET, Dibrugarh University},
             addressline={Dibrugarh},
             postcode={786004},
             state={Assam},
             country={India}}
             \ead{tiken.m@dibru.ac.in}

% \affiliation[label2]{organization={},
%             addressline={},
%             city={},
%             postcode={},
%             state={},
%             country={}}

%\author{} %% Author name
%
%%% Author affiliation
%\affiliation{organization={},%Department and Organization
%            addressline={}, 
%            city={},
%            postcode={}, 
%            state={},
%            country={}}

\begin{abstract}
Rotating machinery, such as bearings and gearboxes, plays a critical role in industrial operations; however, faults in these components can lead to costly downtime, safety hazards, and inefficient maintenance. Traditional fault diagnosis methods often rely on manual expertise or complex deep learning models that require vast labeled data, high computational resources, and lack transparency, limiting their practicality in real-world settings. To address these challenges, this study introduces a novel graph-based framework for multiclass fault diagnosis that emphasizes interpretability, efficiency, and robustness. Vibration signals were adaptively segmented using entropy optimization to focus on informative fault-related bursts, and then transformed into time-frequency representations. Each segment was modeled as a graph, extracting structural features such as average shortest path length, modularity, and spectral gap, which were combined with statistical signal descriptors to train simple, interpretable classifiers: logistic regression, support vector machines, and random forests.  Evaluated on the Case Western Reserve University (CWRU) bearing dataset across 0--3 HP loads and the Southeast University (SU) gearbox dataset under varied speed-load conditions, the framework outperformed traditional and deep learning approaches. The random forest classifier achieved up to 99.9\% accuracy on the CWRU and 100\% on the SU datasets. Even with added noise (standard deviation = 0.5), it maintains 89.3\% and 98.1\% accuracy on the CWRU and SU datasets, respectively. In cross-domain scenarios, it delivered F1-scores of 97.73\% for cross-load and 97.99\% for cross-fault transfers, showcasing strong generalizability without deep architectures. By combining graph-theoretic analysis with statistical signal processing, this method eliminates the need for resource-intensive models while enhancing adaptability to noise, varying loads, and diverse fault types. Its transparency allows engineers to trace anomalies to specific causes, making it ideal for real-time industrial fault diagnosis and predictive maintenance applications.

\end{abstract}

% Keywords

\end{frontmatter}

%% Add \usepackage{lineno} before \begin{document} and uncomment 
%% following line to enable line numbers
%% \linenumbers

%% main text
\section{Introduction}
\label{sec:introduction}
Rotating machinery, including motors, gearboxes, and bearings, is integral to industries such as manufacturing, energy, transportation, and aerospace. The reliability of these systems hinges on early fault detection in critical components, which are prone to failure under mechanical stress \citep{lei2013review}. Undetected faults can lead to costly downtime, maintenance expenses, or safety hazards, necessitating accurate and efficient fault diagnosis systems \citep{jardine2006review}.Vibration signals, which capture fault-induced patterns, are widely used for diagnosis, but their analysis poses challenges due to dynamic operating conditions, noise, and complex fault patterns \citep{lei2020applications}.

Traditional fault diagnosis methods for rotating machinery rely on handcrafted features extracted from vibration signals in the time or frequency domain, followed by classifiers like support vector machines (SVMs)~\citep{widodo2007support}. Time-domain features (e.g., mean, RMS, skewness~\citep{joanes1998}) and frequency-domain features (e.g., FFT~\citep{welch1967}, spectral envelope~\citep{zheng2022spectral}) capture fault signatures, while techniques like empirical mode decomposition (EMD)~\citep{lei2013review} and Teager-Kaiser energy~\citep{kaiser1990} enhance signal decomposition. These methods achieve high accuracy (e.g., 95--98\% on the CWRU dataset~\citep{ZHANG2015164}) in controlled settings but struggle with dynamic conditions and noise due to fixed segmentation and manual feature selection, requiring significant domain expertise~\citep{lei2020applications, rauber2014heterogeneous}.

To address the limitations of manual feature engineering, machine learning methods automate feature selection and classification, marking a significant evolution in fault diagnosis \citep{R2025103892}. Ensemble methods~\citep{ZHANG2015164} and transfer learning~\citep{zhang2017transfer} improve robustness across operating conditions. For example, Zhang et al.~\citep{zhang2017transfer} used neural networks for bearing fault diagnosis under varying loads, achieving 90-95\% accuracy on CWRU cross-load scenarios. However, reliance on handcrafted features limits their ability to capture complex patterns in noisy or high-dimensional data. Classifiers like k-nearest neighbors (kNN)~\citep{cover1967} and SVMs~\citep{widodo2007support, Yazdani-Asrami_2023} are computationally efficient but lack the generalization of deep learning for large datasets~\citep{lei2020applications}.

Building on the automation introduced by machine learning, deep learning revolutionizes fault diagnosis by learning hierarchical features directly from raw vibration signals~\citep{ZHAO2019213}. Convolutional neural networks (CNNs)~\citep{wen2017new, niyongabo2022bearing, Xu28022025}, long short-term memory (LSTM) networks~\citep{zhang2020attention}, and attention-based models~\citep{li2024tdanet} achieve high accuracies (98-99.9\%) on benchmarks like CWRU and SU datasets~\citep{cwru_bearing_dataset, cathy_mechanical_datasets}. For instance, Shao et al.~\citep{shao2018highly} reported 99.5\% accuracy using deep transfer learning on the SU dataset. Recent advancements include multi-source fusion~\citep{makrouf2023multi}, sparse CNNs~\citep{Xu28022025}, and transformer-based models~\citep{zhou2024drswin}. However, deep learning models require large labeled datasets, incur high computational costs (e.g., 536 s inference for DL-CNN~\citep{shao2018highly}), and lack interpretability, with performance dropping in cross-domain scenarios (e.g., 90-95\% F1-score on cross-load CWRU~\citep{li2022multi, zhu2023review}).

Several advanced methods have emerged to address the challenges of limited generalization and high computational costs in conventional deep learning approaches for fault diagnosis. For example, the MDIFN~\citep{GAO2024102278} uses complex-valued CNNs and BiGRUs in an adversarial transfer-learning setup. This extracts domain-invariant spatiotemporal features from multi-source data, improving cross-domain performance and decreasing the need for extensive labeled datasets. However, its computational complexity and limited interpretability persist as drawbacks. Also, the Domain Feature Decoupling Network (DFDN)~\citep{GAO2024110449} separates domain and fault features before adaptively merging them, increasing interpretability and data efficiency for cross-fault transfer learning.
	
While many traditional deep learning classifiers demand extensive labeled data, recent innovations mitigate this through lightweight designs and few-shot strategies. For example, \citet{YAN2024121338} introduces LiConvFormer, a CNN/Transformer hybrid using separable multiscale convolution and broadcast self-attention for compact, robust fault diagnosis. Similarly, \citet{10759278} proposes a few-shot class-incremental learning framework that enables wind turbine system-level fault diagnosis by incorporating new classes with minimal labeled examples, leveraging forward-backward compatible representations to avoid catastrophic forgetting. Additionally, \citet{CHEN2024746} develops a multi-modal self-supervised learning method that pre-trains on unlabeled vibration and acoustic signals for cross-domain one-shot bearing fault diagnosis, achieving high accuracy with just one target-domain label. These approaches collectively show that deep networks can deliver strong performance in data-scarce scenarios via efficient architectures and learning paradigms. Nevertheless, these techniques are compatible with the proposed framework, yielding improved accuracy and noise robustness, independent of deep learning, focusing on explainability and efficiency with limited data or domain changes.

Graph-based methods offer a novel approach by modeling structural relationships in vibration signals, addressing the adaptability issues of traditional and machine learning methods~\citep{yang2021supergraph, singh2025ensemble}. Yang et al.~\citep{yang2021supergraph} used spatial-temporal graphs to achieve 97\% accuracy on CWRU, leveraging metrics like modularity (a measure of how well a graph divides into cohesive sub-groups~\citep{newman2006modularity}) and spectral gap (the difference between the largest eigenvalues of the graph’s adjacency matrix, indicating connectivity robustness~\citep{cheeger1970lower, chung1996lectures}). Singh~\citep{singh2025ensemble} integrated graph autoencoders with transformers for robust fault detection, using small-world properties (efficient connectivity with short average path lengths~\citep{watts1998collective}) to capture fault-induced disruptions. These methods enhance adaptability to dynamic conditions by modeling inter-segment relationships, unlike the fixed segmentation of traditional approaches. However, they are computationally intensive and prioritize structural relationships over frequency-domain features~\citep{yang2021supergraph, lei2020applications}. The proposed framework addresses this by integrating local signal characteristics with global graph metrics, improving both accuracy and interoperability.

The CWRU bearing dataset~\citep{cwru_bearing_dataset} and SU gearbox dataset~\citep{cathy_mechanical_datasets} are standard benchmarks. CWRU includes vibration data under varying loads (0--3 HP) and fault types, while SU covers gearbox faults under diverse conditions~\citep{shao2018highly}. Performance metrics like accuracy, F1-score, and inference time are evaluated, often using statistical tests (e.g., ANOVA~\citep{fisher1925statistical}, Tukey HSD~\citep{tukey1949comparing}) for cross-load robustness~\citep{zhang2017transfer}. Recent studies report 98-99.9\% accuracy on these datasets, but cross-load performance remains lower (90-95\% F1-score~\citep{li2022multi}).

Despite advancements, challenges persist across fault diagnosis methods. Traditional and machine learning approaches lack adaptability to dynamic conditions due to manual feature engineering~\citep{lei2020applications, rauber2014heterogeneous}. Deep learning models, while powerful, require large datasets, incur high computational costs, and lack interpretability, with post-hoc methods like Grad-CAM offering limited insight~\citep{zhang2020deep}. Cross-domain generalization remains a hurdle, with performance drops due to domain shifts~\citep{zhang2017transfer, li2022multi}. Graph-based methods, though effective for structural modeling, are computationally intensive and underutilize frequency-domain features~\citep{yang2021supergraph, singh2025ensemble}. The underuse of inter-segment relationships in vibration signals across all methods misses opportunities to capture fault-induced patterns, underscoring the need for adaptive, structurally informed, and generalizable approaches~\citep{lei2020applications, yang2021supergraph}.

%A critical gap across these methods is the underuse of structural information within vibration signals, which exhibit temporal and physical dependencies that are often overlooked by processing segments independently \citep{yang2021supergraph}. Graph-based methods offer a promising solution by modeling inter-segment relationships, achieving high accuracy (e.g., 97\% on CWRU \citep{yang2021supergraph}) through metrics like modularity \citep{newman2006modularity} and spectral gap \citep{cheeger1970lower}. However, existing graph-based approaches, such as those using Graph Neural Networks (GNNs) \citep{xiao2023graph} or spatial-temporal architectures \citep{singh2024spatial}, are computationally intensive and prioritize structural relationships over frequency-domain features \citep{lei2020applications}.

To address these challenges, this paper proposes a novel graph-based framework that emphasizes interpretability and computational efficiency through feature engineering for multiclass fault diagnosis in rotating machinery. Unlike end-to-end deep learning, our approach integrates entropy-driven segmentation to optimize window and step sizes, ensuring adaptive and robust feature extraction under varying conditions.

Each segment is characterized by a comprehensive feature set, including time-domain statistics, frequency-domain descriptors (e.g., spectral density, envelope analysis), and Teager–Kaiser energy~\citep{welch1967}. These features are clustered to form representative patterns, from which $k$-nearest neighbor (kNN) graphs are constructed. Graph-theoretic metrics, such as average shortest path length, modularity, and spectral gap, capture fault-induced structural changes~\citep{newman2018networks}, providing a lightweight yet powerful basis for fault classification. By combining global graph metrics with local signal features, the framework achieves a dual-level representation that enhances the detection of subtle faults and improves noise resilience.

To leverage this enriched representation for classification, the proposed framework integrates a multiclass strategy using logistic regression (LR), random forests (RF), and support vector machines (SVM). These classifiers are chosen to balance interpretability, generalization, and robustness: LR provides a transparent baseline; SVM captures complex decision boundaries in high-dimensional space; and RF mitigates feature redundancy through ensemble learning. Integrating Shapley Additive exPlanations (SHAP)~\citep{lundberg2017unified} further improves interpretability by quantifying feature contributions, reducing model-specific bias, and promoting trustworthy fault explanations. The complete pipeline is validated on the benchmark CWRU and SU datasets across diverse load and fault conditions, demonstrating superior diagnostic accuracy, robustness to noise, and cross-domain adaptability compared to conventional, machine learning, and deep learning approaches.

This study aims to:
\begin{itemize}
	\item Enhance vibration signal interpretation through entropy-guided segmentation.
	\item Leverage graph learning to capture structural patterns in machinery health data.
	\item Combine graph-based and statistical features for robust fault characterization.
	\item Evaluate scalable, interpretable classifiers on standard benchmark datasets.
	\item Demonstrate robustness to noise and transferability across domains.
\end{itemize}
This approach bridges the gap between structural modeling and feature engineering, offering a transparent, computationally efficient, and generalizable solution for real-time industrial monitoring and predictive maintenance.

\section{Methodology}

\begin{figure*}[h!]
	\centering
	\includegraphics[width=1\textwidth]{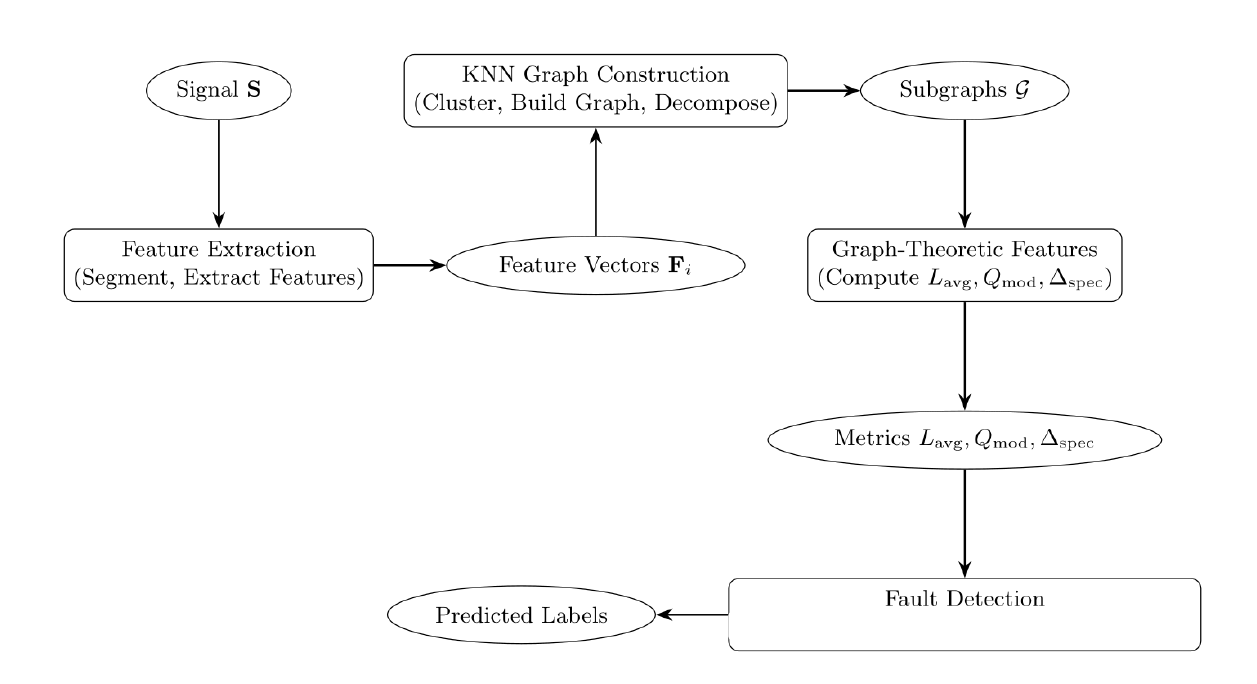}
	\caption{Overview of the methodology for multiclass fault detection. Source: Authors own work}
	\label{fig:model}
\end{figure*}

We propose a methodology that begins by extracting time- and frequency-domain features \( \mathbf{F}_i \) from a multichannel vibrational signal \( \mathbf{S} \). The signal is segmented, and clustered segments are used to construct a K-Nearest Neighbors (KNN) graph \( \mathcal{G} \). From this graph, we compute key graph-theoretic features, including average shortest path length  \( L_{\text{avg}} \), modularity $Q_{\text{mod}}$, and spectral gap $\Delta_{\text{spec}}$. These features serve as input to a multiclass classifier model designed for fault detection. An overview of the proposed pipeline is illustrated in Figure~\ref{fig:model}, outlining the progression from raw signal input to fault classification via feature extraction, graph construction, graph-based feature computation, and model training. In the following subsection, we detail each component of the proposed approach. 

Table~\ref{tab:symbols} summarizes the key symbols used in this work, focusing on essential variables for brevity.

\begin{table*}[h!]
	\centering
	\caption{Summary of Key Symbols. Source: Authors own work.}
	\label{tab:symbols}
	\begin{tabular}{|l|l|}
		\toprule
		\textbf{Symbol} & \textbf{Description} \\
		\midrule
		\( \mathbf{S} \) & Original multi $ch$-channel signal (\( \mathbb{R}^{N \times ch} \)) \\
		\( \omega^* \) & Optimal window size for segmentation \\
		\( \sigma^* \) & Optimal step size for segmentation \\
		\( n \) & Number of segments (\( \lfloor (N - \omega^*) / \sigma^* \rfloor \)) \\
		\( \mathbf{F}_i \) & Feature vector for the \( i \)-th segment (\( \mathbb{R}^d \)) \\
		\( d \) & Dimension of feature vectors \\
		\( K_c \) & Number of clusters in MiniBatch K-Means \\
		\( k_{\text{NN}} \) & Number of nearest neighbors in KNN graph \\
		\( \tau \) & Distance threshold for KNN graph edges \\
		\( \mathcal{G} \) & Set of subgraphs \( \{\mathcal{G}_1, \mathcal{G}_2, \dots, \mathcal{G}_{m_G}\} \) \\
		\( m_G \) & Number of subgraphs \\
		\( N_{\text{max}} \) & Maximum number of nodes per subgraph \\
		\( w_{ij} \) & Edge weight between nodes \( i \) and \( j \) \\
		\( N_{\text{total}} \) & Total number of nodes across all subgraphs (\( \approx n \)) \\
		\( L_{\text{avg}} \) & System-wide average shortest path length \\
		\( Q_{\text{mod}} \) & System-wide average modularity \\
		\( \Delta_{\text{spec}} \) & System-wide average spectral gap \\
		\( \ell_v \) & Multiclass fault label of node \( v \) (\( \{0, 1, \ldots, K_F-1\} \)) \\
		\( K_F \) & Number of fault classes \\
		\( \mathbf{z}_v \) & Combined feature vector for node \( v \) (\( \mathbb{R}^{d+3} \)) \\
		\( \mathbf{Z} \) & Feature matrix (\( \mathbb{R}^{N_{\text{total}} \times (d+3)} \)) \\
		\( \hat{\ell}_v \) & Predicted class label for node \( v \) \\
		\bottomrule
	\end{tabular}
\end{table*}

\subsection{Optimizing Segmentation Process}

In the vibrational signal fault detection pipeline, we propose a methodology to optimize window parameters for signal segmentation. This approach combines multiple entropy measures with a systematic parameter search to determine the optimal window size (\(\omega^*\)) and step size (\(\sigma^*\)). By leveraging diverse entropy metrics, the method captures complementary signal characteristics, enabling effective segmentation that enhances fault detection accuracy.

We include the step size (\(\sigma\)) to control the degree of overlap between consecutive segments during signal segmentation. While the window size (\(\omega\)) determines the duration of each segment, the step size governs how far the window advances across the signal. Choosing an appropriate step size is essential for balancing temporal resolution and redundancy—smaller values of \(\sigma\) yield higher overlap and finer temporal detail but increase computational cost and potential data redundancy. Conversely, larger step sizes reduce overlap, speeding up computation but potentially missing transient features. Therefore, optimizing both \(\omega\) and \(\sigma\) ensures effective signal representation for entropy-based feature extraction.

We begin by calculating Shannon entropy as the foundational metric for a discrete probability distribution \( P = \{p_1, p_2, \dots, p_n\} \), expressed as:
\begin{equation}
	H(P) = -\sum_{i=1}^n p_i \log(p_i)
\end{equation}

where \( p_i \) represents the probability of the \( i \)-th outcome \citep{shannon1948}.

For a continuous-time signal \( x(t) \), we derive the probability distribution \( P \) from amplitude histograms of windowed segments \( x_w(t) \) of duration \( \omega \). The discrete representation of each segment is
\[
\mathbf{x}_w = [x(t_0), x(t_0 + \Delta t), \dots, x(t_0 + \omega)] \in \mathbb{R}^\omega,
\]
where \( \Delta t \) denotes the sampling interval and histograms are constructed with a bin size \( \beta \), dynamically calculated using the Freedman-Diaconis rule:
\begin{equation}
	\beta = \lceil 2 \cdot \text{IQR}(\mathbf{x}_w) \cdot \omega^{-1/3} / (\max(\mathbf{x}_w) - \min(\mathbf{x}_w)) \rceil
\end{equation}
where \(\text{IQR}(\mathbf{x}_w)\) denotes the interquartile range \citep{freedman1981}. Each bin’s probability is estimated as:
\[
p_i = n_i / (\omega + \epsilon),
\]
where \( n_i \) is the bin count and \( \epsilon = 10^{-10} \) is a regularization constant to avoid numerical instability.

We computed four specific entropy measures to comprehensively characterize each windowed segment:

\begin{enumerate}
	\item \textbf{Amplitude Entropy (\( H_a \))}: Calculated directly from the raw signal segment \( \mathbf{x}_w \) using the histogram described above:
	\begin{equation}
		H_a = -\sum_{i=1}^\beta p_i \log(p_i)
	\end{equation}

	\item \textbf{Envelope Entropy (\( H_e \))}: Captures amplitude variations by applying the Shannon entropy formulation to the analytic signal envelope \( |H\{x_w(t)\}| \), where \( x_w(t) \) is the windowed segment of duration \( \omega \). The envelope is obtained via the Hilbert transform, defined as
	
	\begin{equation}
		H\{x_w(t)\} = x_w(t) + j \cdot \frac{1}{\pi} \int_{-\infty}^\infty \frac{x_w(\tau)}{t - \tau} \, d\tau
	\end{equation}
	where \( x_w(t) \) is the original real-valued signal within the window and \( j = \sqrt{-1} \) is the imaginary unit. The envelope is:
	
	\begin{equation}
		|H\{x_w(t)\}| = \sqrt{x_w(t)^2 + \hat{x}_w(t)^2}
	\end{equation}

	where \( \hat{x}_w(t) \) denotes the Hilbert transform of \( x_w(t) \). The envelope entropy is computed from a histogram of \( |H\{x_w(t)\}| \) with \( \beta \) bins, as follows:
	
	\begin{equation}
		H_e = -\sum_{i=1}^\beta p_i \log(p_i)
	\end{equation}
	where $p_i$ is the probability of the envelope amplitude being in the ( i )th bin.

	\item \textbf{Spectral Entropy (\( H_s \))}: Estimated in the frequency domain using Welch’s method to calculate the power spectral density (PSD) of the windowed segments \( \mathbf{x}_w \) \citep{welch1967}. For a discrete signal segment, we compute
	\begin{equation}
		S_{xx}(f) = \frac{1}{K} \sum_{i=1}^K |X_i(f)|^2    
	\end{equation}
	
	where \( X_i(f) \) are the Fourier transforms of \( K \) overlapping segments of \( \mathbf{x}_w \), each of length \( \omega \). We normalized the spectral probabilities as:
	\begin{equation}
		p_i = S_{xx}(f_i) / \left(\sum_{j=1}^M S_{xx}(f_j) + \epsilon\right)
	\end{equation}
	where \( M \) is the number of frequency bins, yielding:
	\begin{equation}
		H_s = -\sum_{i=1}^M p_i \log(p_i)
	\end{equation}

	\item \textbf{Envelope Spectrum Entropy (\( H_{es} \))}: Determined by applying Welch’s method to the envelope of the windowed segment \( x_w(t) \), denoted \( |H\{x_w(t)\}| \). The PSD is computed as:
	\begin{equation}
		S_{\text{env}}(f) = \frac{1}{K} \sum_{i=1}^K |E_i(f)|^2
	\end{equation}
	where \( E_i(f) \) are the Fourier transforms of \( K \) overlapping segments of the envelope \( |H\{x_w(t)\}| \) derived from the Hilbert transform of \( \mathbf{x}_w \). We normalize the probabilities as
	\begin{equation}
		p_i = S_{\text{env}}(f_i) / \left(\sum_{j=1}^M S_{\text{env}}(f_j) + \epsilon\right)
	\end{equation}
	and compute:
	\begin{equation}
		H_{es} = -\sum_{i=1}^M p_i \log(p_i)
	\end{equation}    
\end{enumerate}

We formulate the optimization of window parameters as a constrained search over the parameter space \( \Theta = \{(\omega, \sigma)\} \), where \( \omega \in \Omega \subset \mathbb{Z}^+ \) represents window sizes and \( \sigma \in \Sigma \subset \mathbb{Z}^+ \) represents the step size. The step sizes are derived from the overlap ratios \( \rho \in [0, 1) \) using $\sigma = \lfloor \omega (1 - \rho) \rfloor$.

The objective function is defined as
\begin{equation}
	J(\omega, \sigma) = \frac{\alpha \cdot H_t + (1 - \alpha) \cdot H_f}{\log(1 + \omega)}
\end{equation}
where \( \alpha \) balances the time- and frequency-domain contributions and \( \log(1 + \omega) \) normalizes for the window size. The time-domain component is
$H_t = \alpha_t \cdot H_a + (1 - \alpha_t) \cdot H_e$,
and the frequency-domain component is $H_f = \alpha_s \cdot H_s + (1 - \alpha_s) \cdot H_{es}$ where \( \alpha_t \) and \( \alpha_s \) balance their respective entropy contributions.

The optimal parameters are solved as follows:
\begin{equation}
	(\omega^*, \sigma^*) = \arg\max_{(\omega, \sigma) \in \Theta} J(\omega, \sigma)
\end{equation}
to yield the final optimal window and step sizes for segmenting the time series.

\subsection{Features Extraction}

Following the optimization of the window parameters, we integrate signal processing, machine learning, and graph theory to transform time-series data into a graph representation. In the following, we outline the process, detailing the feature extraction pipeline, and graph construction framework.

First, we extract a comprehensive set of features from each $ch$-channel signal segment \( \mathbf{X}_i \in \mathbb{R}^{N \times ch} \), where the segments are derived from the original two-channel signal \( \mathbf{S} \in \mathbb{R}^{N \times ch} \) using the optimal window size \( \omega^* \) and step size \( \sigma^* \) determined in the previous section. Segmentation, defined as \( \mathbf{X}_i = \mathbf{S}[i \cdot \sigma^* : i \cdot \sigma^* + \omega^*] \), ensures that the features capture the signal characteristics optimized for bearing fault detection by using properties in the time domain and in the frequency domain.

For each segment \( \mathbf{X}_i \), we computed statistical and energy-based metrics. We calculate the mean (\( \mu_{X_i} \)) and standard deviation (\( \sigma_{X_i} \)) as: $\mu_{X_i} = \frac{1}{2N} \sum_{k=1}^{ch} \sum_{j=1}^N X_{i,j,k}$ and $\sigma_{X_i} = \sqrt{\frac{1}{2N} \sum_{k=1}^{ch} \sum_{j=1}^N (X_{i,j,k} - \mu_{X_i})^2}$, where \( X_{i,j,k} \) is the \( j \)th sample of the \( k \)th channel (with \( k = 1, 2, \cdots, ch \)) in segment \( \mathbf{X}_i \). We then compute skewness (\( \gamma_{X_i} \)) and kurtosis (\( \kappa_{X_i} \)) to assess the distribution shape \citep{joanes1998} as:

\begin{equation}
	\gamma_{X_i} = \frac{\mathbb{E}\left[\left(\frac{1}{2} \sum_{k=1}^{ch} (X_{i,j,k} - \mu_{X_i})\right)^3\right]}{\sigma_{X_i}^3}
\end{equation}

\begin{equation}
\kappa_{X_i} = \frac{\mathbb{E}\left[\left(\frac{1}{2} \sum_{k=1}^{ch} (X_{i,j,k} - \mu_{X_i})\right)^4\right]}{\sigma_{X_i}^4}.
\end{equation}

We derived the energy metrics, including the root mean square (\( \text{RMS}_{X_i} \)) and peak amplitude (\( \text{Peak}_{X_i} \)):

\begin{equation}
\text{RMS}_{X_i} = \sqrt{\frac{1}{2N} \sum_{k=1}^{ch} \sum_{j=1}^N X_{i,j,k}^2}
\end{equation}

\begin{equation}
\text{Peak}_{X_i} = \max_{k,j} |X_{i,j,k}|
\end{equation}

To capture impulsive behavior indicative of faults, we applied the Teager-Kaiser Energy operator (\( \text{TKEO}_{X_i,k} \)) \citep{kaiser1990}, computed channel-wise, and averaged:
\begin{equation}
\text{TKEO}_{X_i,k} = \frac{1}{N-2} \sum_{j=1}^{N-2} (X_{i,j+1,k}^2 - X_{i,j,k} X_{i,j+2,k})
\end{equation}

\begin{equation}
	\text{TKEO}_{X_i} = \frac{1}{2} \sum_{k=1}^{ch} \text{TKEO}_{X_i,k}
\end{equation}
emphasizing the rapid amplitude changes.

Next, we analyzed the frequency domain using the Fast Fourier Transform (FFT) and Welch’s method. We computed the magnitude spectrum (\( \mathbf{S}_{F,i,k} \)) for each channel of segment \( \mathbf{X}_i \):
\[
\mathbf{S}_{F,i,k} = |\mathcal{F}\{\mathbf{X}_{i,:,k}\}|, \quad k = 1, 2,\cdots, ch
\]
where \( \mathbf{X}_{i,:,k} \) is the \( k \)-th channel of \( \mathbf{X}_i \) and the power spectral density(PSD) (\( P_{X_iX_i}(f) \)) using Welch’s method \citep{welch1967}:

\begin{equation}
P_{X_iX_i}(f) = \frac{1}{K_w} \sum_{m=1}^{K_w} |\mathcal{F}\{\mathbf{X}_{i,m}\}|^2	
\end{equation}
where \( \mathbf{X}_{i,m} \) are windowed subsegments of \( \mathbf{X}_i \) with 50\% overlap, and \( K_w \) is the number of subsegments. To isolate the amplitude modulations, we extracted the envelope spectrum (\( A_{X_i,k}(t) \)) via the Hilbert transform \citep{hilbert1930}:

\begin{equation}
	\begin{aligned}
		A_{X_i,k}(t) &= \left|\mathcal{H}\left\{\mathbf{X}_{i,:,k}(t)\right\}\right| \\
		&= \sqrt{\mathbf{X}_{i,:,k}^2(t) + \left( \frac{1}{\pi} \, \text{p.v.} \int_{-\infty}^\infty \frac{\mathbf{X}_{i,:,k}(\tau)}{t - \tau} \, d\tau \right)^2} \\
		A_{X_i}(t)   &= \frac{1}{2} \sum_{k=1}^{ch} A_{X_i,k}(t)
	\end{aligned}
	\label{eq:single}
\end{equation}

where \(\text{p.v.}\) denotes the Cauchy principal value that enhances the detection of periodic fault signatures.

The selected time-frequency features align with the established bearing fault detection literature, particularly the feature sets recommended by \citep{randall2011rolling} for vibration analysis and \citep{zheng2022spectral} for envelope spectrum diagnostics.

Using the optimized window size \( \omega^* \) and step size \( \sigma^* \) from the previous section, we segmented the signal \( \mathbf{S} \):
\[
\mathbf{X}_i = \mathbf{S}[i \cdot \sigma^* : i \cdot \sigma^* + \omega^*],
\]
producing segments \( \mathbf{X}_i \). For each segment, we extracted a feature vector \( \mathbf{F}_i \) by combining the time- and frequency-domain features. We then cluster these vectors using MiniBatch K-Means \citep{sculley2010} to group similar signal behaviors efficiently:
\begin{equation}
\arg\min_{\mathbf{C}} \sum_{i=1}^n \min_{\boldsymbol{\mu}_j \in \mathbf{C}} \|\mathbf{F}_i - \boldsymbol{\mu}_j\|^2	
\end{equation}
where \( \mathbf{C} \) denotes the set of cluster centroids and \( n \) denotes the number of segments. This method reduces the computational complexity from \( O(n^2) \) to approximately \( O(n) \) by processing the data in small batches to ensure scalability for large datasets. Clustering identifies representative patternssuch as fault-specific signatures, simplifying the feature space for subsequent graph construction, providing a basis for modular subgraphs, and enhancing robustness to noise. These clusters facilitate efficient k-nearest neighbor (k-NN) graph building and fault detection by preorganizing segments into meaningful groups. 

\subsection{K-Nearest Neighbor Graph Construction}

We construct k-nearest neighbor (k-NN) graphs to define the connectivity between segments \citep{cover1967}, where each node represents a segment characterized by its feature vector \( \mathbf{F}_i \). Edge set \( \mathcal{E}_G \) is defined as follows:
\begin{equation}
	\mathcal{E}_G = \{(i,j) \mid j \in \text{Top-}k(\text{NN}(\mathbf{F}_i)), \, d(\mathbf{F}_i, \mathbf{F}_j) < \tau\}
\end{equation}
where each node (segment) is connected to its \( k_{\text{NN}} \) nearest neighbors based on the Euclidean distance \( d(\mathbf{F}_i, \mathbf{F}_j) = \|\mathbf{F}_i - \mathbf{F}_j\|_2 \) subject to a distance threshold \( \tau \). The edge weights, capturing feature-space similarity, are defined as
\begin{equation}
	w_{ij} = \frac{1}{1 + \|\mathbf{F}_i - \mathbf{F}_j\|_2}
\end{equation}

To improve computational efficiency, we leveraged the MiniBatch K-Means clustering results from the previous section, reducing the number of pairwise distance comparisons by prioritizing connections within and between clusters. Large clusters were decomposed into modular subgraphs
\begin{eqnarray}
\mathcal{G} = \{\mathcal{G}_1, \mathcal{G}_2, \dots, \mathcal{G}_{m_G}\}, \quad |\mathcal{G}_l| \leq N_{\text{max}}	
\end{eqnarray}
where \( m_G \) is the number of subgraphs and \( N_{\text{max}} \) is a hyperparameter defining the maximum number of nodes allowed in a subgraph. This decomposition ensures computational efficiency while preserving the structural coherence for fault analysis.

\subsection{Graph-Theoretic Features}

Each subgraph \( \mathcal{G}_l = (\mathcal{V}_l, \mathcal{E}_{\mathcal{G}_l}, \mathbf{w}_l) \) consists of a node set \( \mathcal{V}_l \) representing segments, an edge set \( \mathcal{E}_{\mathcal{G}_l} \) that captures functional dependencies, and a weight function \( \mathbf{w}_l = \{w_{uv} \mid (u,v) \in \mathcal{E}_{\mathcal{G}_l}\} \), where \( w_{uv} \) are the edge weights previously defined to ensure consistency with the graph topology.

Each node \( v \in \mathcal{V}_l \) corresponds to a segment and carries a feature vector \( \mathbf{F}_v \in \mathbb{R}^d \) that encodes local properties such as time-frequency characteristics, along with a multiclass fault label \( \ell_v \in \{0, 1, \ldots, K_F-1\} \), where \( \ell_v = 0 \) represents normal operation, and \( 1, \ldots, K_F-1 \) corresponds to distinct fault types.

We extract three key graph-theoretic metrics—average shortest path length (\( L_{\text{avg}} \)), modularity (\( Q_{\text{mod}} \)), and spectral gap (\( \Delta_{\text{spec}} \))—to characterize the system’s structural properties and enable fault detection. These metrics provide complementary insights into fault-induced changes: \( L_{\text{avg}} \) reveals disruptions in the information flow, \( Q_{\text{mod}} \) identifies the clustering of fault types, and \( \Delta_{\text{spec}} \) assesses network resilience and fault propagation.

For each connected subgraph \( \mathcal{G}_l = (\mathcal{V}_l, \mathcal{E}_{\mathcal{G}_l}, \mathbf{w}_l) \) with \( |\mathcal{V}_l| > 1 \), the average shortest path length is computed as follows:
\begin{equation}
	L_{\mathcal{G}_l} = \frac{1}{|\mathcal{V}_l|(|\mathcal{V}_l| - 1)} \sum_{u \neq v \in \mathcal{V}_l} d_{\text{path}}(u, v)
\end{equation}
where \( d_{\text{path}}(u, v) \) is the weighted shortest path between nodes \( u \) and \( v \) given by

\begin{equation}
d_{\text{path}}(u, v) = \min_{\text{path } p: u \to v} \sum_{(s,t) \in p} w_{st}	
\end{equation}

We aggregate this across all connected subgraphs as follows:
\begin{equation}
	L_{\text{avg}} = \frac{1}{|\mathcal{C}_{\text{conn}}|} \sum_{\mathcal{G}_l \in \mathcal{C}_{\text{conn}}} L_{\mathcal{G}_l}
\end{equation}
where \( \mathcal{C}_{\text{conn}} = \{ \mathcal{G}_l \mid \mathcal{G}_l \text{ is connected}, |\mathcal{V}_l| > 1 \} \). We also computed the average degree as follows:
\begin{eqnarray}
	d_{\text{avg}} = \frac{1}{m_G} \sum_{\mathcal{G}_l} \frac{2|\mathcal{E}_{\mathcal{G}_l}|}{|\mathcal{V}_l|}
\end{eqnarray}
and the expected small-world path length.

\begin{equation}
	L_{\text{exp-sw}} = \frac{\log N_{\text{total}}}{\log d_{\text{avg}}}, \quad N_{\text{total}} = \sum_l |\mathcal{V}_l|
\end{equation}
by following the small-world network model \citep{watts1998collective}. The metric \( L_{\text{avg}} \) quantifies the information or signal propagation efficiency, where an increase signals fault-induced disconnections or delays, which are critical for detecting degraded system performance across multiple fault types.

Modularity metric \( Q_{\text{mod}} \) evaluates the community structure of each subgraph \( \mathcal{G}_l \) by partitioning nodes into \( K_F \) communities based on their multiclass fault labels. Defining partition \( \mathcal{P}_l = \{ C_0, C_1, \ldots, C_{K_F-1} \} \), where \( C_k = \{ v \in \mathcal{V}_l \mid \ell_v = k \} \) for \( k = 0, \ldots, K_F-1 \), we obtain

\begin{equation}
	Q_{\mathcal{G}_l} = \frac{1}{2|\mathcal{E}_{\mathcal{G}_l}|} \sum_{u, v \in \mathcal{V}_l} \left[ A_{\mathcal{G}_l}(u,v) - \frac{k_u k_v}{2|\mathcal{E}_{\mathcal{G}_l}|} \right] \delta(\ell_u, \ell_v)
\end{equation}
where \( A_{\mathcal{G}_l}(u,v) = w_{uv} \) if \( (u,v) \in \mathcal{E}_{\mathcal{G}_l} \) (else 0), \( k_u = \sum_v A_{\mathcal{G}_l}(u,v) \) is the weighted degree, and \( \delta(\ell_u, \ell_v) = 1 \) if \( \ell_u = \ell_v \) (else 0) following \citep{newman2006modularity}. We consider a system-wide mean

\begin{equation}
	Q_{\text{mod}} = \frac{1}{|\mathcal{E}_{\text{nonempty}}|} \sum_{\mathcal{G}_l \in \mathcal{E}_{\text{nonempty}}} Q_{\mathcal{G}_l}
\end{equation}
where \( \mathcal{E}_{\text{nonempty}} = \{ \mathcal{G}_l \mid |\mathcal{E}_{\mathcal{G}_l}| > 0 \} \). Modularity \( Q_{\text{mod}} \) measures how well fault types cluster into distinct communities, where a decrease signals structural disruptions or overlaps between fault classes, aiding in identifying fault patterns.

For each subgraph \( \mathcal{G}_l \) with \( |\mathcal{V}_l| > 1 \), we construct the weighted Laplacian matrix \( \mathbf{L}_{\mathcal{G}_l} \in \mathbb{R}^{|\mathcal{V}_l| \times |\mathcal{V}_l|} \):
\begin{equation}
	\mathbf{L}_{\mathcal{G}_l}(u, v) =
	\begin{cases} 
		-w_{uv} & \text{if } (u, v) \in \mathcal{E}_{\mathcal{G}_l}, \\
		\sum_{z \in \mathcal{V}_l} w_{uz} & \text{if } u = v, \\
		0 & \text{otherwise}.
	\end{cases}
\end{equation}
Here, \( \mathbf{L}_{\mathcal{G}_l} \) encodes the structure of \( \mathcal{G}_l \), combining its adjacency information and edge weights, such that \( \mathbf{L}_{\mathcal{G}_l} = \mathbf{D}_{\mathcal{G}_l} - \mathbf{A}_{\mathcal{G}_l} \), where \( \mathbf{A}_{\mathcal{G}_l} \) is the weighted adjacency matrix with \( \mathbf{A}_{\mathcal{G}_l}(u,v) = w_{uv} \) if \( (u,v) \in \mathcal{E}_{\mathcal{G}_l} \) (else 0), and \( \mathbf{D}_{\mathcal{G}_l} \) is the diagonal degree matrix with \( \mathbf{D}_{\mathcal{G}_l}(u,u) = \sum_{z \in \mathcal{V}_l} w_{uz} \). We obtain the eigenvalues \( 0 = \lambda_{\mathcal{G}_l,1} \leq \lambda_{\mathcal{G}_l,2} \leq \cdots \leq \lambda_{\mathcal{G}_l,|\mathcal{V}_l|} \) of \( \mathbf{L}_{\mathcal{G}_l} \), where \( \lambda_{\mathcal{G}_l,1} = 0 \) owing to the connectivity of \( \mathcal{G}_l \) and \( \lambda_{\mathcal{G}_l,2} \), the algebraic connectivity, reflects how well connected \( \mathcal{G}_l \) is. We define the spectral gap for \( \mathcal{G}_l \) as
\[
\Delta_{\mathcal{G}_l} = \lambda_{\mathcal{G}_l,2} - \lambda_{\mathcal{G}_l,1} = \lambda_{\mathcal{G}_l,2},
\]
leveraging the spectral graph theory \citep{cheeger1970lower}. We computed the system-level average as follows:

\begin{equation}
\Delta_{\text{spec}} = \frac{1}{|\mathcal{C}_{\text{conn}}|} \sum_{\mathcal{G}_l \in \mathcal{C}_{\text{conn}}} \Delta_{\mathcal{G}_l}	
\end{equation}

The spectral gap \( \Delta_{\text{spec}} \) reflects the network’s robustness and recovery capacity, and a smaller gap (low \( \lambda_{\mathcal{G}_l,2} \)) indicates that \( \mathcal{G}_l \) is more susceptible to disconnection, suggesting vulnerability to fault propagation and cascading failures, whereas a larger gap implies greater resilience, which is essential for distinguishing the fault severity and system-wide impact across multiple classes \citep{chung1996lectures}.

\subsection{Multiclass Fault Detection Model}

To enable multiclass fault detection, we construct a comprehensive feature set that integrates both local and global information. Specifically, for each node \( v \), we define a feature vector: \[\mathbf{z}_v = [\mathbf{F}_v^\top, L_{\text{avg}}, Q_{\text{mod}}, \Delta_{\text{spec}}]^\top \in \mathbb{R}^{d+3},\]

where \( \mathbf{F}_v \) captures node-level attributes, such as time-frequency characteristics, whereas \( L_{\text{avg}} \), \( Q_{\text{mod}} \), and \( \Delta_{\text{spec}} \) are graph-theoretic metrics that encode system-level structural properties. The label for each node remains \( \ell_v' = \ell_v \), with \( \ell_v \in \{0, 1, \ldots, K-1\} \) indicating one of \( K \) fault categories. These features and labels are assembled into a matrix \( \mathbf{Z} \in \mathbb{R}^{N_{\text{total}} \times (d+3)} \) and a label vector \( \mathbf{y}_{\text{label}} \in \{0, 1, \ldots, K-1\}^{N_{\text{total}}} \), where \( N_{\text{total}} \) denotes the total number of nodes in all subgraphs.

We then formulate a multiclass classification problem, where a model \( f(\mathbf{z}_v; \theta) \), parameterized by \( \theta \), maps a feature vector \( \mathbf{z}_v \in \mathbb{R}^{d+3} \) to a score vector \( [s_0(\mathbf{z}_v), \ldots, s_{K-1}(\mathbf{z}_v)] \). In probabilistic models, these scores are converted into class probabilities using the softmax function:
\begin{equation}
		P(\ell_v = k \mid \mathbf{z}_v; \theta) = \frac{\exp(s_k(\mathbf{z}_v))}{\sum_{j=0}^{K-1} \exp(s_j(\mathbf{z}_v))} 	
\end{equation}
while non-probabilistic models rely on decision functions or ensemble voting strategies \citep{hastie2009elements}. The predicted label is selected as \( \hat{\ell}_v = \arg\max_k s_k(\mathbf{z}_v) \).

To train the model, we minimized the regularized loss function:

\begin{equation}
	L(\theta) = \frac{1}{N} \sum_{i=1}^N \ell(f(\mathbf{z}_i; \theta), \ell_i) + \lambda R(\theta)
\end{equation}

where \( \ell(\cdot, \cdot) \) is a task-specific loss (e.g., cross-entropy, hinge loss), and \( R(\theta) \) is a regularization term weighted by \( \lambda \) to prevent overfitting \citep{bishop2006pattern}.

To enhance generalizability across operating conditions, we adopted a transfer learning strategy. Let the source domain be \( \mathcal{D}_S = \{(\mathbf{z}_i^S, \ell_i^S)\}_{i=1}^{N_S} \) and the target domain be \( \mathcal{D}_T = \{(\mathbf{z}_i^T, \ell_i^T)\}_{i=1}^{N_T} \), where both domains share the same feature space (\( \mathbb{R}^{d+3} \)) and label space (\( \{0, 1, \ldots, K-1\} \)), but differ in data distributions owing to domain shifts (e.g., FFT frequency variations from 50–100 Hz to 55–95 Hz) \citep{zhang2017transfer}. The model is first trained on \( \mathcal{D}_S \) by minimizing

\begin{equation}
	L(\theta) = \frac{1}{N_S} \sum_{i=1}^{N_S} \ell(f(\mathbf{z}_i^S; \theta), \ell_i^S) + \lambda R(\theta)
\end{equation}

and then adapted to \( \mathcal{D}_T \) by initializing \( \theta_T^{(0)} = \theta_S \) and fine-tuning via:

\begin{equation}
	L(\theta) = \frac{1}{N_T} \sum_{i=1}^{N_T} \ell(f(\mathbf{z}_i^T; \theta), \ell_i^T) + \lambda R(\theta)
\end{equation}

To ensure transparency and interpretability in fault diagnosis, our framework incorporates interpretable machine learning models, including Linear Regression, Support Vector Machines (SVM), and Random Forests \citep{hastie2009elements}. These models deliver robust classification performance while providing insights into decision boundaries and feature importance, making them well-suited for practical deployment in industrial settings. To comprehensively assess the predictive power of the extracted features, we will evaluate three classifiers: Linear Regression, SVM, and Random Forest.

\subsection{Algorithmic Overview and Complexity Analysis}

The proposed framework for multiclass fault detection processes multichannel vibration signals through five stages: feature extraction, clustering, k-nearest neighbor (KNN) graph construction, graph-theoretic feature extraction, and fault classification. The whole pipeline is defined by Algorithm \ref{alg:graph_fault}. This framework and pipeline integrate signal processing with graph-based techniques to enable robust fault detection in complex systems.

\begin{algorithm}[h!]

	\caption{Graph‑Based Fault Diagnosis Pipeline}
	\label{alg:graph_fault}
	\KwData{Raw multichannel vibration signal $S \in \mathbb{R}^{N\times \text{ch}}$, labels $\ell$ (if supervised)}
	\KwResult{Predicted fault labels $\hat\ell$}
	
	\SetKwBlock{SegBlock}{Entropy‑Driven Segmentation}{}
	\SegBlock{
		Estimate candidate window sizes $\Omega$ and overlap ratios $\rho$\;
		\ForEach{$(\omega,\rho)\in\Omega\times\rho$}{
			$\sigma \leftarrow \lfloor \omega\,(1-\rho)\rfloor$\;
			Segment $S$ into $\{X_i\}$ with window $\omega$ and step $\sigma$\;
			Compute entropy scores $H_t, H_f$ for each segment\;
			Compute $J(\omega,\sigma) = \bigl[\alpha\,H_t + (1-\alpha)\,H_f\bigr]/\log(1+\omega)$\;
		}
		$(\omega^*,\sigma^*) \leftarrow \arg\max J$\;
	}
	
	\SetKwBlock{FeatBlock}{Feature Extraction}{}
	\FeatBlock{
		Segment $S$ into $\{X_i\}$ using $(\omega^*,\sigma^*)$\;
		\ForEach{segment $X_i$}{
			Compute time‑domain stats: mean, std, skewness, kurtosis, RMS, peak\;
			Compute TKEO energy\;
			Compute FFT‑based PSD and envelope spectrum via Hilbert transform\;
			Concatenate into feature vector $F_i\in\mathbb{R}^d$\;
		}
	}
	
	\SetKwBlock{GraphBlock}{Clustering \& k‑NN Graph Construction}{}
	\GraphBlock{
		Run MiniBatch K‑Means on $\{F_i\}$ to obtain $K_c$ clusters\;
		\ForEach{cluster $C_j$}{
			$V_j \leftarrow \{\,i : F_i \in C_j\}$\;
			Build k‑NN graph $G_j=(V_j,E_j)$:\;
			\ForEach{node $u\in V_j$}{
				Find $k$ nearest neighbors in feature space\;
				Add weighted edges $(u,v)$ if $\mathrm{dist}(u,v)<\tau$\;
			}
			\If{$|V_j|>N_{\max}$}{split $C_j$ further\;}
		}
	}
	
	\SetKwBlock{GraphFeat}{Graph‑Theoretic Feature Computation}{}
	\GraphFeat{
		Initialize lists $L,Q,\Delta$\;
		\ForEach{subgraph $G_j$}{
			Compute average shortest path length $L_j$\;
			Compute modularity $Q_j$ w.r.t.\ known labels $\ell$\;
			Compute spectral gap $\Delta_j = \lambda_2(\text{Laplacian}(G_j))$\;
			Append $L_j\to L$, $Q_j\to Q$, $\Delta_j\to\Delta$\;
		}
	}
	
	\SetKwBlock{ClassBlock}{Classification / Transfer Learning}{}
	\ClassBlock{
		Form feature matrix $Z\in\mathbb{R}^{N_{\text{total}}\times(d+3)}$ by augmenting each $F_i$ with $(\overline{L},\overline{Q},\overline{\Delta})$\;
		\If{transfer learning}{
			Pretrain on source domain $(Z_S,\ell_S)$ using Random Forest\;
			Fine‑tune on target domain $(Z_T,\ell_T)$\;
		}
		\Else{
			Train Random Forest classifier on $(Z,\ell)$\;
		}
		$\hat\ell \leftarrow \text{RF.predict}(Z)$\;
		\Return $\hat\ell$\;
	}
	
\end{algorithm}

Feature extraction segments a \( ch \)-channel vibration signal \(\mathbf{S} \in \mathbb{R}^{N \times ch}\) into \( n = \lfloor (N - \omega^*) / \sigma^* \rfloor \) segments by using the optimal window size \(\omega^*\) and step size \(\sigma^*\). Each segment yields a feature vector of dimension \( d = ch \cdot d_c \), where \( d_c \) is the number of features per channel (e.g., \( d = 40 \) for \( ch = 2 \), \( d_c = 20 \)). The FFT, the dominant operation, has complexity \( O(\omega^* \cdot \log \omega^*) \) per segment per channel, resulting in a total complexity of \( O(ch \cdot n \cdot \omega^* \cdot \log \omega^*) \).

The \( n \) feature vectors are clustered into \( K_c \) groups using MiniBatch K-means, with complexity \( O(n \cdot ch \cdot d_c \cdot K_c) \), enabling efficient pattern identification. For each cluster, a KNN graph connects each node to its \( k_{\text{NN}} \) nearest neighbors within a threshold \(\tau\), with large clusters decomposed into subgraphs of size at most \( N_{\text{max}} \). With \( n / K_c \) nodes per cluster, the complexity is \( O\left( \frac{n^2}{K_c} \left( ch \cdot d_c + \log (n / K_c) \right) \right) \), dominating the framework's cost.

Graph-theoretic features are extracted from \( m_G = \sum_{i=1}^{K_c} \lceil (n / K_c) / N_{\text{max}} \rceil \) subgraphs, each containing approximately \( |V_l| \approx n / m_G \) nodes. Dijkstra's algorithm, applied to compute all pairs of shortest paths, has a complexity of \( O\left( \frac{n^2}{m_G} \log (n / m_G) \right) \). The final classification stage, using classifiers such as Support Vector Machines or Random Forests, scales linearly with \( n \), contributing lower-order terms. For a small number of channels (\( ch \)) and moderate \( n \), the KNN graph construction with complexity \( O(n^2 / K_c) \) often dominates. However, for large \( ch \) and \( n \), feature extraction with complexity \( O(ch \cdot n \cdot \omega^* \cdot \log \omega^*) \) becomes dominant, owing to its high computational cost.

\section{Experiment and Result Analysis}

\subsection{Datasets}
We employ the Case Western Reserve University (CWRU) bearing dataset and the bearing and  gearbox dataset from Southeast University (SU), China, which are widely used in research of rotating machinery fault diagnosis.

\begin{table*}[h!]
	\centering
	\caption{Specifications of the datasets used in the analysis. Source: Authors own work.}
		\scriptsize % reduces font size for better fit
	\begin{tabular}{l p{1.5cm} p{1.5cm} p{2cm} p{3.5cm} p{2.5cm}}
		\hline
		\textbf{Motor Load} & \textbf{Samples}& \textbf{Channels} & \textbf{Total Data Points} & \textbf{Fault Types} & \textbf{Fault Severities} \\
		\hline
		1 Hp & 4,753 & 2 & 4,867,072 & \begin{tabular}[t]{@{}l@{}} Ball, Inner Race, \\ Outer Race, Healthy \end{tabular} & \begin{tabular}[t]{@{}l@{}} 0.007", 0.014", \\ 0.021" \end{tabular} \\
		
		2 Hp & 4,640 & 2 & 4,751,360 & \begin{tabular}[t]{@{}l@{}} Ball, Inner Race, \\ Outer Race, Healthy \end{tabular} & \begin{tabular}[t]{@{}l@{}} 0.007", 0.014", \\ 0.021" \end{tabular} \\
		
		3 Hp & 4,753 & 2 & 4,867,072 & \begin{tabular}[t]{@{}l@{}} Ball, Inner Race, \\ Outer Race, Healthy \end{tabular} & \begin{tabular}[t]{@{}l@{}} 0.007", 0.014", \\ 0.021" \end{tabular} \\
		
		0 Hp & 4,753 & 2 & 4,867,072 & \begin{tabular}[t]{@{}l@{}} Ball, Inner Race, \\ Outer Race, Healthy \end{tabular} & \begin{tabular}[t]{@{}l@{}} 0.007", 0.014", \\ 0.021" \end{tabular} \\
		\hline
	\end{tabular}
	\label{tab:datasetspecs}
\end{table*}

The dataset(CWRU) bearing dataset used for fault prediction consists of vibration signals collected from bearings under different operating conditions, including different types of faults (ball fault, Inner Race fault, Outer Race fault, and healthy condition) and fault severity (0.007, 0.014 and 0.021 inches). The data were recorded at a sampling frequency of 12 or 48 kHz, meaning each second contained 12,0000, or 48,000 samples. We use four datasets based on the motor load of 0 Hp, 1 HP, 2 HP, and 3 HP, with approximate data points of 4.8 million each. Table \ref{tab:datasetspecs} details the dataset used in the analysis.

\begin{table*}[h!]
	\centering
	\caption{ANOVA and Tukey HSD Results for Vibration Signal RMS Across Load Conditions in the CWRU Dataset. Source: Authors own work.}
	\label{tab:anova_tukey_results}
		\scriptsize % reduces font size for better fit
	\begin{tabular}{| p{2cm}| p{3.5cm}| p{1.5cm}| p{2cm}| p{1.5cm}| p{2cm}|}
		\hline
		\textbf{Test} & \textbf{Groups} & \textbf{Mean Difference} & \textbf{p-value} & \textbf{95\% CI} & \textbf{Significant} \\ \hline
		\multirow{1}{*}{ANOVA} & All Loads (0 HP, 1 HP, 2 HP, 3 HP) & - & $F = 3966122.36$, $p < 0.001$ & - & Yes \\ \hline
		\multirow{6}{*}{Tukey HSD} 
		& 0 HP vs. 1 HP & 0.0495 & $<0.001$ & [0.0494, 0.0496] & Yes$^*$ \\ 
		& 0 HP vs. 2 HP & 0.0993 & $<0.001$ & [0.0992, 0.0995] & Yes$^*$ \\ 
		& 0 HP vs. 3 HP & 0.1492 & $<0.001$ & [0.1491, 0.1494] & Yes$^*$ \\ 
		& 1 HP vs. 2 HP & 0.0498 & $<0.001$ & [0.0497, 0.0500] & Yes$^*$ \\ 
		& 1 HP vs. 3 HP & 0.0998 & $<0.001$ & [0.0996, 0.0999] & Yes$^*$ \\ 
		& 2 HP vs. 3 HP & 0.0499 & $<0.001$ & [0.0498, 0.0500] & Yes$^*$ \\ \hline
	\end{tabular}
	\begin{threeparttable}
		\begin{tablenotes}
			\small
			\item $^*$Significant at $\alpha = 0.05$ (p < 0.05). The ANOVA test indicates significant differences in RMS among load conditions. Tukey HSD post-hoc tests confirm all pairwise comparisons are significant, with p-values less than 0.001.
		\end{tablenotes}
	\end{threeparttable}
\end{table*}
To investigate the effect of load conditions on bearing vibration signals in the Case Western Reserve University (CWRU) dataset, we conducted a one-way ANOVA \citep{fisher1925statistical} on the root mean square (RMS) of vibration signals across four load conditions: 0 HP, 1 HP, 2 HP, and 3 HP. The ANOVA revealed significant differences among the load conditions, with an F-statistic of 3,966,122.36 and a p-value less than 0.001 (Table \ref{tab:anova_tukey_results}). Post-hoc Tukey HSD tests \citep{tukey1949comparing} were performed to identify specific pairwise differences. All pairwise comparisons were statistically significant (p < 0.001), with mean RMS differences ranging from 0.0495 (0 HP vs. 1 HP) to 0.1492 (0 HP vs. 3 HP). These results indicate that the RMS of vibration signals increases with load, with the largest difference observed between 0 HP and 3 HP conditions. This suggests that load significantly influences vibration characteristics, which may impact bearing fault diagnosis models.

The Southeast University (SU) gearbox dataset, collected using a Drivetrain Dynamic Simulator (DDS), comprises two subsets: bearing and gear data. Both subsets were recorded under controlled experimental conditions with two speed–load configurations: 20 Hz–0 V and 30 Hz–2 V. Each data file includes eight signal channels capturing different drivetrain components. Channel 1 represents motor vibration; channels 2, 3, and 4 measure planetary gearbox vibrations in the x, y, and z directions, respectively; channel 5 records motor torque; and channels 6, 7, and 8 capture parallel gearbox vibrations in the x, y, and z directions. 
The dataset includes five distinct working conditions for bearing and gear data: one healthy state and four fault types. This results in a 10-class classification task for fault diagnosis in the DDS, considering both subsets and operating conditions. Table \ref{tab:seu_fault_types} details the fault types in the dataset.

\begin{table}[h!]
	
	\centering
	\caption{Bearing and Gearbox Fault Types Description (Southeast University DDS Dataset), where each point (*) for each type has an equal number of data points for 20 Hz -0V and 30 Hz - 2V configurations, i.e., 1,048,560 data points for each configuration per type.}.
		
	\scriptsize
	\begin{tabular}{|l|l|p{4cm}|p{3cm}|}
		
		\hline
		\textbf{Location} & \textbf{Type}    & \textbf{Description}                                                   & \textbf{Total Data Points} \\ \hline
		Gearbox           & Chipped          & Crack occurs in the gear feet                                          &  2,097,120*                           \\ \hline
		& Miss             & Missing one of the feet in the gear                                    &   2,097,120*                          \\ \hline
		& Root             & Crack occurs in the root of gear feet                                  &   2,097,120*                          \\ \hline
		& Surface          & Wear occurs on the surface of the gear                                 &     2,097,120*                      \\ \hline
		Bearing           & Ball             & Crack occurs in the ball                                               & 2,097,120*                  \\ \hline
		& Inner            & Crack occurs in the inner ring                                         & 2,097,120*                  \\ \hline
		& Outer            & Crack occurs in the outer ring                                         & 2,097,120*                  \\ \hline
		& Combination      & Crack occurs in both the inner and outer rings                          & 2,097,120*                  \\ \hline
	\end{tabular}
	\label{tab:seu_fault_types}
\end{table}

\begin{table*}[h!]
	\centering
	\caption{Statistical Analysis of RMS Vibration Signals for Bearing Conditions (SEU Dataset). Source: Authors own work.}
		\scriptsize % reduces font size for better fit
	\label{tab:combined_stats}
	\begin{threeparttable}
		\begin{tabular}{| >{\raggedright\arraybackslash}p{3cm} | >{\raggedright\arraybackslash}p{4cm} | l | l | p{1.5cm} | l |}
			\hline
			\textbf{Analysis} & \textbf{Groups/Condition} & \textbf{Statistic} & \textbf{p-value} & \textbf{95\% CI} & \textbf{Significant} \\ \hline
			\multirow{7}{3cm}{ANOVA (20 Hz--0 V)} 
			& All Conditions (Ball, Inner, Outer, Health) & $F = 212.99$ & $<0.001$ & - & Yes \\ \cline{2-6}
			& Ball vs. Health & 0.1098 & $<0.001$ & [0.0964, 0.1231] & Yes$^*$ \\ 
			& Ball vs. Inner & -0.0073 & 0.494 & [-0.0207, 0.0060] & No \\ 
			& Ball vs. Outer & 0.0316 & $<0.001$ & [0.0183, 0.0449] & Yes$^*$ \\ 
			& Health vs. Inner & -0.1171 & $<0.001$ & [-0.1304, -0.1037] & Yes$^*$ \\ 
			& Health vs. Outer & -0.0782 & $<0.001$ & [-0.0915, -0.0648] & Yes$^*$ \\ 
			& Inner vs. Outer & 0.0389 & $<0.001$ & [0.0256, 0.0522] & Yes$^*$ \\ \hline
			\multirow{5}{3cm}{T-tests (20 Hz--0 V vs. 30 Hz--2 V)} 
			& Ball & $t = -14.49$ & $<0.001$ & - & Yes$^*$ \\ 
			& Combination & $t = 5.88$ & $<0.001$ & - & Yes$^*$ \\ 
			& Health & $t = 4.98$ & $<0.001$ & - & Yes$^*$ \\ 
			& Inner & $t = 29.32$ & $<0.001$ & - & Yes$^*$ \\ 
			& Outer & $t = 101.86$ & $<0.001$ & - & Yes$^*$ \\ \hline
		\end{tabular}
		\begin{tablenotes}
			\small
			\item $^*$Significant at $\alpha = 0.05$ (p < 0.05). The ANOVA test shows significant differences among bearing conditions at 20 Hz--0 V, with Tukey HSD identifying significant pairs except Ball vs. Inner. T-tests confirm significant RMS differences between 20 Hz--0 V and 30 Hz--2 V for all conditions.
		\end{tablenotes}
	\end{threeparttable}
\end{table*}

Statistical analysis of the Southeast University (SEU) Bearing and Gearbox dataset, conducted on vibration signals from channel 0 (motor vibration) at 2000 Hz, revealed significant differences in root mean square (RMS) values across bearing conditions and speed–load configurations, as summarized in Table \ref{tab:combined_stats}. 

An ANOVA test for Ball, Inner, Outer, and Health conditions at 20 Hz–0 V yielded a significant $(F = 212.99, p < 0.001)$, indicating that bearing condition impacts vibration intensity. Tukey HSD post-hoc tests confirmed significant pairwise differences $(p < 0.001)$ for most pairs—Ball vs. Health (mean diff $= 0.1098$), Ball vs. Outer (0.0316), Health vs. Inner (-0.1171), Health vs. Outer (-0.0782), and Inner vs. Outer (0.0389)-except Ball vs. Inner (p $=$ 0.494), suggesting similar vibration patterns for these faults. 

T-tests comparing 20 Hz–0 V to 30 Hz–2 V across all conditions (Ball, Combination, Health, Inner, Outer) showed significant differences $(p < 0.001)$, with t-statistics ranging from -14.49 (ball) to 101.86 (outer), indicating that a higher speed-load (30 Hz - 2 V) significantly alters RMS, particularly amplifying vibrations for Outer and Inner faults. These findings highlight the importance of considering bearing and operating conditions in fault diagnosis, with Table \ref{tab:combined_stats} providing a comprehensive overview of statistical significance for developing robust diagnostic models.

We perform significance tests on each dataset to ensure the reliability of classification results, especially for simple models used in fault detection. These tests help distinguish real patterns from random noise. A model might seem effective on a dataset, but without statistical validation, its performance could be coincidental. Significance tests provide p-values to confirm that differences—such as RMS variations across fault types—are not due to chance. This validation is crucial in bearing fault diagnosis, where misclassification can cause equipment failure.

\subsection{Experimental Set-up}
We used a graph-based approach to classify bearing faults across the specified datasets, running all experiments on a free Google Colab environment with a dual-core virtual CPU. Vibration signals, sampled at high frequencies across multiple channels, were transformed into graph structures where nodes captured features such as statistical and spectral properties, and edges represented inter-signal relationships. Labels indicated fault types: for example, Ball, Inner, Outer, and Healthy.

To enrich graph representations, we computed small-world properties (e.g., average path length, node degree), modularity (to assess fault-based community structure), and spectral gaps (to evaluate graph connectivity). These metrics were integrated as features for classification.

We employ logistic regression (LR), random forest (RF), and support vector machine (SVM) classifiers to evaluate the effectiveness of the proposed features. Each model is trained on 70\% of the data and tested on the remaining 30\%. To assess classification performance, we use a comprehensive set of evaluation metrics, including accuracy, precision, recall, and F1-score, computed per fault class. Accuracy measures the overall correctness of predictions, while precision reflects the reliability of fault detection—critical for minimizing false positives in maintenance decisions. Recall captures the model’s ability to correctly identify true fault instances, particularly severe types such as inner and outer defects. The F1-score balances precision and recall, which is especially relevant for imbalanced datasets where healthy samples are more prevalent.

To ensure interpretability, we apply SHAP analysis to quantify the contribution of graph-theoretic metrics and node-level features to model predictions. This provides actionable insights into which structural or statistical characteristics influence classification decisions. Additionally, we visualize confusion matrices, SHAP value distributions, and per-stage computation times to demonstrate the framework’s scalability and practical applicability.

\subsection{Result Analysis}

The first step in the proposed architecture involves determining the optimal window size ($w^*$) and step size ($\alpha^*$). As outlined in the methodology section, we estimate these parameters using entropy-based calculations. 

\subsubsection{Optimal Window and Step Size calculation}

\begin{figure*}[h!]
	\centering
	\includegraphics[width=.7\textwidth]{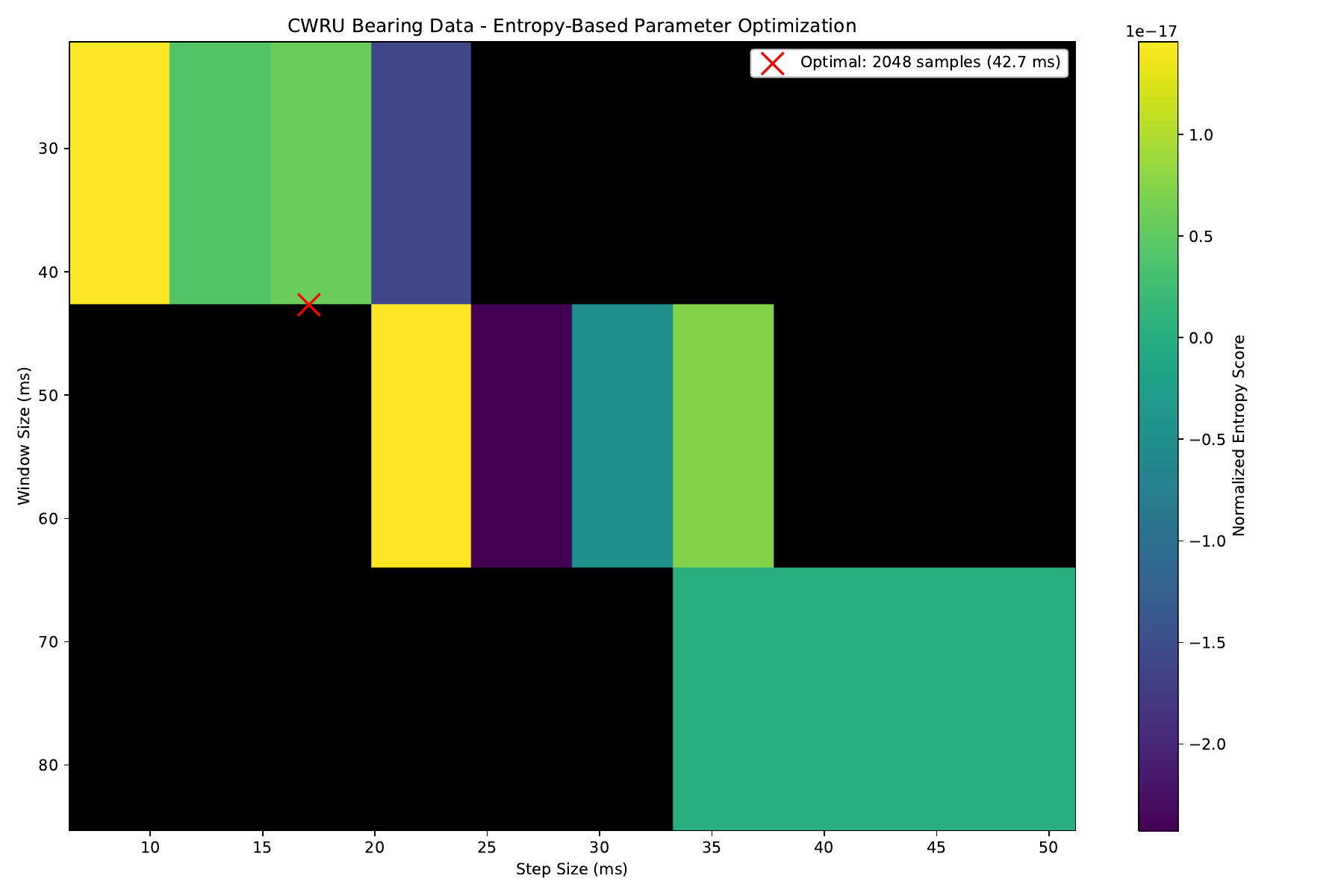}
	\caption{Optimal window and step size for CWRU dataset with 0 Hp. Source: Authors own work.}
	\label{fig:optimWinHp0}
\end{figure*}

The CWRU dataset utilized in this analysis is sampled at 48 kHz, meaning each data point represents approximately 20.83 microseconds. Through entropy-based optimization, an optimal window size of 2048 samples was identified for the 0 Hp load condition (as illustrated in Figure~\ref{fig:optimWinHp0}), corresponding to a time segment of approximately 42.7 milliseconds. The optimal step size was determined to be 1228 samples. Following a similar approach, optimal segmentation parameters were determined for the SU datasets, resulting in a window size of 512 samples and a step size equal to approximately 50\% of the window size.

These window lengths effectively capture multiple cycles of vibration signals caused by bearing defects, preserving critical fault-related features while maintaining high temporal resolution. Several studies support the suitability of this window length. For instance, Zhang et al. \citep{ZHANG2015164} employed window sizes ranging from 1024 to 4096 samples in their CWRU-based bearing fault detection models, highlighting that a 2048-sample window provides a good trade-off between information content and computational efficiency. Moreover, 2048 samples align with the optimal range for extracting spectral features like envelope analysis and FFT, which are sensitive to periodic fault signatures. The power-of-two window size also facilitates efficient implementation in real-time systems using Fast Fourier Transforms (FFT). Therefore, this choice is empirically justified and practically efficient for scalable, accurate fault diagnosis. 

\subsubsection{Graph Construction}
\begin{figure*}[htbp!]
	\centering
	\includegraphics[width=0.8\textwidth]{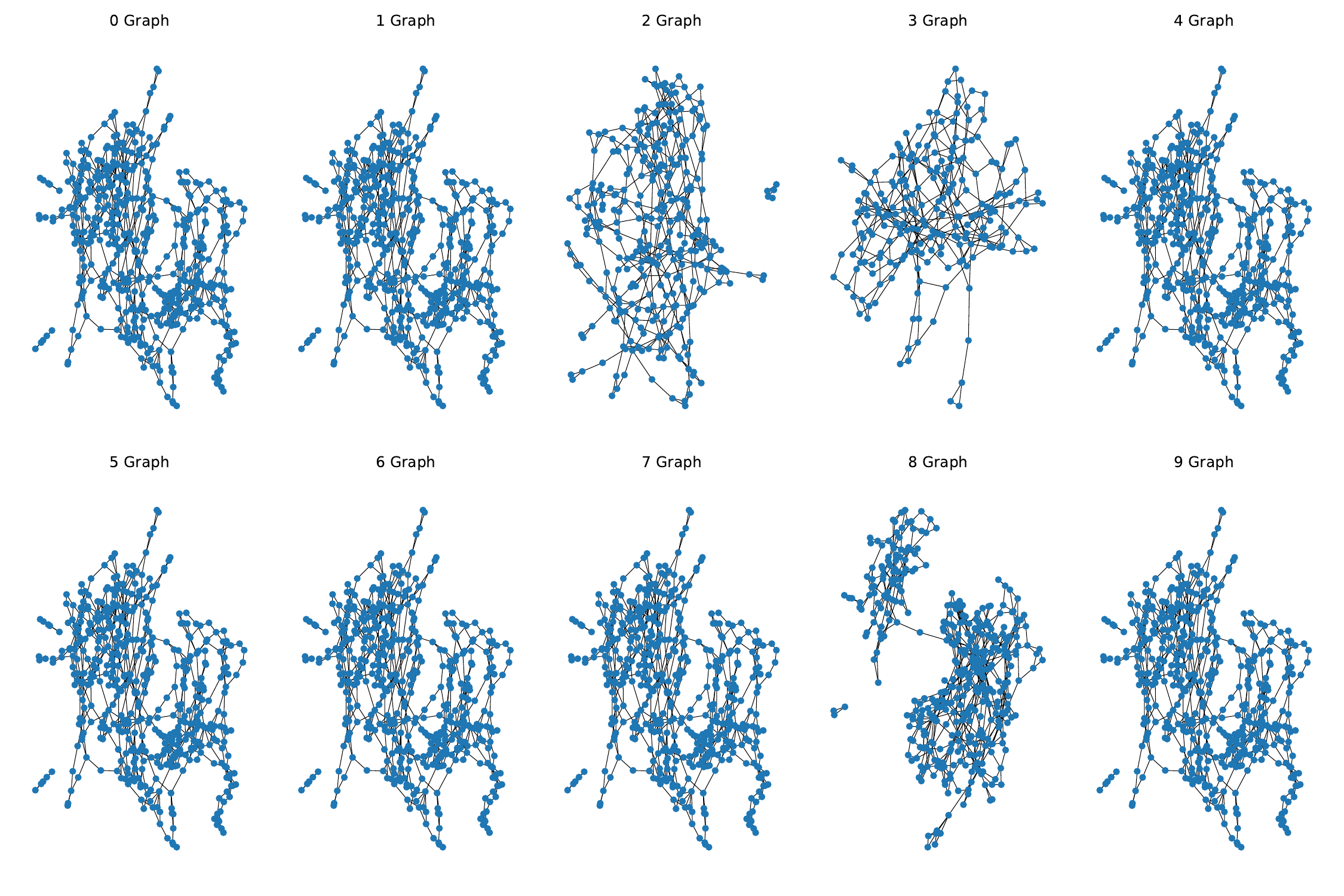}
	\caption{k-Nearest Neighbors (kNN) graph for a cluster split for CWRU 1 Hp Motor Load. Source: Authors own work. }
	\label{fig:knnGraph}
\end{figure*}

We construct a graph composed of multiple subgraphs using optimal window and step sizes. Figure~\ref{fig:knnGraph} shows the k-Nearest Neighbors (kNN) graph for each cluster split in the 1 Hp CWRU dataset. The number of neighbors is dynamically set as \( k_n = \min(k_{\text{max}}, n-1) \), ensuring valid connectivity even in small splits. Splits with fewer than two points are excluded.

\begin{table*}[h!]
	\centering
	\caption{Performance Comparison and Graph Properties for CWRU and Gearbox Datasets. Source: Authors own work.}
		\scriptsize % reduces font size for better fit
	\label{tab:knn_graph}
	\begin{tabular}{
			l  % Dataset
			S[table-format=2.0]  % k_max
			S[table-format=3.0]  % Max Nodes
			S[table-format=2.3]  % <k>
			S[table-format=1.3]  % L
			S[table-format=1.3]  % L_exp
			S[table-format=1.3]  % Q
			S[table-format=1.3]  % Spectral Gap
			S[table-format=1.3]  % Accuracy
			S[table-format=2.2]  % Time
		}
		\toprule
		{Dataset} & 
		{\( k_{\text{max}} \)} & 
		{$N_{max}$} & 
		{\( L \)} & 
		{\( Q \)} & 
		{Spectral Gap} & 
		{Accuracy} & 
		{Time (\si{\second})} \\
		\midrule
		\multirow{6}{*}{CWRU} 
		& 15 & 100    & 0.730  & 0.059 & 0.833 & 0.997 & 5.57 \\
		& 5  & 100    & 1.487      & 0.073 & 0.092 & 0.997 & 1.64 \\
		& \textbf{5}  & \textbf{200}  & \textbf{1.736}    & \textbf{0.127} & \textbf{0.063} & \textbf{0.998} & 3.50 \\
		& 3  & 200   & 2.377     & 0.134 & 0.007 & 0.997 & 1.32 \\
		& 5  & 300 & 1.610 & 0.130 & 0.065 & 0.996 & 5.07 \\
		& 3  & 400     & 2.578     & 0.153 & 0.010 & 0.995 & 1.21 \\
		\midrule
		\multirow{8}{*}{Gearbox} 
		& 3  & 200  & 0.782  & 0.034 & 0.011 & 1.000 &  8.52 \\
		& 3  & 400  & 0.885 & 0.059 & 0.008 & 1.000 & 14.25 \\
		& 5  & 200  & 0.475 & 0.033 & 0.071 & 1.000 & 16.14 \\
		& 5  & 400  & 0.548 & 0.057 & 0.053 & 1.000 & 30.04 \\
		& 8  & 200 & 0.368 & 0.031 & 0.186 & 1.000 & 26.18 \\
		& 8  & 400 & 0.424 & 0.056 & 0.137 & 1.000 & 51.78 \\
		& 11 & 200 & 0.317 & 0.031 & 0.319 & 1.000 & 36.48 \\
		& 11 & 400 & 0.372 & 0.054 & 0.235 & 1.000 & 77.25 \\
		\bottomrule
	\end{tabular}
\end{table*}

To optimize kNN graph structure for fault detection, we varied \( k_{\text{max}} \) and subgraph size (\( |\mathcal{G}_i| \)). A dense baseline (\( k_{\text{max}} = 15 \), \( |\mathcal{G}_i| \leq 100 \)) achieved 0.997 accuracy but lacked small-world characteristics and was slower (5.57 s). Reducing \( k_{\text{max}} \) to 5 improved efficiency (1.64 s) and path length while maintaining accuracy. The best trade-off was found at \( k_{\text{max}} = 5 \), \( |\mathcal{G}_i| \leq 200 \), achieving the highest accuracy (0.998) and modularity (\( Q = 0.127 \)). Smaller \( k_{\text{max}} = 3 \) settings aligned better with small-world metrics and had the lowest runtime.

Larger subgraphs slightly reduce classification accuracy and increase computational cost. The configuration with \( k_{\text{max}} = 3 \) and subgraph size \( |\mathcal{G}_i| \leq 400 \) achieves the best trade-off, aligning well with small-world network characteristics. Table~\ref{tab:knn_graph} summarizes the performance results using a simple logistic regression (LR) baseline. All configurations achieve perfect accuracy for the SU gearbox dataset (1.000); therefore, the setup with the lowest graph construction time is selected for efficiency.

\subsubsection{Classification Analysis}

We evaluated the performance of Support Vector Machine (SVM), Logistic Regression (LR), and Random Forest (RF) models on the CWRU and SU bearing fault datasets. Table \ref{tab:performance} presents their performance on the CWRU dataset across four horsepower loads (0, 1, 2, and 3 Hp) and SU dataset across gearbox and bearing, comparing accuracy, cross-validation (CV) mean accuracy with standard deviation and inference time.

\begin{table}[h!]
	
	\centering
	
	\scriptsize
	
	\caption{Performance comparison of SVM, Logistic Regression, and Random Forest across horsepower loads for CWRU and SU datasets.}
	
	\label{tab:performance}
	
	\begin{tabular}{llccc}
		
		\toprule
		
		\textbf{Dataset} & \textbf{Model} & \textbf{Acc.} & \textbf{CV Mean Acc. ($\pm$ SD)} & \textbf{Infer. Time (s)} \\
		
		\midrule
		
		\multirow{3}{*}{0 Hp} & SVM & 98.7 & $98.8 \pm 0.9$ & 0.0274 \\
		
		& Logistic Regression & 98.7 & $98.8 \pm 0.9$ & \textbf{0.0007} \\
		
		& Random Forest & \textbf{100.0} & \textbf{100.0 $\pm$ 0.000} & 0.0092 \\
		
		\midrule
		
		\multirow{3}{*}{1 Hp} & SVM & 99.0 & $99.6 \pm 0.1$ & 0.2816 \\
		
		& Logistic Regression & 99.4 & $99.6 \pm 0.2$ & \textbf{0.0009} \\
		
		& Random Forest & \textbf{99.9} & \textbf{99.9 $\pm$ 0.1} & 0.0170 \\
		
		\midrule
		
		\multirow{3}{*}{2 Hp} & SVM & 99.3 & $99.3 \pm 0.2$ & 1.3235 \\
		
		& Logistic Regression & 99.5 & $99.6 \pm 0.3$ & \textbf{0.0142} \\
		
		& Random Forest & \textbf{99.7} & \textbf{99.8 $\pm$ 0.2} & 0.0325 \\
		
		\midrule
		
		\multirow{3}{*}{3 Hp} & SVM & 99.8 & $99.7 \pm 0.2$ & 0.7218 \\
		
		& Logistic Regression & 99.9 & $99.7 \pm 0.2$ & \textbf{0.0167} \\
		
		& Random Forest & \textbf{100.0} & \textbf{100.0 $\pm$ 0.1} & 0.0233 \\
		
		\midrule
		
		\multirow{3}{*}{Gearbox} & SVM & \textbf{100.0} & $100.0 \pm 0.000$ & 0.001165 \\
		
		& Logistic Regression & \textbf{100.0} & $100.0 \pm 0.000$ & \textbf{0.000002} \\
		
		& Random Forest & \textbf{100.0} & \textbf{100.0 $\pm$ 0.000} & 0.000010 \\
		
		\midrule
		
		\multirow{3}{*}{Bearing} & SVM & 99.9 & $99.9 \pm 0.000$ & 0.001074 \\
		
		& Logistic Regression & \textbf{100.0} & $100.0 \pm 0.000$ & \textbf{0.000002} \\
		
		& Random Forest & \textbf{100.0} & \textbf{100.0 $\pm$ 0.000} & 0.000021 \\
		
		\bottomrule
		
	\end{tabular}
	
\end{table}

The results in Table \ref{tab:performance} highlight the strengths and trade-offs of each model for the CWRU dataset. Random Forest consistently achieved the highest accuracy (99.7-100.0) and CV mean accuracy (99.8-100.0, $\pm$0.000-0.2), with perfect classification at 0 and 3 Hps. Logistic Regression delivered strong accuracy (98.7-099.9) and CV mean accuracy (98.8-99.7, $\pm$0.2-0.9) and had the fastest inference times (0.0007-0.0167 s). This speed is critical for real-time applications, such as bearing fault diagnosis, in which rapid decision-making is essential. SVM, while reliable, showed slightly lower accuracy (0.987-0.998) and CV mean accuracy (98.8-99.7, $\pm$0.1-0.9) and the slowest inference times (0.0274-1.3235 s), especially at 2 Hp, making it less suitable for time-sensitive tasks.  

\begin{figure}[h!]
	\includegraphics[width=\textwidth]{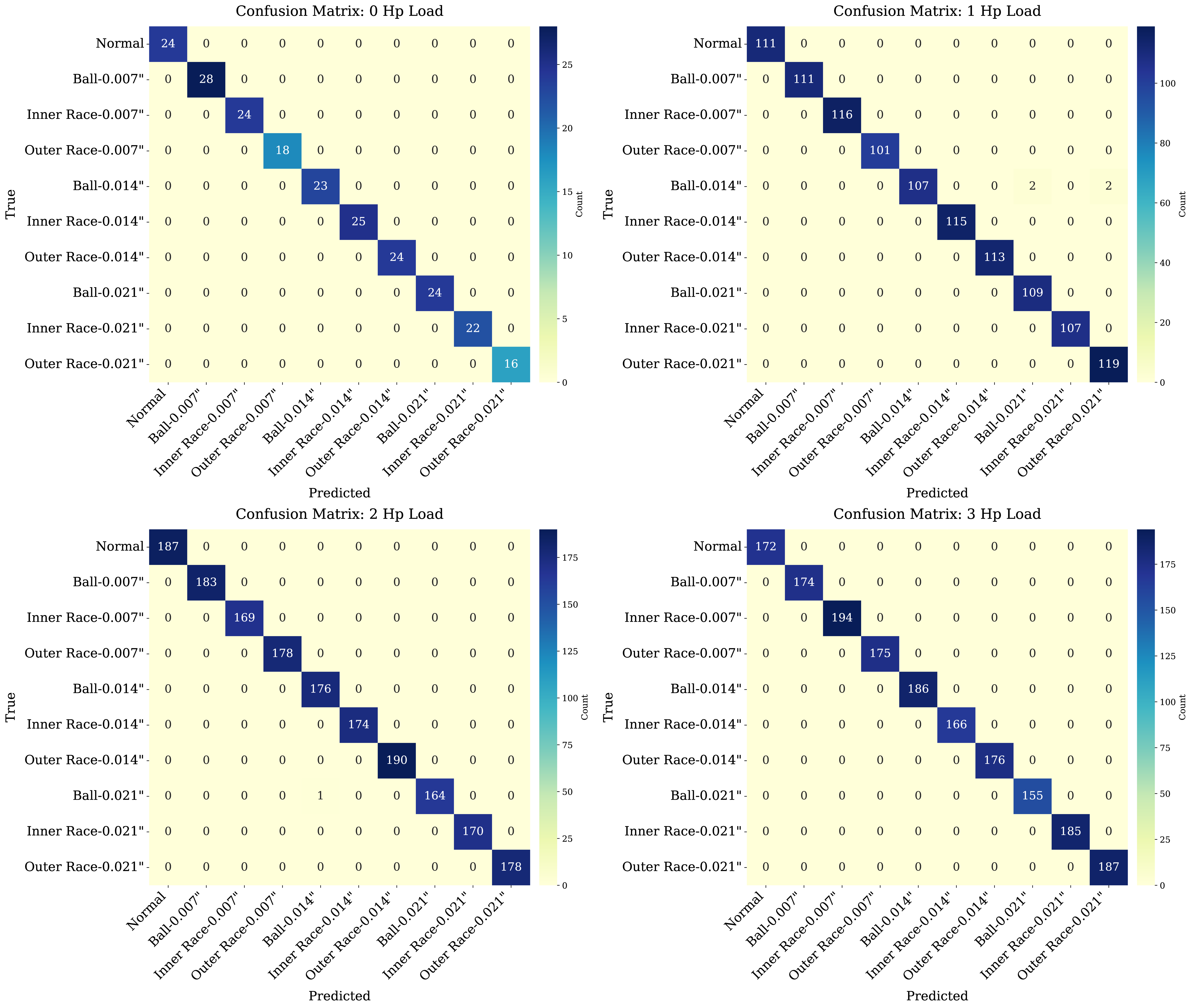} 
	\caption{Confusion matrices of the Random Forest classifier evaluated on the CWRU bearing dataset under four different motor loads: 0 Hp, 1 Hp, 2 Hp, and 3 Hp. Each matrix illustrates the model’s performance in classifying various bearing fault types (Ball, Inner Race, Outer Race) with different defect sizes, as well as the healthy condition. The strong diagonal dominance across all loads indicates the model’s high accuracy and robustness across varying operating conditions.} \label{fig:heatmapcwru}
\end{figure}

\begin{figure}[h!]
	\centering
	\includegraphics[width=0.6\textwidth]{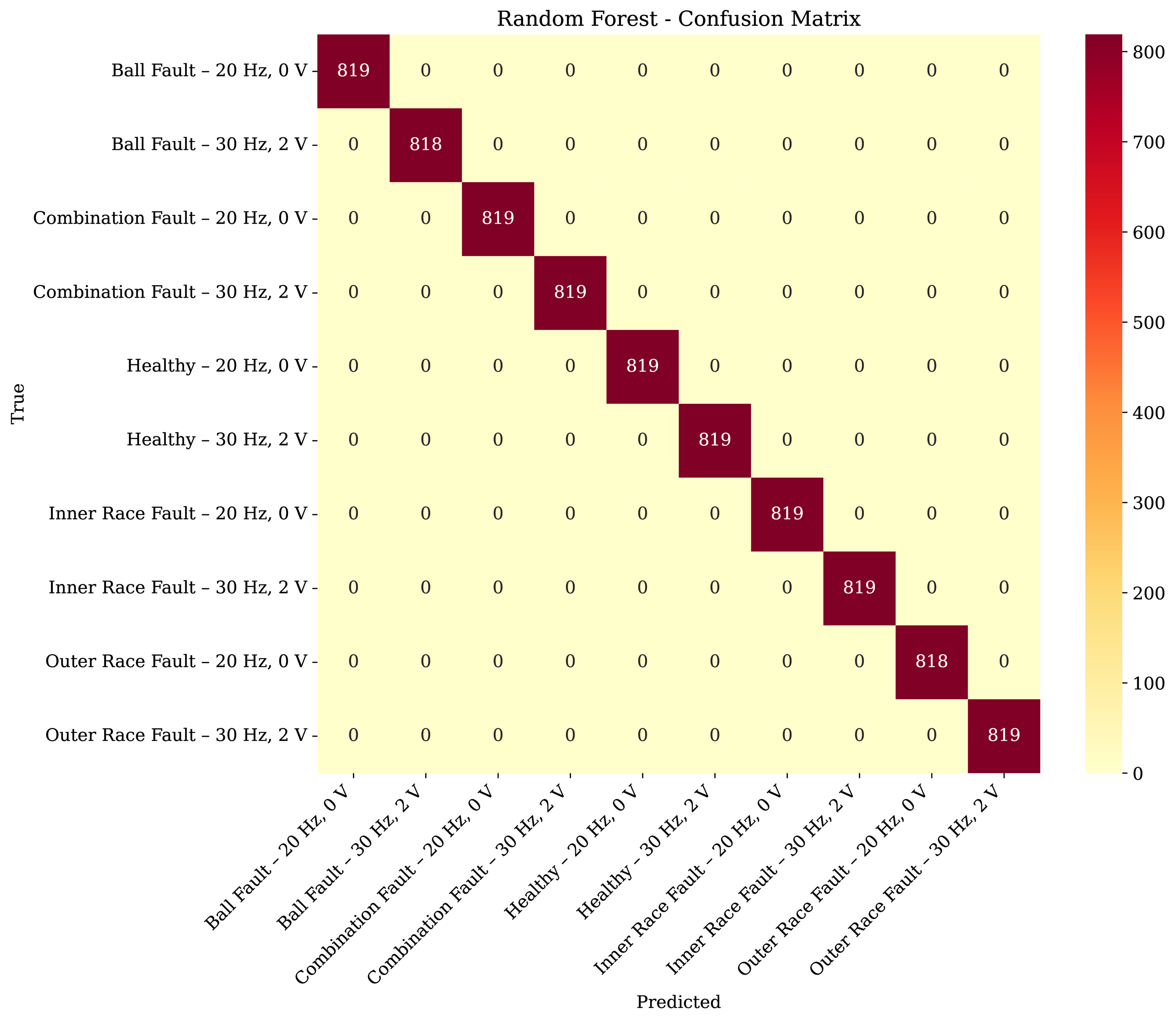} 
	\caption{Confusion matrices of the Random Forest classifier applied to the SU Bearing dataset.}
	\label{fig:classificationReport_bearing}
\end{figure}

\begin{figure}[h!]
	\centering
	\includegraphics[width=0.6\textwidth]{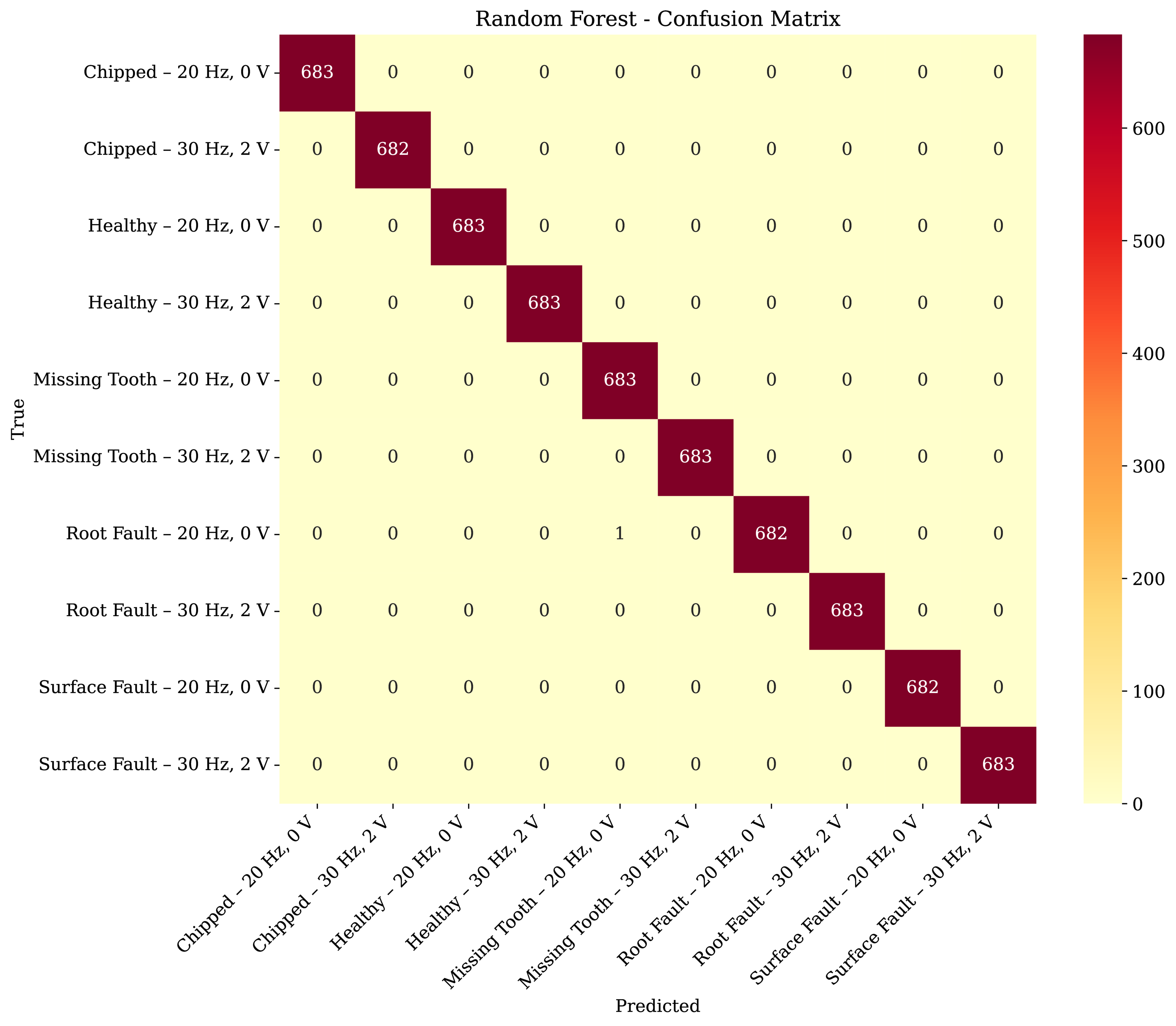} 
	\caption{Confusion matrices of the Random Forest classifier applied to the SU Gearbox dataset}
	\label{fig:classificationReport_gearbox}
\end{figure}

All models exhibited exceptional performance on the SU dataset, which includes the Gearbox and Bearing subsets, reflecting the dataset characteristics, such as well-separated fault classes or less noisy data. For Gearbox, all three models, SVM, Logistic Regression, and Random Forest, achieved perfect accuracy (100.0\%) and CV means accuracy (100.0\%, ±0.000), indicating robust classification across gearbox-related faults. Logistic Regression stands out with the fastest inference time (0.000002 s), followed closely by Random Forest (0.000010 s), whereas SVM is slower (0.001165 s). For Bearing, Logistic Regression and Random Forest again achieved perfect accuracy (100.0\%) and CV mean accuracy (100.0\%, ±0.000), with Logistic Regression maintaining its speed advantage (0.000002 s) over Random Forest (0.000021 s) and SVM (0.001074 s). SVM, while nearly perfect (99.9\% accuracy and CV mean accuracy), was slightly less accurate on Bearing-SU and consistently slower across both SU subsets.

Random Forest is the most robust and accurate model across the CWRU and SU datasets, making it a top choice for fault diagnosis. On the CWRU dataset, RF consistently achieved the highest accuracy (99.7-100.0\%) and cross-validation (CV) mean accuracy (99.8-100.0\%, ±0.000-0.2), with perfect classification at 0 and 3 Hp. This demonstrates its exceptional ability to handle complex fault patterns and varying load conditions, ensuring reliable performance in diverse operational scenarios. Similarly, on the SU dataset, RF delivered perfect accuracy (100.0\%) and CV mean accuracy (100.0\%, ±0.000) for both Gearbox-SU and Bearing-SU, showcasing its capability to excel in datasets with well-separated fault classes. Although its inference times are not the fastest, they remain highly competitive (0.0092-0.0325 s for CWRU, 0.000010–0.000021 s for SU), striking an effective balance between accuracy and computational efficiency.

In contrast, Logistic Regression offers impressive speed, with inference times as low as 0.0007-0.0167 s on CWRU and 0.000002 s on SU, alongside strong accuracy (98.7-100.0\% on CWRU, 100.0\% on SU). However, its accuracy occasionally falls short of RF, particularly at lower horsepower loads on the CWRU (e.g., 98.7\% at 0 Hp), and its performance is less consistent across varying conditions. SVM, while reliable with accuracies of 98.7-99.8\% on CWRU and 99.9-100.0\% on SU, is significantly hindered by slower inference times (0.0274-1.3235 s on CWRU, 0.001074-0.001165 s on SU), making it less practical for real-time applications. Additionally, SVM’s slightly lower accuracy of SVM on CWRU (e.g., 98.7\% at 0 Hp) and Bearing-SU (99.9\%) underscores the advantage of RF in precision.

\begin{table}[h!]
	\centering
	
	\caption{Cross-load transfer performance of Random Forest (RF) classifier across different source and target load conditions on the CWRU dataset, and selected scenarios from the SU dataset. Metrics include Accuracy, F1-score, Precision, and Recall.}
	\label{tab:rf_cross_load_transfer}
	\scriptsize % reduces font size for better fit
%	\resizebox{\columnwidth}{!}{%
		\begin{tabular}{cccccc}
			\toprule
			\textbf{Source Load} & \textbf{Target Load} & \textbf{Accuracy} & \textbf{F1-score} & \textbf{Precision} & \textbf{Recall} \\
			\midrule
			\multicolumn{6}{c}{\textit{CWRU Cross-load Transfer}} \\
			\midrule
			0 Hp & 0 Hp & 1.000000 & 1.000000 & 1.000000 & 1.000000 \\
			& 1 Hp & 0.967655 & 0.967440 & 0.968459 & 0.967655 \\
			& 2 Hp & 0.941808 & 0.941617 & 0.942108 & 0.941808 \\
			& 3 Hp & 0.963277 & 0.962806 & 0.962601 & 0.963277 \\
			\midrule
			1 Hp & 0 Hp & 0.995614 & 0.995614 & 0.995805 & 0.995614 \\
			& 1 Hp & 0.999102 & 0.999101 & 0.999109 & 0.999102 \\
			& 2 Hp & 0.995480 & 0.995475 & 0.995492 & 0.995480 \\
			& 3 Hp & 0.988701 & 0.988704 & 0.988869 & 0.988701 \\
			\midrule
			2 Hp & 0 Hp & 0.986842 & 0.986803 & 0.988421 & 0.986842 \\
			& 1 Hp & 0.997305 & 0.997309 & 0.997320 & 0.997305 \\
			& 2 Hp & 1.000000 & 1.000000 & 1.000000 & 1.000000 \\
			& 3 Hp & 0.997175 & 0.997178 & 0.997188 & 0.997175 \\
			\midrule
			3 Hp & 0 Hp & 0.982456 & 0.982375 & 0.984157 & 0.982456 \\
			& 1 Hp & 0.982929 & 0.982913 & 0.982951 & 0.982929 \\
			& 2 Hp & 0.981921 & 0.981913 & 0.982003 & 0.981921 \\
			& 3 Hp & 0.999435 & 0.999435 & 0.999438 & 0.999435 \\
			\midrule
			\multicolumn{6}{c}{\textit{SU Cross-load Transfer}} \\
			\midrule
			Bearing & Bearing & 1.000000 & 1.000000 & 1.000000 & 1.000000 \\
			& Gearbox & 0.973000 & 0.911000 & 0.953100 & 0.893000 \\
			\midrule
			Gearbox & Bearing & 0.996000 & 0.857000 & 0.866000 & 0.834000 \\
			& Gearbox & 1.000000 & 1.000000 & 1.000000 & 0.990000 \\
			\bottomrule
		\end{tabular}%
	%}
\end{table}

To evaluate the Random Forest (RF) model's effectiveness, we visualized its classification performance using confusion matrices for the CWRU and SU datasets. Figure~\ref{fig:heatmapcwru} shows RF's performance on the CWRU dataset across four motor loads (0--3 Hp) and ten bearing conditions (e.g., Normal, Ball Fault, Inner Race Fault). Figures~\ref{fig:classificationReport_bearing} and \ref{fig:classificationReport_gearbox} highlight RF's accuracy in identifying fault classes under varying conditions for SU's bearing and gearbox systems.

\subsubsection{Cross Learning}

Real-world fault detection often involves diverse operating conditions, where models trained on specific loads may struggle to generalize. Transfer learning is thus critical for effective cross-load fault diagnosis.
	
	\begin{table}[h!]
	\centering
	\caption{Cross-Fault Transfer Results: Classification performance for models trained on fault size \SI{0.007}{\inch} and tested on \SI{0.007}{\inch} and \SI{0.021}{\inch} (CWRU dataset).}
	\scriptsize % reduces font size for better fit
	\label{tab:cwru_cross_fault_transfer}
	\begin{tabular}{ccccccc}
		\toprule
		\textbf{Source} & \textbf{Target} & \textbf{Model} & \textbf{Accuracy} & \textbf{F1-Score} & \textbf{Precision} & \textbf{Recall} \\
		\midrule
		\multicolumn{6}{c}{\textit{CWRU Cross-Fault Transfer}} \\
		\midrule
		\multirow{3}{*}{\SI{0.007}{\inch}} & \multirow{3}{*}{\SI{0.007}{\inch}} 
		& LR  & 1.0000 & 1.0000 & 1.0000 & 1.0000 \\
		&     & RF  & 1.0000 & 1.0000 & 1.0000 & 1.0000 \\
		&     & SVM & 0.6667 & 0.5556 & 0.5000 & 0.6667 \\
		\midrule
		\multirow{3}{*}{\SI{0.007}{\inch}} & \multirow{3}{*}{\SI{0.021}{\inch}} 
		& LR  & 0.1940 & 0.2446 & 0.3422 & 0.1940 \\
		&     & RF  & \textbf{0.9718} & \textbf{0.9727} & \textbf{0.9738} & \textbf{0.9718} \\
		&     & SVM & 0.3446 & 0.1909 & 0.3137 & 0.3446 \\
		\bottomrule
	\end{tabular}
\end{table}
	\begin{figure}[h!]
	\centering
	\includegraphics[width=0.6\textwidth]{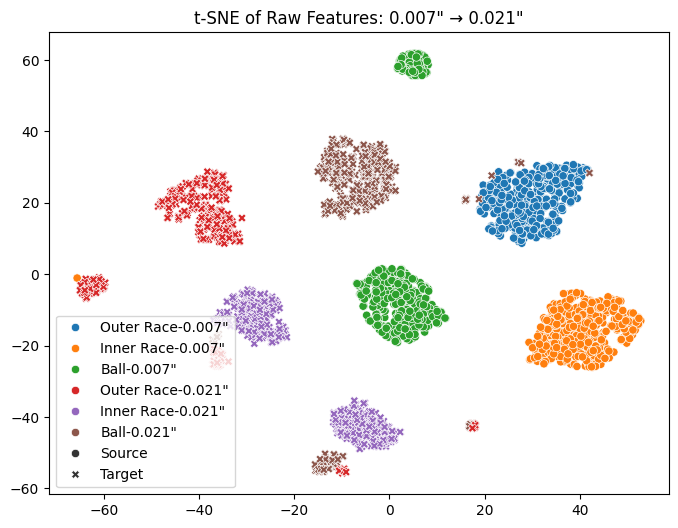}
	\caption{t-SNE visualization of raw features for the cross-fault transfer scenario from fault size \SI{0.007}{\inch} (source) to \SI{0.021}{\inch} (target) on the CWRU dataset. Fault types and sizes are represented with distinct colors; markers distinguish source ($\cdot$) and target ($\times$) domains.}
	\label{fig:cwsu_tsne}
\end{figure}

In cross-load scenarios, RF demonstrates robust performance on the CWRU dataset, achieving perfect classification (accuracy, F1-score, precision, recall = 1.0) when trained and tested on the same load (e.g., 0 Hp $\to$ 0 Hp). Even when tested on different loads (e.g., 0 Hp $\to$ 1 Hp), RF maintains high performance, with accuracy and F1-scores above 94\% (Table~\ref{tab:rf_cross_load_transfer}). For SU's cross-domain transfers (e.g., bearing $\to$ gearbox), RF achieves up to 99.6\% accuracy, with slight F1-score reductions due to domain shifts, underscoring its adaptability across diverse conditions.
	
	The RF model also excels in cross-fault scenarios. On the CWRU dataset, when trained on \SI{0.007}{\inch} faults and tested on \SI{0.021}{\inch} faults, RF achieves 97.18\% accuracy and 97.27\% F1-score, significantly outperforming Logistic Regression (LR, ~19\%) and Support Vector Machine (SVM, ~34\%) (Table~\ref{tab:cwru_cross_fault_transfer}). Similarly, on the SU dataset, RF delivers 99.4\% accuracy for bearing-to-gearbox transfers and 96.2\% for gearbox-to-bearing transfers, surpassing LR ($\leq$82.3\%) and SVM ($\leq$90.1\%) (Table~\ref{tab:SU_cross_fault_transfer}).

	\begin{table}[h!]
		\centering
		\caption{Cross-Fault Load Classification Performance (SU dataset).}
			\scriptsize % reduces font size for better fit
		\label{tab:SU_cross_fault_transfer}
		\begin{tabular}{lllS[table-format=1.3]S[table-format=1.3]}
			\toprule
			\textbf{Method} & \textbf{Train Domain} & \textbf{Test Domain} & \textbf{Accuracy} & \textbf{F1-Score} \\
			\midrule
			RF & Bearing (20Hz, 0V) & Gearbox (20Hz, 0V) & 0.994 & 0.992 \\
			& Gearbox (20Hz, 0V) & Bearing (20Hz, 0V) & 0.962 & 0.955 \\
			\midrule
			SVM & Bearing (20Hz, 0V) & Gearbox (20Hz, 0V) & 0.901 & 0.847 \\
			& Gearbox (20Hz, 0V) & Bearing (20Hz, 0V) & 0.000 & 0.000 \\
			\midrule
			LR & Bearing (20Hz, 0V) & Gearbox (20Hz, 0V) & 0.823 & 0.784 \\
			& Gearbox (20Hz, 0V) & Bearing (20Hz, 0V) & 0.019 & 0.012 \\
			\bottomrule
		\end{tabular}
	\end{table}
	
	\begin{figure}[h!]
		\centering
		\includegraphics[width=\linewidth]{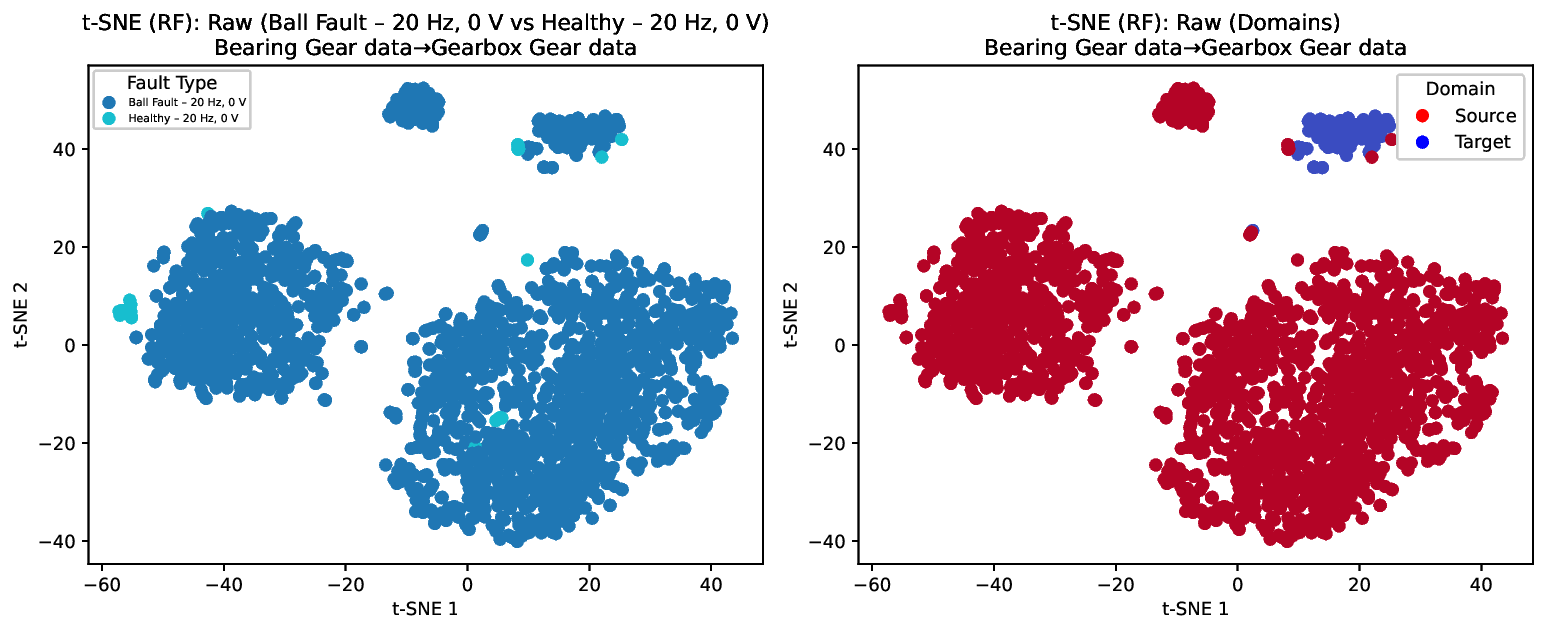}
		\caption{t-SNE visualization of Random Forest features for (A) Fault-type separation (Ball Fault vs. Healthy) and (B) Domain alignment (Bearing$\to$Gearbox).}
		\label{fig:SU_tsne}
	\end{figure}
	
	t-SNE visualizations further illustrate RF’s strengths (Figures~\ref{fig:cwsu_tsne}, \ref{fig:SU_tsne}). These plots show RF’s ability to separate fault types across different sizes and align source and target domains, reflecting its robust feature learning. Unlike LR’s linear constraints or SVM’s sensitivity to margin-based boundaries, RF’s ensemble approach effectively captures complex, nonlinear patterns. Collectively, RF’s high accuracy, robust generalization across loads and fault sizes, and effective domain adaptation (Tables~\ref{tab:rf_cross_load_transfer}, \ref{tab:cwru_cross_fault_transfer}, \ref{tab:SU_cross_fault_transfer}; Figures~\ref{fig:cwsu_tsne}, \ref{fig:SU_tsne}) make it an ideal choice for real-world fault diagnosis under varying conditions.

\subsubsection{Feature Importance Analysis}

To evaluate the Random Forest (RF) model’s predictive behavior under varying loads, SHAP (SHapley Additive exPlanations) analysis quantifies feature contributions in the CWRU and SU datasets. The CWRU dataset includes vibration signals from two channels (Ch1, Ch2), while the SU dataset comprises eight channels (Ch1–Ch8).

\begin{table}[h!]
	\centering
	%\scriptsize
	\caption{Summary of SHAP-Based Feature Importance Across Load Conditions for CWRU dataset}
	\label{tab:cwrushap_summary}
		\scriptsize % reduces font size for better fit
	\begin{tabular}{clccccl}
		\toprule
		\textbf{Rank} & \textbf{Feature} & \textbf{Domain} & \textbf{Ch.} & \textbf{0 Hp} & \textbf{3 Hp} & \textbf{Trend} \\
		\midrule
		1  & Mean Ch1               & Time             & Ch1    & Very High  & Low       & ↓ Decreasing \\
		2  & PSD Low Freq Ch1       & Frequency        & Ch1    & Very High  & Very High & ↑ Then Stable \\
		3  & FFT Low Freq Ch2       & Frequency        & Ch2    & High       & Medium    & Stable / ↓ \\
		4  & Kurtosis Ch1/Ch2       & Time             & Ch1/2  & High       & Low       & ↓ Decreasing \\
		5  & PSD Mid Freq Ch2       & Frequency        & Ch2    & Medium     & Very High & ↑ Increasing \\
		6  & PSD High Freq Ch2      & Frequency        & Ch2    & Low        & Very High & ↑ Increasing \\
		7  & TKE Ch1                & Time-Frequency   & Ch1    & Low        & Medium    & ↑ Then Stable \\
		8  & RMS Ch1                & Time             & Ch1    & Medium     & Low       & ↓ Slightly \\
		9  & FFT Std Ch1            & Frequency        & Ch1    & Low        & Medium    & Stable \\
		10 & Env FFT Mean Ch2       & Envelope Freq    & Ch2    & Medium     & High      & ↑ Increasing \\
		11 & FFT Mean Ch2           & Frequency        & Ch2    & Low        & Medium    & Stable \\
		12 & PSD Peak Ch1/Ch2       & Frequency        & Ch1/2  & Low        & Medium    & Stable \\
		13 & Median Ch2             & Time             & Ch2    & High       & Low       & ↓ Disappears \\
		\bottomrule
	\end{tabular}
\end{table}

\begin{figure}[h!]
	\centering
	\begin{subfigure}[t]{0.48\textwidth}
		\centering
		\includegraphics[width=\linewidth]{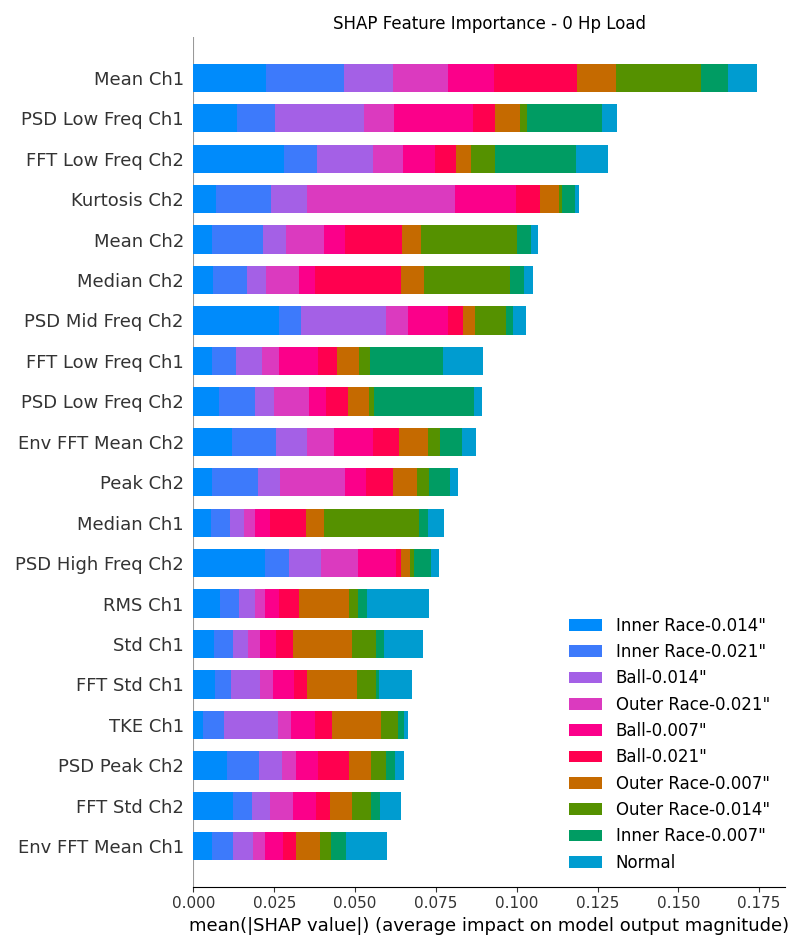} % Replace with actual filename
		\caption{SHAP feature importance at 0 HP (no load).}
		\label{fig:shap0hp}
	\end{subfigure}
	\hfill
	\begin{subfigure}[t]{0.48\textwidth}
		\centering
		\includegraphics[width=\linewidth]{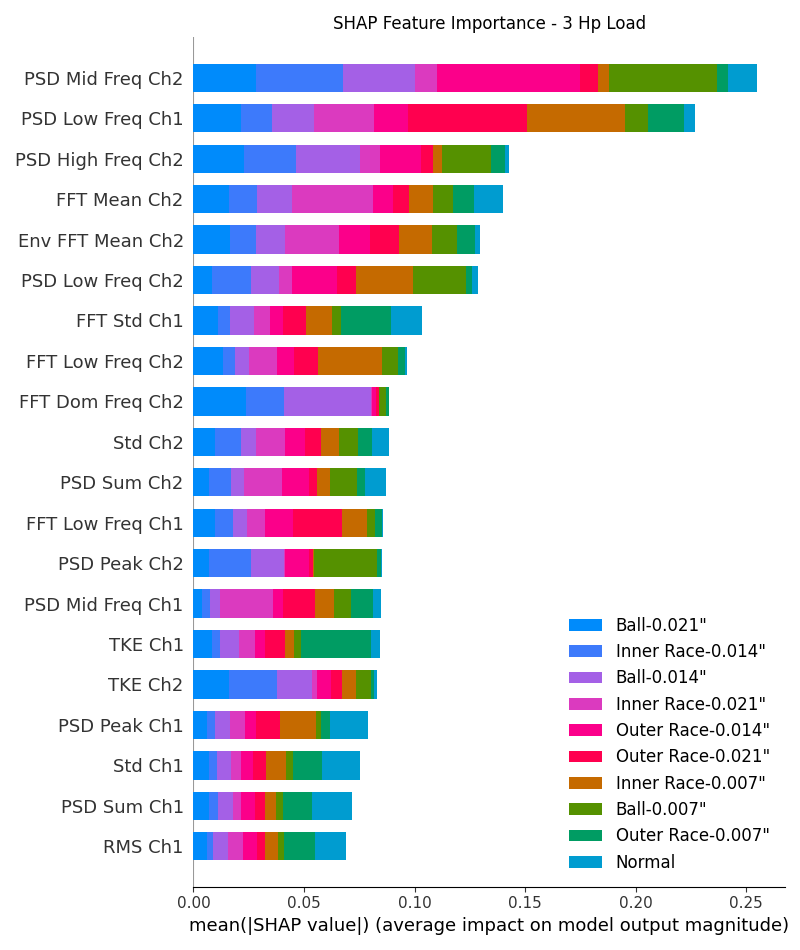} % Replace with actual filename
		\caption{SHAP feature importance at 3 HP (full load).}
		\label{fig:shap3hp}
	\end{subfigure}
	\begin{subfigure}[b]{0.48\textwidth}
		\centering
		\includegraphics[width=\textwidth]{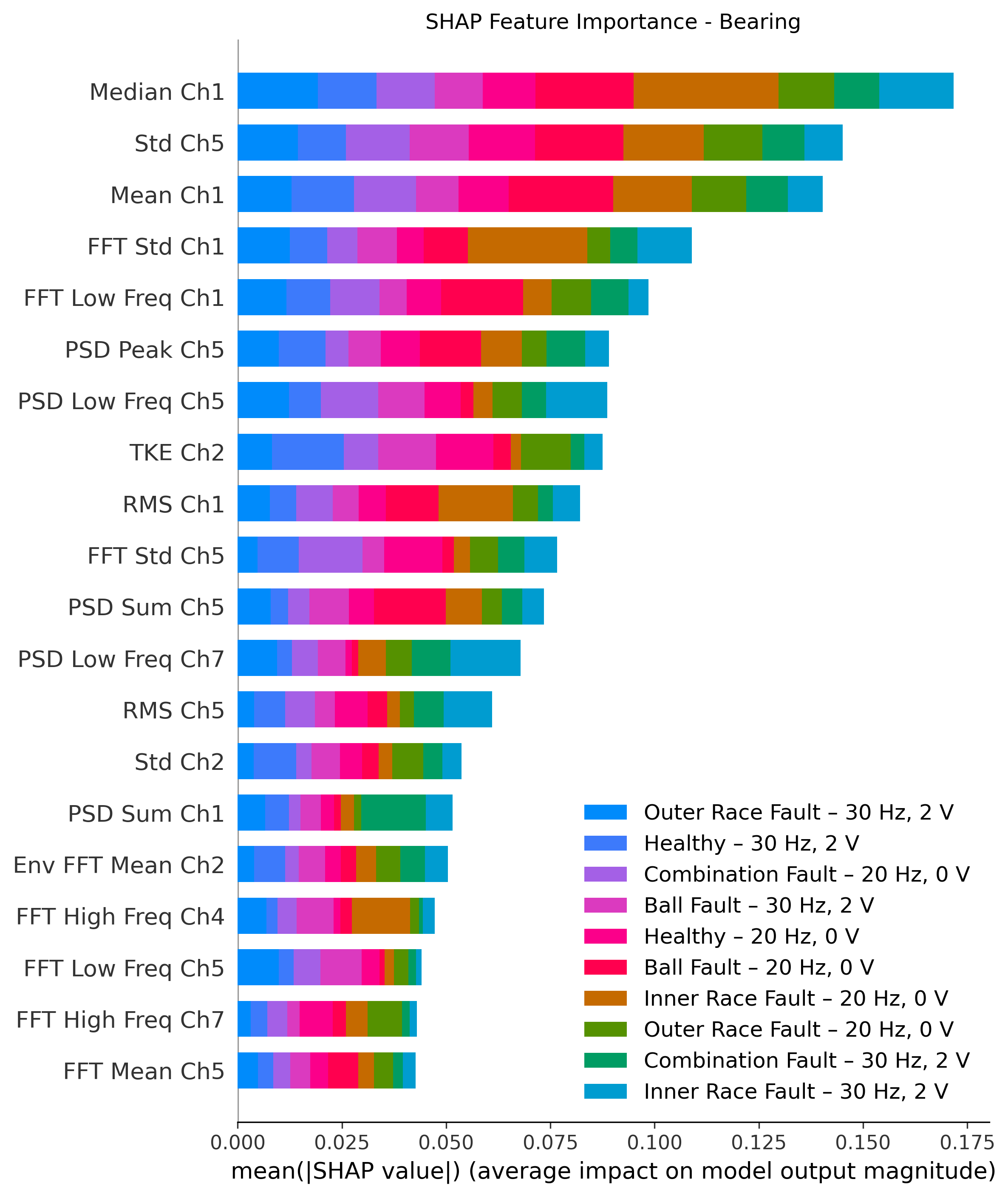}
		\caption{SHAP Feature Importance – Bearing.}
		\label{fig:shap_bearing}
	\end{subfigure}
	\hfill
	\begin{subfigure}[b]{0.48\textwidth}
		\centering
		\includegraphics[width=\textwidth]{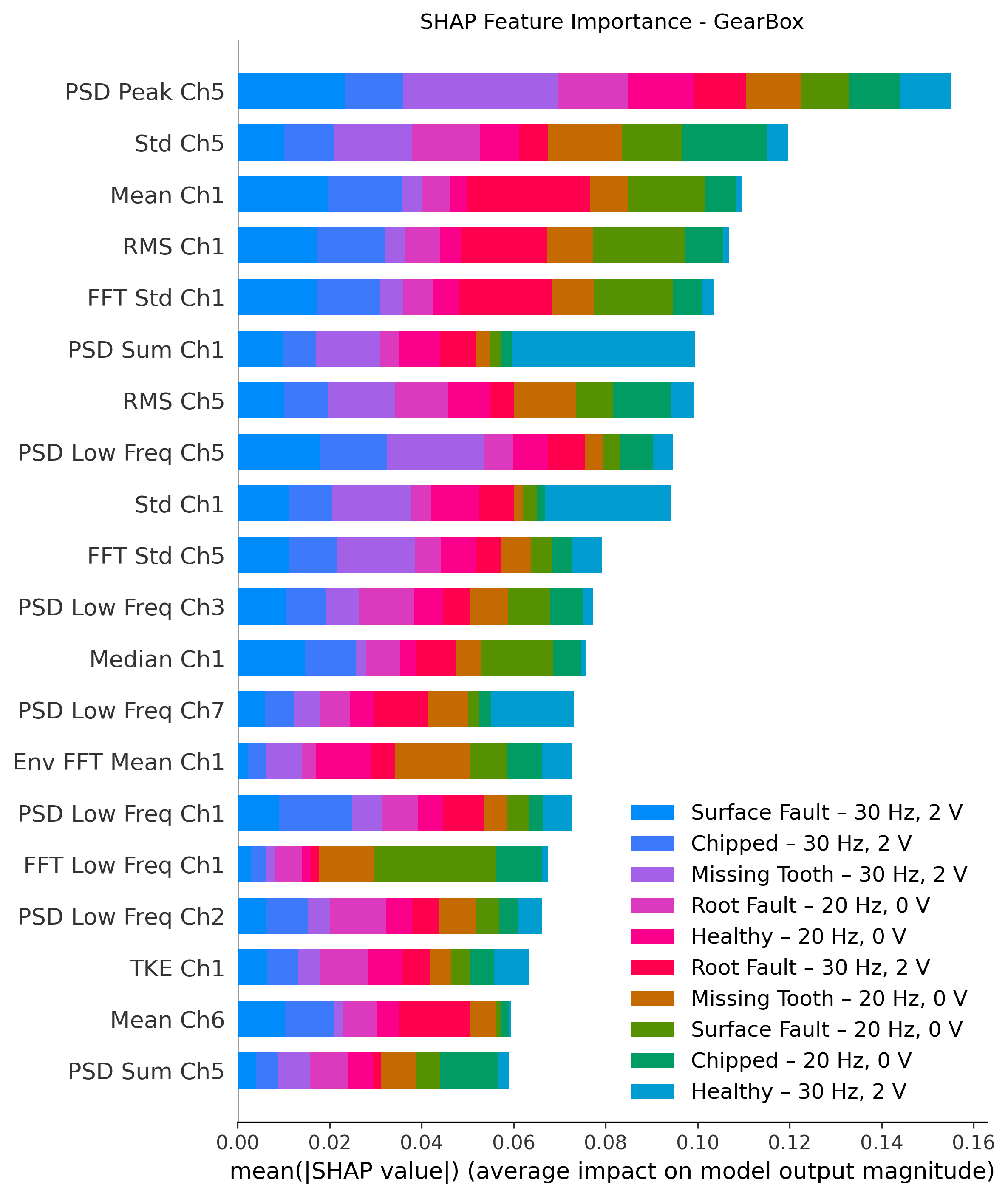}
		\caption{SHAP Feature Importance – Gearbox.}
		\label{fig:shap_gearbox}
	\end{subfigure}
	\caption{Comparison of SHAP feature importance across CWRU and SU Datasets.}
	\label{fig:shap_comparison}
\end{figure}

\begin{table}[h!]
	\centering
	%\scriptsize
	\caption{Comparison of SHAP-Based Feature Importance Between SU Bearing and Gearbox Fault Classification}
	\label{tab:SU_shap_comparison_summary}
		\scriptsize % reduces font size for better fit
	\begin{tabular}{cccccp{3 cm}}
		\toprule
		\textbf{Rank} & \textbf{Feature} & \textbf{Domain}  & \textbf{Bearing Impact} & \textbf{Gearbox Impact} & \textbf{Trend} \\
		\midrule
		1  & Median Ch1           & Time           & Very High    & Medium       & ↓ Less Important in Gearbox \\
		2  & PSD Peak Ch5         & Frequency       & High         & Very High    & ↑ More Dominant in Gearbox \\
		3  & Std Ch5              & Time            & Very High    & Very High    & → Stable Across Both \\
		4  & Mean Ch1             & Time            & Very High    & High         & ↓ Slight Drop in Gearbox \\
		5  & FFT Std Ch1          & Frequency       & High         & Medium       & ↓ Lower in Gearbox \\
		6  & RMS Ch1              & Time            & Medium       & High         & ↑ Higher in Gearbox \\
		7  & PSD Sum Ch1          & Frequency       & Medium       & Medium       & → Similar Impact \\
		8  & RMS Ch5              & Time            & Medium       & Medium       & → Stable \\
		9  & PSD Low Freq Ch5     & Frequency       & High         & Medium       & ↓ Slight Drop \\
		10 & FFT Low Freq Ch1     & Frequency       & High         & Medium       & ↓ Slight Drop \\
		11 & FFT Std Ch5          & Frequency       & Medium       & Medium       & → Similar Impact \\
		12 & TKE Ch2              & Time-Frequency  & Medium       & Low          & ↓ Less in Gearbox \\
		13 & Env FFT Mean Ch2     & Envelope Freq   & Medium       & Low          & ↓ Declining \\
		14 & PSD Low Freq Ch3     & Frequency       & Low          & Medium       & ↑ Only Important in Gearbox \\
		15 & FFT High Freq Ch4    & Frequency       & Medium       & Low          & ↓ Less Important in Gearbox \\
		\bottomrule
	\end{tabular}
\end{table}

The SHAP analysis results are summarized in Table \ref{tab:cwrushap_summary} and detailed through visualizations in Figures \ref{fig:shap0hp} and \ref{fig:shap3hp}, which illustrate the magnitude and direction of feature influences under the two load extremes. Under no-load conditions (0 HP), time-domain features like Mean Ch1 and Kurtosis Ch1/Ch2 strongly contribute to model predictions, reflecting their sensitivity to low-stress scenarios. However, their influence diminishes significantly under full-load conditions (3 HP). In contrast, frequency-domain features, such as PSD Mid Frequency Ch2, PSD High Frequency Ch2, and Envelope FFT Mean Ch2, become more prominent at full load, underscoring their importance in high-stress environments. Features like FFT Standard Deviation Ch1 and PSD Peak Ch1/Ch2 maintain relatively stable contributions across both load levels, indicating robustness to changing operational conditions.

Table\ref{tab:SU_shap_comparison_summary} and Figures\ref{fig:shap_bearing} and~\ref{fig:shap_gearbox} compare SHAP feature importance for bearing and gearbox fault classification in the SU dataset. For bearings, time-domain features (Median Ch1, Std Ch5, Mean Ch1) dominate, capturing amplitude variations from localized faults (e.g., inner race defects). For gearboxes, frequency-domain features (PSD Peak Ch5, FFT Std Ch1) are critical, detecting spectral patterns linked to gear faults (e.g., root cracks). Features like Std Ch5 and RMS Ch1 maintain high importance across both, while TKE Ch2 and Env FFT Mean Ch2 are bearing-specific. Thus, time-domain features drive bearing diagnostics, while frequency-domain features are key for gearboxes.

\subsubsection{Noise Sensitivity Analysis}

\begin{figure}[h!]
	\centering
	\begin{subfigure}{0.45\textwidth}
		\includegraphics[width=\linewidth]{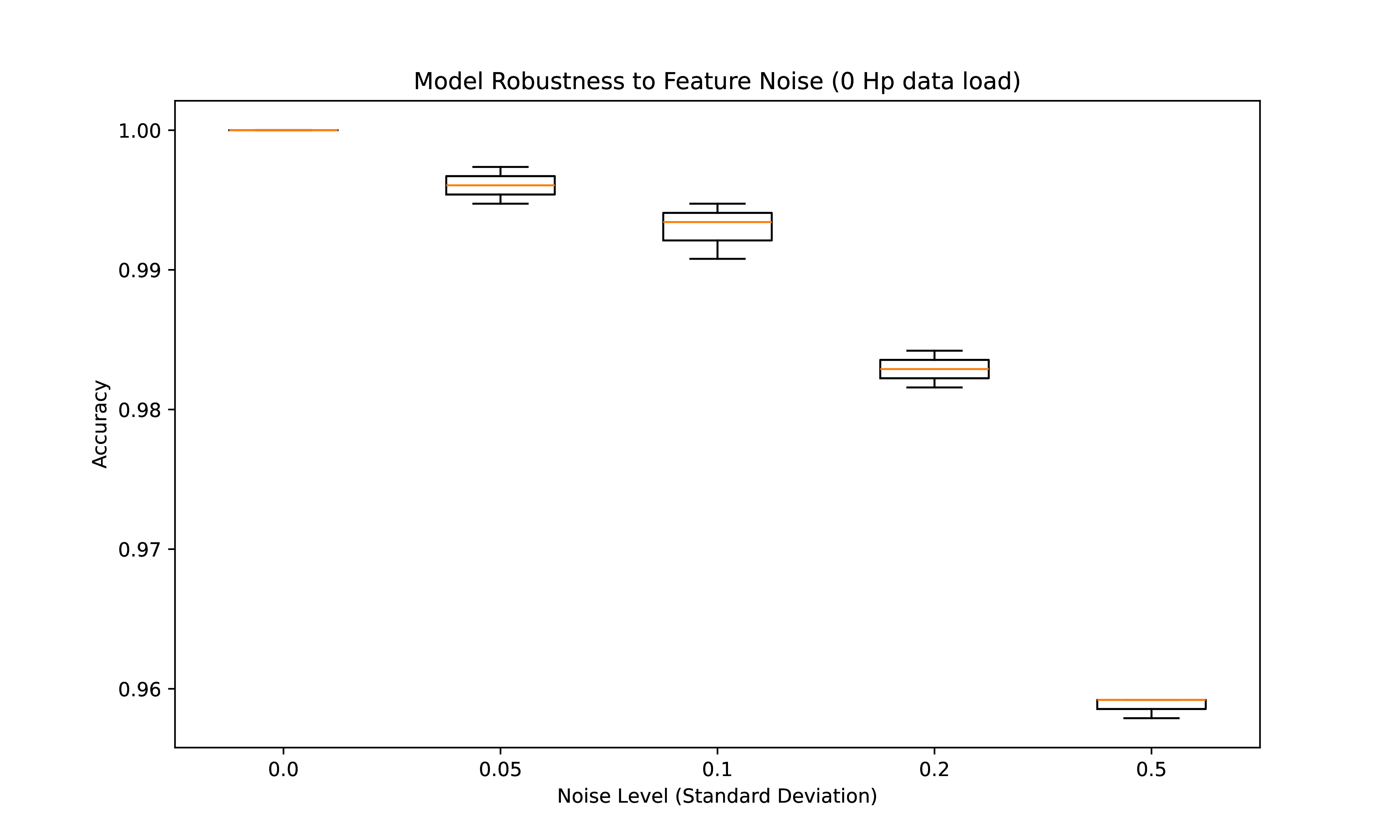}
		\caption{Accuracy with feature noise (0 Hp data load)}
		\label{fig:Noise0hp}
	\end{subfigure}
	\vspace{0.3cm}
	\begin{subfigure}{0.45\textwidth}
		\includegraphics[width=\linewidth]{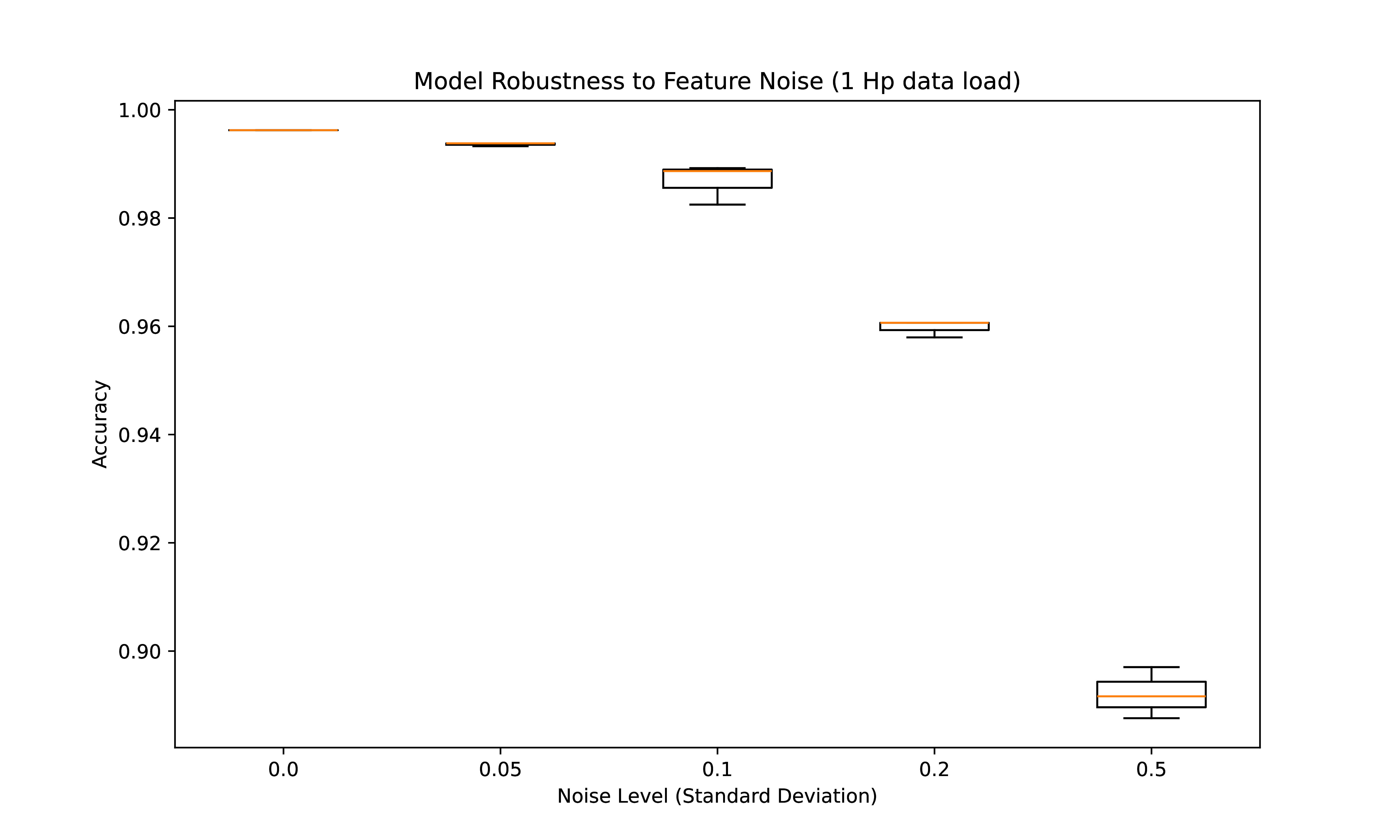}
		\caption{Accuracy with feature noise (1 Hp data load)}
		\label{fig:Noise1hp}
	\end{subfigure}
	\begin{subfigure}{0.45\textwidth}
		\includegraphics[width=\linewidth]{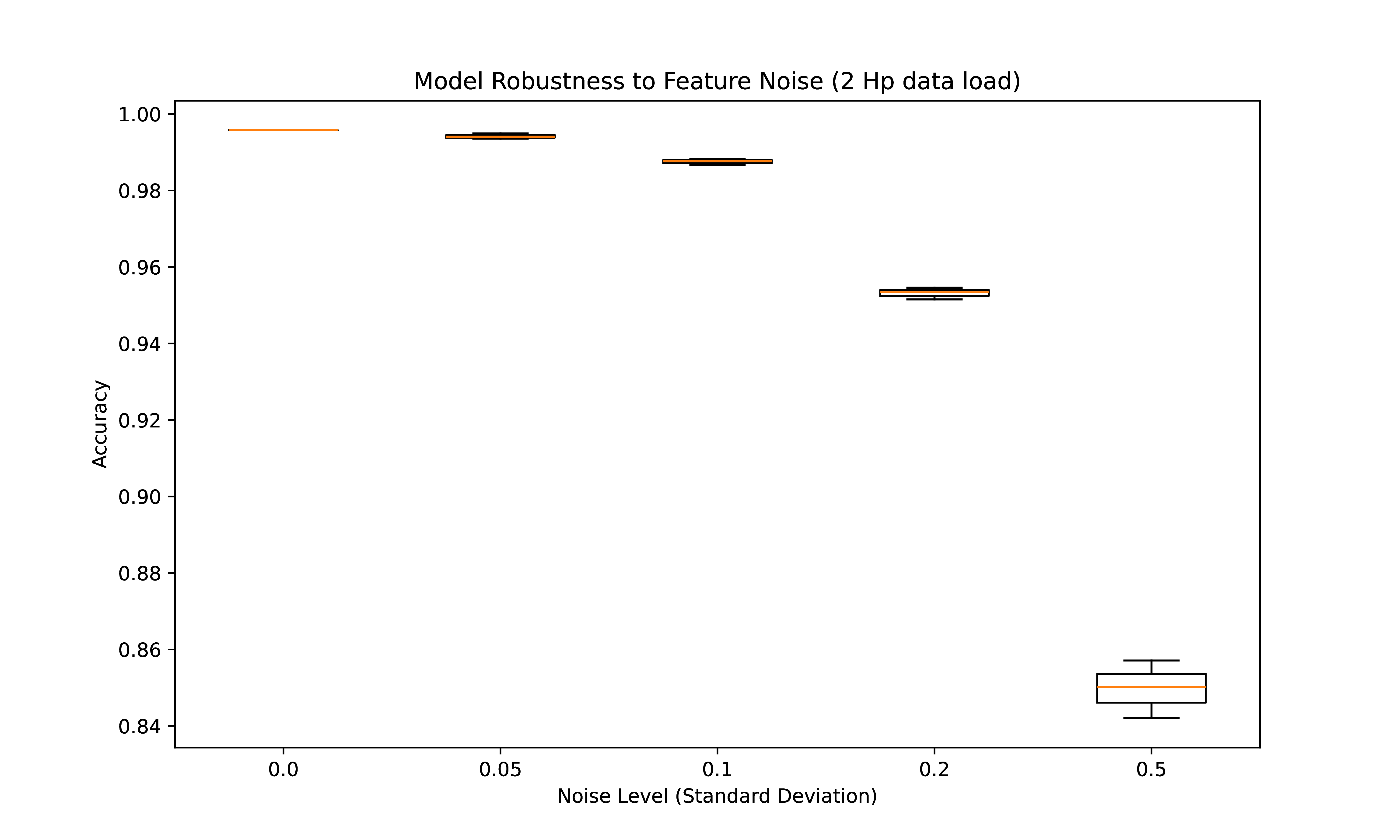}
		\caption{Accuracy with feature noise (2 Hp data load)}
		\label{fig:Noise2hp}
	\end{subfigure}
	\begin{subfigure}{0.45\textwidth}
		\includegraphics[width=\linewidth]{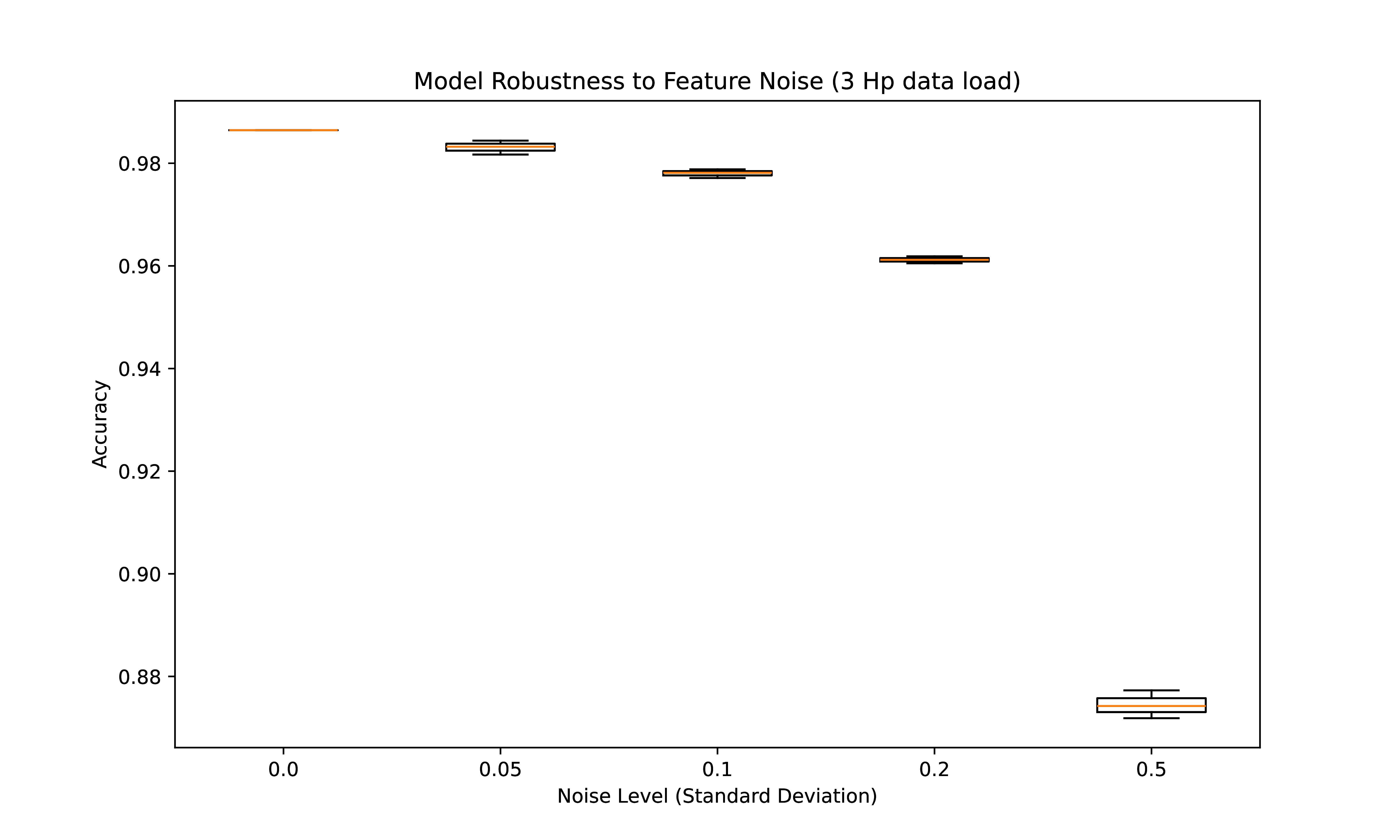}
		\caption{Accuracy with feature noise (1 Hp data load)}
		\label{fig:Noise3hp}
	\end{subfigure}
	\vspace{0.3cm}
	\begin{subfigure}{0.45\textwidth}
		\includegraphics[width=\linewidth]{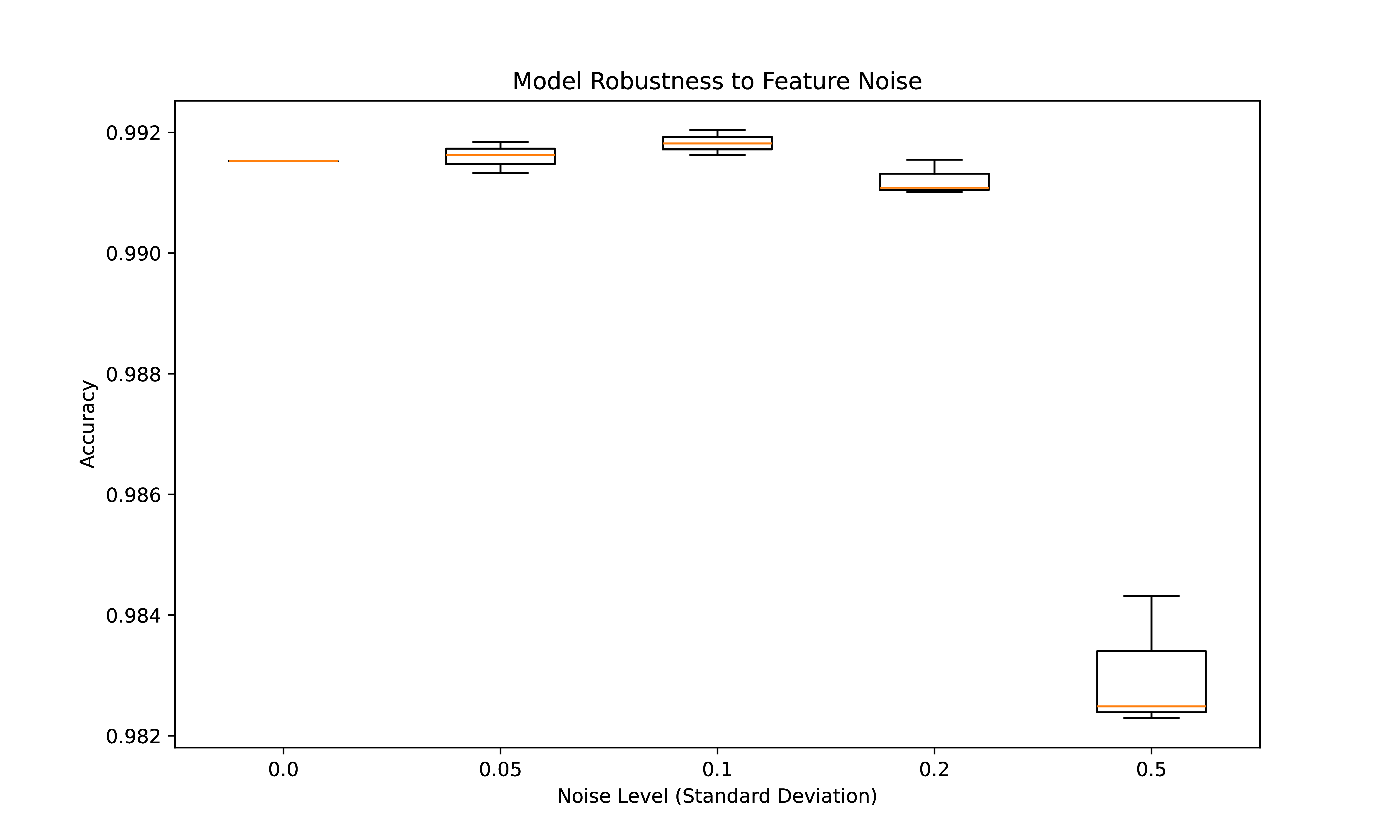}
		\caption{Accuracy with feature noise (SU Gearbox)}
		\label{fig:NoiseGear}
	\end{subfigure}
	\begin{subfigure}{0.45\textwidth}
		\includegraphics[width=\linewidth]{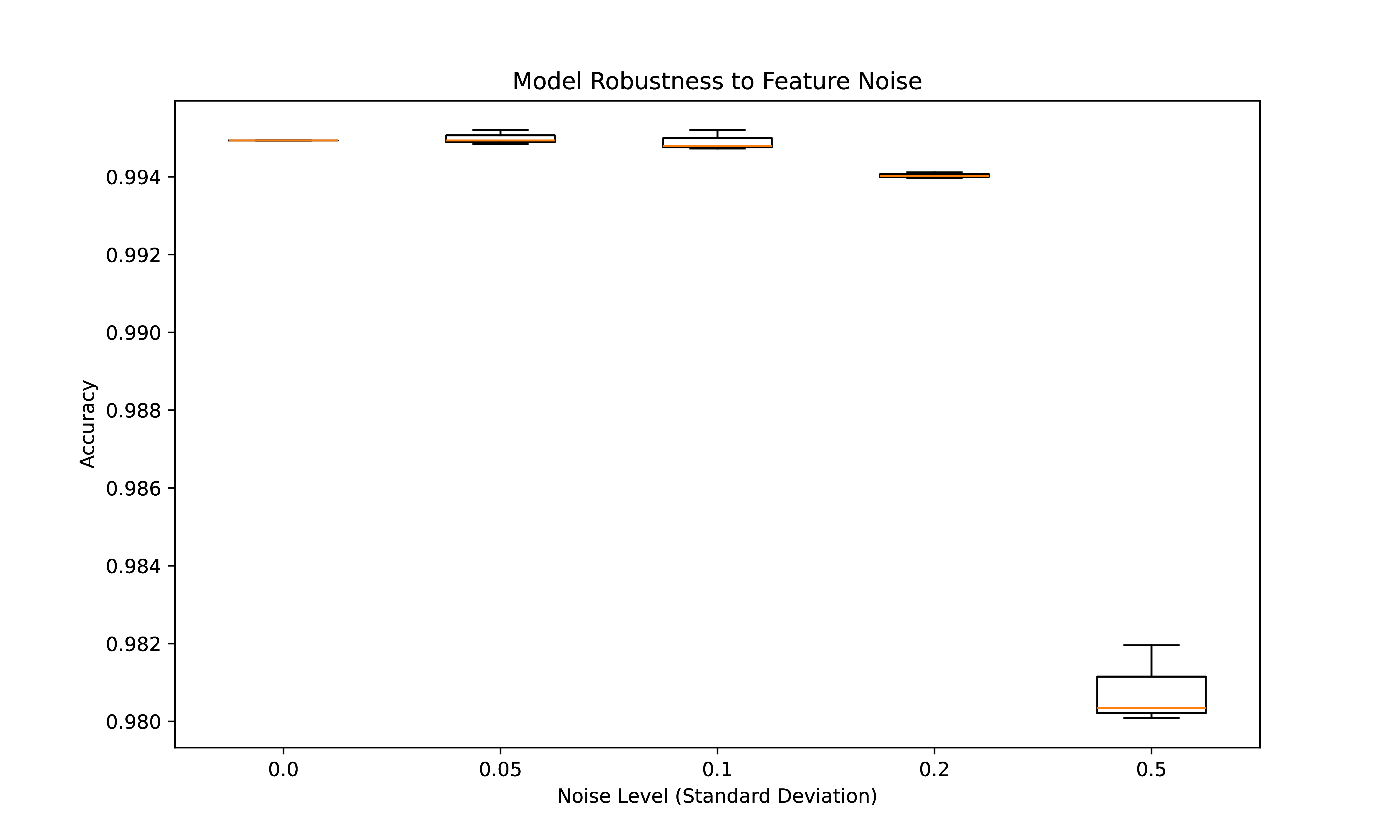}
		\caption{Accuracy with feature noise (SU Bearing)}
		\label{fig:NoiseBearing}
	\end{subfigure}
	
	\caption{Classification accuracy of Random Forest under varying levels of Gaussian noise across datasets.}
	\label{fig:rf_noise_plot}
\end{figure}

Random Forest demonstrates robust performance against noisy inputs due to its ensemble structure and randomized feature selection. This study evaluates RF’s resilience under Gaussian noise with zero-mean and standard deviations of $\sigma = 0, 0.05, 0.1, 0.2, 0.5$ applied to input features of the CWRU bearing dataset (0--3 HP loads) and SU gearbox/bearing dataset. Classification accuracy, assessed via stratified 5-fold cross-validation, is reported in Figure~\ref{fig:rf_noise_plot}. 

For the SU dataset (Figure \ref{fig:NoiseGear}, and \ref{fig:NoiseBearing}), RF maintains high accuracy (0.994--0.980 and 0.992--0.982, respectively) across all noise levels, with a minimal decline of 1.2--1.4\% at $\sigma = 0.5$. 

For the CWRU dataset at 1--3 HP (Figures~\ref{fig:Noise1hp} to~\ref{fig:Noise3hp}), accuracy remains robust at 0.99--0.98 for $\sigma \leq 0.1$, but drops to 0.88--0.84 at $\sigma = 0.5$, indicating higher noise sensitivity in higher-load conditions. In contrast, at 0 HP (Figure~\ref{fig:Noise0hp}), accuracy is more stable, declining from 0.98 at $\sigma = 0.05$ to 0.96 at $\sigma = 0.5$, suggesting lower noise sensitivity in low-load conditions. This robustness, consistent with Breiman~\citep{breiman2001random}, stems from RF’s ensemble averaging, where decision trees trained on bootstrapped samples and random feature subsets mitigate the impact of noisy features. The increased noise sensitivity at 1--3 HP may reflect greater signal complexity in higher-load vibration data. 
 
These findings affirm RF’s suitability for fault diagnosis in low-to-moderate noise industrial settings, particularly for gearbox and low-load bearing applications. However, preprocessing or feature engineering may be required for higher-load scenarios under high noise ($\sigma \geq 0.5$).

\subsection{Comparative Analysis}

The proposed bearing fault detection model consistently outperforms conventional deep learning approaches (e.g., CNN, LSTM) as well as advanced graph-based techniques (e.g., GCN, GAT) on both the CWRU and SU datasets. By integrating multiscale feature engineering, signal processing, and graph-theoretic metrics—such as path length ($L$), modularity ($Q$), and spectral gap—within the random forest framework, the model achieves a compelling balance of accuracy, computational efficiency, and interpretability. Performance benchmarks in Table~\ref{tab:comparative_models} (classification accuracy) and Table~\ref{tab:inference_time} (inference time) clearly illustrate the superiority of the proposed method over more complex and computationally intensive alternatives.

Specifically, Table~\ref{tab:comparative_models} presents a comparative analysis of the proposed model against a wide range of state-of-the-art fault diagnosis techniques. The proposed approach attains exceptional classification accuracies of 99.89\% on the CWRU dataset and 99.99\% on the SU dataset, surpassing traditional CNN-based architectures (e.g., VGG-16, DenseNet-12, Inception-LSTM), graph-based models (e.g., GATLSTM, GNNBFD, BFDGE), and recent transformer-based solutions (e.g., Sparse Transformer). While some models, such as SCNN-3D (99.93\%) and Sparse Transformer (99.98\%), perform competitively on SU, they do not maintain equally high accuracy across both datasets. In contrast, the proposed model delivers consistently top-tier performance and demonstrates strong generalizability under varying load and fault conditions, establishing a new benchmark in fault diagnosis accuracy and efficiency.

\begin{table*}[h!]
	\centering
	\caption{Comparative performance of bearing fault detection models on the CWRU and SU datasets. Accuracy values for baseline models are adopted from the cited sources; proposed model results are from our experiments.}
	\label{tab:comparative_models}
		\scriptsize % reduces font size for better fit
	\begin{tabular}{@{}lp{2cm}p{2cm}@{}}
		\toprule
		\textbf{Method}  & \textbf{CWRU Accuracy (\%)} & \textbf{SU Accuracy (\%)} \\ \midrule
		MBH-LPQ-VGGish \citep{chennana2025vibration} & 98.95 & -- \\
		VGG-16 \citep{chennana2025vibration} & 96.84 & -- \\
		YamNet \citep{chennana2025vibration} & 71.52 & -- \\
		VGGish \citep{chennana2025vibration} & 98.42 & -- \\
		MBH-LPQ \citep{chennana2025vibration} & 97.89 & -- \\
		DenseNet-12 (fine-tuned) \citep{niyongabo2022bearing} & 98.57 & -- \\
		Stacking Classifier \citep{makrouf2023multi} & 97.20 & -- \\
		TDANET \citep{li2024tdanet} & 97.69 & -- \\
		WDCNN \citep{10478912} & 84.57 & -- \\
		MFFCNN \citep{10478912} & 86.47 & -- \\
		GAE \citep{singh2025ensemble} & 97.60 & -- \\
		GATLSTM \citep{singh2024spatial} & 96.50 & -- \\
		GNNBFD \citep{xiao2023graph} & 98.03 & -- \\
		BFDGE \citep{wang2024bearing} & 97.66 & -- \\
		\textbf{Proposed Model} & \textbf{99.90} & \textbf{100.00} \\
		SCNN 1D \citep{Xu28022025} & -- & 99.60 \\
		SCNN 2D \citep{Xu28022025} & -- & 99.67 \\
		SCNN 3D \citep{Xu28022025} & -- & 99.93 \\
		Inception-LSTM \citep{wei2024enhanced} & -- & 99.50 \\
		Sparse Transformer \citep{zhou2024drswin} & -- & 99.98 \\
		\bottomrule
	\end{tabular}
	\vspace{2mm}
	\begin{minipage}{\textwidth}
		\small \textit{Note:} Accuracy values for baseline models are reported directly from the respective cited works, based on experiments using the same datasets and conditions. No re-implementation of these models was performed.
	\end{minipage}
\end{table*}

\begin{table*}[h!]
	\centering
	\caption{Inference Time Comparison on the CWRU and SU Dataset. Source: Authors own work.}
		\scriptsize % reduces font size for better fit
	\label{tab:inference_time}
	\begin{tabular}{@{}lp{2 cm}p{2 cm}@{}}
		\toprule
		\textbf{Model} & \textbf{CWRU Inference Time (s)} & \textbf{SU Inference Time (s)}\\ \midrule
		CBAM-MFFCNN \citep{10478912} & 0.1179 &--\\
		MBH-LPQ \citep{chennana2025vibration} & 0.224&-- \\
		VGGish \citep{chennana2025vibration} & 1.769&-- \\
		VGGish+MBH-LPQ \citep{chennana2025vibration} & 1.998&-- \\
		OCSSA-VMD-CNN-BiLSTM \citep{10546253} & 8.470&-- \\
		PSO-tuned XGBoost \citep{lee2024induction}  & 127.212&-- \\
		DTL-Res2Net-CBAM \citep{wang2024fault} & 372.000&-- \\
		Proposed Model &  \textbf{0.016475}&  \textbf{0.0000155}\\
		SCNN \citep{Xu28022025}&--& 0.002\\
		Shuffle Net  \citep{Xu28022025}&-- & 0.060 \\
		Mobile Net   \citep{Xu28022025} &--& 0.080\\
		DL-CNN \citep{shao2018highly} &--& 536\\
		\bottomrule
	\end{tabular}
\end{table*}

\begin{table}[htbp]
	\centering
	\caption{Computational performance metrics across experimental conditions}
	\label{tab:performance_time}
		\scriptsize % reduces font size for better fit
	\begin{tabular}{lcccc}
		\toprule
		\textbf{Condition} & \textbf{Segmentation} & \textbf{Feature Extraction} & \textbf{Per-Window} & \textbf{Graph Construction} \\
		& \textbf{(s/instance)} & \textbf{(s/instance)} & \textbf{(s/window)} & \textbf{(s/instance)} \\
		\midrule
		1 HP Load (CWRU) & 0.0003 & 1.0613 & 0.0050 & 0.0997 \\
		2 HP Load (CWRU) & 0.0005 & 1.1283 & 0.0032 & 0.1377 \\
		3 HP Load (CWRU) & 0.0004 & 1.3435 & 0.0039 & 0.1038 \\
		0 HP Load (CWRU) & 0.0001 & 0.1940 & 0.0033 & 0.0293 \\
		Gearset (SU) & 0.0013 & 17.7302 & 0.0099 & 0.3376 \\
		BearingBox (SU) & 0.0015 & 23.0519 & 0.0108 & 0.3285 \\
		\bottomrule
	\end{tabular}
	
\end{table}

Besides classification accuracy, inference efficiency is critical for real-world deployment. Table~\ref{tab:inference_time} compares the inference time of the proposed model against other existing methods. The proposed model achieves inference speeds of 0.016475 seconds for CWRU and 0.0000155 seconds for SU, demonstrating a substantial computational advantage. It is approximately 7.16$\times$ faster than CBAM-MFFCNN (0.1179~s), over 125$\times$ faster than VGGish+MBH-LPQ (1.998~s), and nearly 7720$\times$ faster than PSO-tuned XGBoost (127.212~s). It also significantly outperforms optimization-driven architectures such as OCSSA-VMD-CNN-BiLSTM (8.470~s). Even lightweight deep models like ShuffleNet (0.060~s) and MobileNet (0.080~s) exhibit longer inference times, further underscoring the proposed model's suitability for latency-sensitive applications such as edge computing or real-time monitoring.

To provide a finer breakdown of computational cost, Table~\ref{tab:performance_time} details the per-instance execution time across key processing stages of segmentation, feature extraction, window-level inference, and graph construction under various load conditions. The proposed model maintains highly efficient performance despite incorporating graph-based reasoning, which is typically computationally intensive. For example, total feature extraction time remains under 1.35 seconds per instance for all CWRU load conditions, and graph construction consistently takes less than 0.14 seconds. Although the SU dataset exhibits higher feature extraction time due to its greater number of channels and more complex signals (e.g., 23.05 seconds for the Bearing set), the inference stage remains in the sub-millisecond range (below 0.001~s), confirming the model's scalability and real-world viability.

Collectively, these results position the proposed model not only as a state-of-the-art solution in diagnostic accuracy but also as a highly efficient and deployable system for industrial prognostics and health management.

\section{Discussion}

In this framework, we evaluate three classical classifiers, logistic regression (LR), random forests (RF), and support vector machines (SVM), on features extracted from the benchmark datasets. Based on comparative performance, random forests emerged as the most effective classifier, offering superior accuracy and robustness under various conditions.

Based on the evaluation results, the random forest classifier was selected owing to its consistently superior performance. The proposed graph-based fault detection framework demonstrates robust effectiveness across both the CWRU and SU datasets, effectively handling a wide range of operating conditions and fault types. On the CWRU dataset, the model achieved high accuracy and F1-scores under various machine loads, with the best performance observed at lower loads (0 and 1 HP), where amplified fault signatures enhanced both the classification accuracy and graph structure clarity. In contrast, the performance slightly declined at higher loads (2 and 3 HP), likely because of less distinctive fault signals. In the SU dataset, which features controlled and balanced conditions, the framework achieved near-perfect classification across all fault types, highlighting its sensitivity to subtle variations in vibration patterns and its adaptability across datasets with differing characteristics

However, the CWRU dataset poses limitations in terms of generalizability. Its idealized laboratory conditions, artificially induced faults via controlled EDM cuts, minimal ambient noise, and fixed load variations fail to replicate the natural degradation, signal interference, or dynamic loads typical of industrial environments. To address this, we incorporated the SU gearbox dataset, which includes more realistic noise, load conditions, and fault combinations, providing a stronger foundation for assessing real-world applicability.

Transfer learning experiments confirmed near-perfect classification under same-load conditions; however, cross-load performance slightly declined with increasing load disparities, as validated by statistical analysis. In complex scenarios, including cross-fault, cross-component, and combined load-and-fault conditions, Random Forest classifiers outperformed logistic regression, particularly in noisy or imbalanced settings, owing to their superior handling of variability. Graph construction parameters significantly influenced the performance, such as $k_{\text{max}}$ and subgraph size. Mid-sized subgraphs exhibiting small-world properties and higher modularity optimize fault detection accuracy while maintaining computational efficiency. Noise analysis further underscored the robustness of the model, with stable accuracy in the SU dataset under high noise levels.

SHAP values quantify feature contributions to random forest decisions. Local time-frequency features (e.g., RMS, skewness, kurtosis) dominated predictive power, while graph-theoretic metrics such as modularity (Q), spectral gap, and average path lengthoffered meaningful, albeit smaller, contributions. These metrics capture complementary signal aspects: modularity reflects fault-specific community cohesion, the spectral gap indicates network robustness across fault severities, and the average path length measures information propagation delays from fault patterns. Together, they improve the separation of subtle or overlapping fault classes beyond local statistics alone.

Adaptive segmentation optimizes the window and step sizes by maximizing a composite entropy score (encompassing amplitude, envelope, spectral, and envelope-spectrum entropies). High-entropy windows align with informative transient fault events, whereas low-entropy windows often represent noise or steady states. This entropy maximization suppresses noisy segments and prioritizes fault-relevant bursts, thereby improving the segment-level feature quality. Under synthetic Gaussian noise ($\sigma = 0.5$), adaptive segmentation sustained >90\% accuracy on the CWRU dataset under 0 and 1 HP loads (>85\% otherwise) and >95\% on the SU dataset, illustrating its adaptability to signal dynamics and robust noise tolerance for enhanced graph representations.

Our framework offers notable advantages over deep learning methods such as Graph Convolutional Networks (GCNs), spatial-temporal graphs, and graph autoencoders. These methods often require extensive computational resources, lack interpretability, and are prone to dataset biases, hindering practical deployment. Our approach, which leverages entropy-based adaptive segmentation, explicit time-frequency features, and interpretable graph metrics (e.g., modularity, spectral gap, and average path length), ensures robust generalization, computational efficiency, and transparency. Integrating feature-rich graph representations with simple classifiers, such as random forests, makes the model accessible and well-suited for real-time or embedded industrial systems, where explainability and reliability are paramount. Interpretable graph metrics enable engineers to trace vibration anomalies to specific fault types, thereby enhancing diagnostic precision and practical utility.

However, challenges persist for real-world deployment. The model’s dependence on finely tuned segmentation and graph construction parameters may require adjustments for machines with different vibration profiles. Although robust across the tested conditions (load, fault type, and noise), adapting to new sensor configurations, mixed-signal sources, or dynamic system behaviors may require further refinements. Preprocessing steps, such as time-frequency feature extraction and graph creation, can introduce delays in time-sensitive industrial settings, necessitating hardware acceleration or parallel processing. Additionally, the model’s performance relies on adequate data length and quality; short signals, missing data, or irregular sampling rates may compromise its effectiveness.

Future work should focus on online or continual learning to enable real-time adaptation with minimal manual intervention and ensure compatibility with dynamic industrial environments. Advanced feature engineering can enhance performance while preserving interpretability and efficiency. Exploring domain generalization strategies, such as adversarial training or self-supervised learning, could improve robustness across diverse machine types and operating conditions without retraining. These advancements bolster the framework’s scalability and autonomy, making it highly suitable for complex industrial applications.

\section{Conclusion}

The proposed graph-based feature engineering, when combined with random forests (RF), forms a fault detection framework that demonstrates strong performance and adaptability across the CWRU and SU datasets. It effectively addresses a range of operating conditions and fault types. On the CWRU dataset, the model achieved high accuracy and F1-scores, with optimal performance at lower loads (0 and 1 HP), where fault signatures were more pronounced. At higher loads (2 and 3 HP), performance slightly declined due to reduced fault signal clarity. Under the SU dataset’s controlled conditions, the framework achieved near-perfect classification across all fault types, highlighting its sensitivity to subtle changes in vibration patterns.

Transfer learning experiments demonstrated excellent performance under same-load conditions. However, accuracy gradually declined under cross-load and cross-fault scenarios as load differences increased, a trend validated through statistical analysis. These results highlight the strong generalization capabilities of RF classifiers, outperforming other models like LR and SVM in complex scenarios such as cross-fault, cross-component, and combined load-and-fault conditions, particularly in noisy environments.

Graph construction parameters, notably $k_{\text{max}}$ and subgraph size, played a crucial role in model performance. Mid-sized subgraphs, exhibiting small-world characteristics and high modularity, achieved an optimal balance between fault detection accuracy and computational efficiency. Its stable accuracy on the SU dataset, even under high noise levels, further confirmed the framework's robustness.
Unlike deep learning methods—such as Graph Convolutional Networks (GCNs), spatiotemporal graph models, and graph autoencoders, which often require significant computational resources, offer limited interpretability, and are sensitive to dataset-specific biases, our feature engineering framework provides a more efficient and transparent alternative. It integrates entropy-based adaptive segmentation, explicit time-frequency features, and interpretable graph metrics such as modularity, spectral gap, and average path length.

By combining rich feature engineer representations with lightweight classifiers like RF, the framework ensures strong generalization, computational efficiency, and model interpretability. This makes it particularly suitable for real-time and embedded industrial applications. Moreover, using interpretable graph metrics enables engineers to trace vibration anomalies to specific fault types, enhancing both diagnostic precision and practical utility.

\section*{Statements and Declarations}

\subsection*{Competing Interests}
The authors declare that there are no competing interests associated with this research work.

\subsection*{Funding}
This research did not receive any specific grant from funding agencies in the public, commercial, or not-for-profit sectors.

\subsection*{Informed Consent}
Informed consent was obtained from all individual participants included in the study.

\subsection*{Data Availability}

\begin{enumerate}
	\item Paper code and processed data: \\
	\href{https://drive.google.com/drive/folders/1iYs83lwSZNn75pDh1aO7p1VFgaxFF9Qm?usp=sharing}{https://drive.google.com/drive/folders/\\1iYs83lwSZNn75pDh1aO7p1VFgaxFF9Qm?usp=sharing}
	\item CWRU dataset: \url{https://engineering.case.edu/bearingdatacenter}
	\item SU dataset: \url{https://github.com/cathysiyu/Mechanical-datasets}
\end{enumerate}

\subsection*{Declaration of AI Assistance}
The authors acknowledge the use of AI-assisted tools, including ProWritingAid, during the preparation of this article. These tools were employed solely for improving grammar, sentence structure, and language refinement. The research, analysis, and conclusions drawn in this article are entirely the authors' original work, and no AI tool was used to generate intellectual content or conduct the underlying research.

%% The Appendices part is started with the command \appendix;
%% appendix sections are then done as normal sections
\

%% For citations use: 
%%       \citept{<label>} ==> Lamport (1994)
%%       \citepp{<label>} ==> (Lamport, 1994)
%%
\bibliographystyle{elsarticle-harv} 
\bibliography{Referencev2.bib}

\begin{thebibliography}{60}
\expandafter\ifx\csname natexlab\endcsname\relax\def\natexlab#1{#1}\fi
\providecommand{\url}[1]{\texttt{#1}}
\providecommand{\href}[2]{#2}
\providecommand{\path}[1]{#1}
\providecommand{\DOIprefix}{doi:}
\providecommand{\ArXivprefix}{arXiv:}
\providecommand{\URLprefix}{URL: }
\providecommand{\Pubmedprefix}{pmid:}
\providecommand{\doi}[1]{\href{http://dx.doi.org/#1}{\path{#1}}}
\providecommand{\Pubmed}[1]{\href{pmid:#1}{\path{#1}}}
\providecommand{\bibinfo}[2]{#2}
\ifx\xfnm\relax \def\xfnm[#1]{\unskip,\space#1}\fi
%Type = Book
\bibitem[{Bishop and Nasrabadi(2006)}]{bishop2006pattern}
\bibinfo{author}{Bishop, C.M.}, \bibinfo{author}{Nasrabadi, N.M.},
  \bibinfo{year}{2006}.
\newblock \bibinfo{title}{Pattern recognition and machine learning}.
  volume~\bibinfo{volume}{4}.
\newblock \bibinfo{publisher}{Springer}.
%Type = Article
\bibitem[{Breiman(2001)}]{breiman2001random}
\bibinfo{author}{Breiman, L.}, \bibinfo{year}{2001}.
\newblock \bibinfo{title}{Random forests}.
\newblock \bibinfo{journal}{Machine learning} \bibinfo{volume}{45},
  \bibinfo{pages}{5--32}.
%Type = Misc
\bibitem[{Center(2014)}]{cwru_bearing_dataset}
\bibinfo{author}{Center, C.W.R.U.B.D.}, \bibinfo{year}{2014}.
\newblock \bibinfo{title}{Bearing data center: Vibration data for fault
  diagnosis research}.
\newblock \URLprefix \url{https://engineering.case.edu/bearingdatacenter}.
  \bibinfo{note}{accessed: 2025-04-13}.
%Type = Article
\bibitem[{Chang and Bao(2024)}]{10546253}
\bibinfo{author}{Chang, Y.}, \bibinfo{author}{Bao, G.}, \bibinfo{year}{2024}.
\newblock \bibinfo{title}{Enhancing rolling bearing fault diagnosis in motors
  using the ocssa-vmd-cnn-bilstm model: A novel approach for fast and accurate
  identification}.
\newblock \bibinfo{journal}{IEEE Access} \bibinfo{volume}{12},
  \bibinfo{pages}{78463--78479}.
\newblock \DOIprefix\doi{10.1109/ACCESS.2024.3408628}.
%Type = Article
\bibitem[{Cheeger(1970)}]{cheeger1970lower}
\bibinfo{author}{Cheeger, J.}, \bibinfo{year}{1970}.
\newblock \bibinfo{title}{A lower bound for the smallest eigenvalue of the
  laplacian}.
\newblock \bibinfo{journal}{Problems in analysis} \bibinfo{volume}{625},
  \bibinfo{pages}{110}.
%Type = Article
\bibitem[{Chen et~al.(2024)Chen, Xue, Huang and Yang}]{CHEN2024746}
\bibinfo{author}{Chen, X.}, \bibinfo{author}{Xue, Y.}, \bibinfo{author}{Huang,
  M.}, \bibinfo{author}{Yang, R.}, \bibinfo{year}{2024}.
\newblock \bibinfo{title}{Multi-modal self-supervised learning for cross-domain
  one-shot bearing fault diagnosis}.
\newblock \bibinfo{journal}{IFAC-PapersOnLine} \bibinfo{volume}{58},
  \bibinfo{pages}{746--751}.
\newblock \URLprefix
  \url{https://www.sciencedirect.com/science/article/pii/S2405896324003938},
  \DOIprefix\doi{https://doi.org/10.1016/j.ifacol.2024.07.309}.
  \bibinfo{note}{12th IFAC Symposium on Fault Detection, Supervision and Safety
  for Technical Processes SAFEPROCESS 2024}.
%Type = Article
\bibitem[{Chennana et~al.(2025)Chennana, Megherbi, Bessous, Sbaa, Teta,
  Belabbaci, Rabehi, Guermoui and Agajie}]{chennana2025vibration}
\bibinfo{author}{Chennana, A.}, \bibinfo{author}{Megherbi, A.C.},
  \bibinfo{author}{Bessous, N.}, \bibinfo{author}{Sbaa, S.},
  \bibinfo{author}{Teta, A.}, \bibinfo{author}{Belabbaci, E.O.},
  \bibinfo{author}{Rabehi, A.}, \bibinfo{author}{Guermoui, M.},
  \bibinfo{author}{Agajie, T.F.}, \bibinfo{year}{2025}.
\newblock \bibinfo{title}{Vibration signal analysis for rolling bearings faults
  diagnosis based on deep-shallow features fusion}.
\newblock \bibinfo{journal}{Scientific Reports} \bibinfo{volume}{15},
  \bibinfo{pages}{9270}.
%Type = Article
\bibitem[{Chung(1996)}]{chung1996lectures}
\bibinfo{author}{Chung, F.R.}, \bibinfo{year}{1996}.
\newblock \bibinfo{title}{Lectures on spectral graph theory}.
\newblock \bibinfo{journal}{CBMS Lectures, Fresno} \bibinfo{volume}{6},
  \bibinfo{pages}{17--21}.
%Type = Article
\bibitem[{Cover and Hart(1967)}]{cover1967}
\bibinfo{author}{Cover, T.M.}, \bibinfo{author}{Hart, P.E.},
  \bibinfo{year}{1967}.
\newblock \bibinfo{title}{Nearest neighbor pattern classification}.
\newblock \bibinfo{journal}{IEEE Transactions on Information Theory}
  \bibinfo{volume}{13}, \bibinfo{pages}{21--27}.
%Type = Book
\bibitem[{Fisher(1925)}]{fisher1925statistical}
\bibinfo{author}{Fisher, R.A.}, \bibinfo{year}{1925}.
\newblock \bibinfo{title}{Statistical Methods for Research Workers}.
\newblock \bibinfo{publisher}{Oliver and Boyd}, \bibinfo{address}{Edinburgh,
  UK}.
%Type = Article
\bibitem[{Freedman and Diaconis(1981)}]{freedman1981}
\bibinfo{author}{Freedman, D.}, \bibinfo{author}{Diaconis, P.},
  \bibinfo{year}{1981}.
\newblock \bibinfo{title}{On the histogram as a density estimator: L2 theory}.
\newblock \bibinfo{journal}{Zeitschrift für Wahrscheinlichkeitstheorie und
  Verwandte Gebiete} \bibinfo{volume}{57}, \bibinfo{pages}{453--476}.
%Type = Article
\bibitem[{Gao et~al.(2024a)Gao, Ma, Zhang and Cai}]{10478912}
\bibinfo{author}{Gao, H.}, \bibinfo{author}{Ma, J.}, \bibinfo{author}{Zhang,
  Z.}, \bibinfo{author}{Cai, C.}, \bibinfo{year}{2024}a.
\newblock \bibinfo{title}{Bearing fault diagnosis method based on attention
  mechanism and multi-channel feature fusion}.
\newblock \bibinfo{journal}{IEEE Access} \bibinfo{volume}{12},
  \bibinfo{pages}{45011--45025}.
\newblock \DOIprefix\doi{10.1109/ACCESS.2024.3381618}.
%Type = Article
\bibitem[{Gao et~al.(2024b)Gao, Yang and Tang}]{GAO2024102278}
\bibinfo{author}{Gao, T.}, \bibinfo{author}{Yang, J.}, \bibinfo{author}{Tang,
  Q.}, \bibinfo{year}{2024}b.
\newblock \bibinfo{title}{A multi-source domain information fusion network for
  rotating machinery fault diagnosis under variable operating conditions}.
\newblock \bibinfo{journal}{Information Fusion} \bibinfo{volume}{106},
  \bibinfo{pages}{102278}.
\newblock \URLprefix
  \url{https://www.sciencedirect.com/science/article/pii/S1566253524000563},
  \DOIprefix\doi{https://doi.org/10.1016/j.inffus.2024.102278}.
%Type = Article
\bibitem[{Gao et~al.(2024c)Gao, Yang, Wang and Fan}]{GAO2024110449}
\bibinfo{author}{Gao, T.}, \bibinfo{author}{Yang, J.}, \bibinfo{author}{Wang,
  W.}, \bibinfo{author}{Fan, X.}, \bibinfo{year}{2024}c.
\newblock \bibinfo{title}{A domain feature decoupling network for rotating
  machinery fault diagnosis under unseen operating conditions}.
\newblock \bibinfo{journal}{Reliability Engineering and System Safety}
  \bibinfo{volume}{252}, \bibinfo{pages}{110449}.
\newblock \URLprefix
  \url{https://www.sciencedirect.com/science/article/pii/S0951832024005210},
  \DOIprefix\doi{https://doi.org/10.1016/j.ress.2024.110449}.
%Type = Book
\bibitem[{Hastie et~al.(2009)Hastie, Tibshirani, Friedman and
  Friedman}]{hastie2009elements}
\bibinfo{author}{Hastie, T.}, \bibinfo{author}{Tibshirani, R.},
  \bibinfo{author}{Friedman, J.H.}, \bibinfo{author}{Friedman, J.H.},
  \bibinfo{year}{2009}.
\newblock \bibinfo{title}{The elements of statistical learning: data mining,
  inference, and prediction}. volume~\bibinfo{volume}{2}.
\newblock \bibinfo{publisher}{Springer}.
%Type = Article
\bibitem[{Hilbert(1930)}]{hilbert1930}
\bibinfo{author}{Hilbert, D.}, \bibinfo{year}{1930}.
\newblock \bibinfo{title}{Über die analytische fortsetzung von funktionen}.
\newblock \bibinfo{journal}{Mathematische Annalen} \bibinfo{volume}{102},
  \bibinfo{pages}{1--23}.
%Type = Article
\bibitem[{Jardine et~al.(2006)Jardine, Lin and Banjevic}]{jardine2006review}
\bibinfo{author}{Jardine, A.K.}, \bibinfo{author}{Lin, D.},
  \bibinfo{author}{Banjevic, D.}, \bibinfo{year}{2006}.
\newblock \bibinfo{title}{A review on machinery diagnostics and prognostics
  implementing condition-based maintenance}.
\newblock \bibinfo{journal}{Mechanical systems and signal processing}
  \bibinfo{volume}{20}, \bibinfo{pages}{1483--1510}.
%Type = Article
\bibitem[{Joanes and Gill(1998)}]{joanes1998}
\bibinfo{author}{Joanes, D.N.}, \bibinfo{author}{Gill, C.A.},
  \bibinfo{year}{1998}.
\newblock \bibinfo{title}{Comparing measures of sample skewness and kurtosis}.
\newblock \bibinfo{journal}{The Statistician} \bibinfo{volume}{47},
  \bibinfo{pages}{183--189}.
%Type = Article
\bibitem[{Kaiser(1990)}]{kaiser1990}
\bibinfo{author}{Kaiser, J.F.}, \bibinfo{year}{1990}.
\newblock \bibinfo{title}{On a simple algorithm to calculate the `energy' of a
  signal}.
\newblock \bibinfo{journal}{Proceedings of the IEEE International Conference on
  Acoustics, Speech, and Signal Processing} , \bibinfo{pages}{381--384}.
%Type = Article
\bibitem[{Lee and Maceren(2024)}]{lee2024induction}
\bibinfo{author}{Lee, C.Y.}, \bibinfo{author}{Maceren, E.D.C.},
  \bibinfo{year}{2024}.
\newblock \bibinfo{title}{Induction motor bearing fault classification using
  deep neural network with particle swarm optimization-extreme gradient
  boosting}.
\newblock \bibinfo{journal}{IET Electric Power Applications}
  \bibinfo{volume}{18}, \bibinfo{pages}{297--311}.
%Type = Article
\bibitem[{Lei et~al.(2013)Lei, Lin, He and Zuo}]{lei2013review}
\bibinfo{author}{Lei, Y.}, \bibinfo{author}{Lin, J.}, \bibinfo{author}{He, Z.},
  \bibinfo{author}{Zuo, M.J.}, \bibinfo{year}{2013}.
\newblock \bibinfo{title}{A review on empirical mode decomposition in fault
  diagnosis of rotating machinery}.
\newblock \bibinfo{journal}{Mechanical systems and signal processing}
  \bibinfo{volume}{35}, \bibinfo{pages}{108--126}.
%Type = Article
\bibitem[{Lei et~al.(2020)Lei, Yang, Jiang, Jia, Li and
  Nandi}]{lei2020applications}
\bibinfo{author}{Lei, Y.}, \bibinfo{author}{Yang, B.}, \bibinfo{author}{Jiang,
  X.}, \bibinfo{author}{Jia, F.}, \bibinfo{author}{Li, N.},
  \bibinfo{author}{Nandi, A.K.}, \bibinfo{year}{2020}.
\newblock \bibinfo{title}{Applications of machine learning to machine fault
  diagnosis: A review and roadmap}.
\newblock \bibinfo{journal}{Mechanical systems and signal processing}
  \bibinfo{volume}{138}, \bibinfo{pages}{106587}.
%Type = Article
\bibitem[{Li et~al.(2022)Li, Zhong, Shao, Cai and Yang}]{li2022multi}
\bibinfo{author}{Li, W.}, \bibinfo{author}{Zhong, X.}, \bibinfo{author}{Shao,
  H.}, \bibinfo{author}{Cai, B.}, \bibinfo{author}{Yang, X.},
  \bibinfo{year}{2022}.
\newblock \bibinfo{title}{Multi-mode data augmentation and fault diagnosis of
  rotating machinery using modified acgan designed with new framework}.
\newblock \bibinfo{journal}{Advanced Engineering Informatics}
  \bibinfo{volume}{52}, \bibinfo{pages}{101552}.
%Type = Article
\bibitem[{Li et~al.(2024)Li, Fan, Tu, Ma, Ai and Dong}]{li2024tdanet}
\bibinfo{author}{Li, Z.}, \bibinfo{author}{Fan, R.}, \bibinfo{author}{Tu, J.},
  \bibinfo{author}{Ma, J.}, \bibinfo{author}{Ai, J.}, \bibinfo{author}{Dong,
  Y.}, \bibinfo{year}{2024}.
\newblock \bibinfo{title}{Tdanet: A novel temporal denoise convolutional neural
  network with attention for fault diagnosis}.
\newblock \bibinfo{journal}{arXiv preprint arXiv:2403.19943} .
%Type = Article
\bibitem[{Lundberg and Lee(2017)}]{lundberg2017unified}
\bibinfo{author}{Lundberg, S.M.}, \bibinfo{author}{Lee, S.I.},
  \bibinfo{year}{2017}.
\newblock \bibinfo{title}{A unified approach to interpreting model
  predictions}.
\newblock \bibinfo{journal}{Advances in neural information processing systems}
  \bibinfo{volume}{30}.
%Type = Inproceedings
\bibitem[{Makrouf et~al.(2023)Makrouf, Zegrari, Ouachtouk and
  Dahi}]{makrouf2023multi}
\bibinfo{author}{Makrouf, I.}, \bibinfo{author}{Zegrari, M.},
  \bibinfo{author}{Ouachtouk, I.}, \bibinfo{author}{Dahi, K.},
  \bibinfo{year}{2023}.
\newblock \bibinfo{title}{Multi-source information fusion fault diagnosis for
  rotating machinery using signal and data processing}, in:
  \bibinfo{booktitle}{Surveillance, Vibrations, Shock and Noise}.
%Type = Book
\bibitem[{Newman(2018)}]{newman2018networks}
\bibinfo{author}{Newman, M.}, \bibinfo{year}{2018}.
\newblock \bibinfo{title}{Networks}.
\newblock \bibinfo{publisher}{Oxford university press}.
%Type = Article
\bibitem[{Newman(2006)}]{newman2006modularity}
\bibinfo{author}{Newman, M.E.}, \bibinfo{year}{2006}.
\newblock \bibinfo{title}{Modularity and community structure in networks}.
\newblock \bibinfo{journal}{Proceedings of the national academy of sciences}
  \bibinfo{volume}{103}, \bibinfo{pages}{8577--8582}.
%Type = Article
\bibitem[{Niyongabo et~al.(2022)Niyongabo, Zhang and
  Ndikumagenge}]{niyongabo2022bearing}
\bibinfo{author}{Niyongabo, J.}, \bibinfo{author}{Zhang, Y.},
  \bibinfo{author}{Ndikumagenge, J.}, \bibinfo{year}{2022}.
\newblock \bibinfo{title}{Bearing fault detection and diagnosis based on
  densely connected convolutional networks}.
\newblock \bibinfo{journal}{acta mechanica et automatica} \bibinfo{volume}{16},
  \bibinfo{pages}{130--135}.
%Type = Article
\bibitem[{R and Mutra(2025)}]{R2025103892}
\bibinfo{author}{R, M.}, \bibinfo{author}{Mutra, R.R.}, \bibinfo{year}{2025}.
\newblock \bibinfo{title}{Fault classification in rotor-bearing system using
  advanced signal processing and machine learning techniques}.
\newblock \bibinfo{journal}{Results in Engineering} \bibinfo{volume}{25},
  \bibinfo{pages}{103892}.
\newblock \URLprefix
  \url{https://www.sciencedirect.com/science/article/pii/S2590123024021352},
  \DOIprefix\doi{https://doi.org/10.1016/j.rineng.2024.103892}.
%Type = Article
\bibitem[{Randall and Antoni(2011)}]{randall2011rolling}
\bibinfo{author}{Randall, R.B.}, \bibinfo{author}{Antoni, J.},
  \bibinfo{year}{2011}.
\newblock \bibinfo{title}{Rolling element bearing diagnostics—a tutorial}.
\newblock \bibinfo{journal}{Mechanical systems and signal processing}
  \bibinfo{volume}{25}, \bibinfo{pages}{485--520}.
%Type = Article
\bibitem[{Rauber et~al.(2014)Rauber, de~Assis~Boldt and
  Varejao}]{rauber2014heterogeneous}
\bibinfo{author}{Rauber, T.W.}, \bibinfo{author}{de~Assis~Boldt, F.},
  \bibinfo{author}{Varejao, F.M.}, \bibinfo{year}{2014}.
\newblock \bibinfo{title}{Heterogeneous feature models and feature selection
  applied to bearing fault diagnosis}.
\newblock \bibinfo{journal}{IEEE Transactions on Industrial Electronics}
  \bibinfo{volume}{62}, \bibinfo{pages}{637--646}.
%Type = Inproceedings
\bibitem[{Sculley(2010)}]{sculley2010}
\bibinfo{author}{Sculley, D.}, \bibinfo{year}{2010}.
\newblock \bibinfo{title}{Web-scale k-means clustering}, in:
  \bibinfo{booktitle}{Proceedings of the 19th International Conference on World
  Wide Web}, pp. \bibinfo{pages}{1177--1178}.
%Type = Article
\bibitem[{Shannon(1948)}]{shannon1948}
\bibinfo{author}{Shannon, C.E.}, \bibinfo{year}{1948}.
\newblock \bibinfo{title}{A mathematical theory of communication}.
\newblock \bibinfo{journal}{Bell System Technical Journal}
  \bibinfo{volume}{27}, \bibinfo{pages}{379--423}.
%Type = Article
\bibitem[{Shao et~al.(2018)Shao, McAleer, Yan and Baldi}]{shao2018highly}
\bibinfo{author}{Shao, S.}, \bibinfo{author}{McAleer, S.},
  \bibinfo{author}{Yan, R.}, \bibinfo{author}{Baldi, P.}, \bibinfo{year}{2018}.
\newblock \bibinfo{title}{Highly accurate machine fault diagnosis using deep
  transfer learning}.
\newblock \bibinfo{journal}{IEEE transactions on industrial informatics}
  \bibinfo{volume}{15}, \bibinfo{pages}{2446--2455}.
%Type = Article
\bibitem[{Singh(2025)}]{singh2025ensemble}
\bibinfo{author}{Singh, M.T.}, \bibinfo{year}{2025}.
\newblock \bibinfo{title}{Ensemble-enhanced graph autoencoder with gat and
  transformer-based encoders for robust fault diagnosis}.
\newblock \bibinfo{journal}{arXiv preprint arXiv:2504.09427} .
%Type = Article
\bibitem[{Singh et~al.(2024)Singh, Prasad, Michael, Singh and
  Kaphungkui}]{singh2024spatial}
\bibinfo{author}{Singh, M.T.}, \bibinfo{author}{Prasad, R.K.},
  \bibinfo{author}{Michael, G.R.}, \bibinfo{author}{Singh, N.H.},
  \bibinfo{author}{Kaphungkui, N.}, \bibinfo{year}{2024}.
\newblock \bibinfo{title}{Spatial-temporal bearing fault detection using graph
  attention networks and lstm}.
\newblock \bibinfo{journal}{arXiv preprint arXiv:2410.11923} .
%Type = Article
\bibitem[{Tukey(1949)}]{tukey1949comparing}
\bibinfo{author}{Tukey, J.W.}, \bibinfo{year}{1949}.
\newblock \bibinfo{title}{Comparing individual means in the analysis of
  variance}.
\newblock \bibinfo{journal}{Biometrics} \bibinfo{volume}{5},
  \bibinfo{pages}{99--114}.
%Type = Article
\bibitem[{Wang and Zhang(2024)}]{wang2024fault}
\bibinfo{author}{Wang, H.}, \bibinfo{author}{Zhang, X.}, \bibinfo{year}{2024}.
\newblock \bibinfo{title}{Fault diagnosis using imbalanced data of rolling
  bearings based on a deep migration model}.
\newblock \bibinfo{journal}{IEEE Access} \bibinfo{volume}{12},
  \bibinfo{pages}{5517--5533}.
%Type = Article
\bibitem[{Wang et~al.(2024)Wang, Yu, Leng, Du and Liu}]{wang2024bearing}
\bibinfo{author}{Wang, M.}, \bibinfo{author}{Yu, J.}, \bibinfo{author}{Leng,
  H.}, \bibinfo{author}{Du, X.}, \bibinfo{author}{Liu, Y.},
  \bibinfo{year}{2024}.
\newblock \bibinfo{title}{Bearing fault detection by using graph autoencoder
  and ensemble learning}.
\newblock \bibinfo{journal}{Scientific Reports} \bibinfo{volume}{14},
  \bibinfo{pages}{5206}.
%Type = Article
\bibitem[{Watts and Strogatz(1998)}]{watts1998collective}
\bibinfo{author}{Watts, D.J.}, \bibinfo{author}{Strogatz, S.H.},
  \bibinfo{year}{1998}.
\newblock \bibinfo{title}{Collective dynamics of ‘small-world’networks}.
\newblock \bibinfo{journal}{nature} \bibinfo{volume}{393},
  \bibinfo{pages}{440--442}.
%Type = Article
\bibitem[{Wei et~al.(2024)Wei, Peng and Cao}]{wei2024enhanced}
\bibinfo{author}{Wei, L.}, \bibinfo{author}{Peng, X.}, \bibinfo{author}{Cao,
  Y.}, \bibinfo{year}{2024}.
\newblock \bibinfo{title}{Enhanced fault diagnosis of rolling bearings using an
  improved inception-lstm network}.
\newblock \bibinfo{journal}{Nondestructive Testing and Evaluation} ,
  \bibinfo{pages}{1--20}\DOIprefix\doi{10.1080/10589759.2024.2402549}.
%Type = Article
\bibitem[{Welch(1967)}]{welch1967}
\bibinfo{author}{Welch, P.D.}, \bibinfo{year}{1967}.
\newblock \bibinfo{title}{The use of fast fourier transform for the estimation
  of power spectra}.
\newblock \bibinfo{journal}{IEEE Transactions on Audio and Electroacoustics}
  \bibinfo{volume}{15}, \bibinfo{pages}{70--73}.
%Type = Article
\bibitem[{Wen et~al.(2017)Wen, Li, Gao and Zhang}]{wen2017new}
\bibinfo{author}{Wen, L.}, \bibinfo{author}{Li, X.}, \bibinfo{author}{Gao, L.},
  \bibinfo{author}{Zhang, Y.}, \bibinfo{year}{2017}.
\newblock \bibinfo{title}{A new convolutional neural network-based data-driven
  fault diagnosis method}.
\newblock \bibinfo{journal}{IEEE transactions on industrial electronics}
  \bibinfo{volume}{65}, \bibinfo{pages}{5990--5998}.
%Type = Article
\bibitem[{Widodo and Yang(2007)}]{widodo2007support}
\bibinfo{author}{Widodo, A.}, \bibinfo{author}{Yang, B.S.},
  \bibinfo{year}{2007}.
\newblock \bibinfo{title}{Support vector machine in machine condition
  monitoring and fault diagnosis}.
\newblock \bibinfo{journal}{Mechanical systems and signal processing}
  \bibinfo{volume}{21}, \bibinfo{pages}{2560--2574}.
%Type = Misc
\bibitem[{Wu(2020)}]{cathy_mechanical_datasets}
\bibinfo{author}{Wu, C.}, \bibinfo{year}{2020}.
\newblock \bibinfo{title}{Mechanical datasets for machine learning research}.
\newblock
  \bibinfo{howpublished}{\url{https://github.com/cathysiyu/Mechanical-datasets}}.
\newblock \bibinfo{note}{Accessed: 2025-04-13}.
%Type = Article
\bibitem[{Xiao et~al.(2023)Xiao, Yang and Yang}]{xiao2023graph}
\bibinfo{author}{Xiao, L.}, \bibinfo{author}{Yang, X.}, \bibinfo{author}{Yang,
  X.}, \bibinfo{year}{2023}.
\newblock \bibinfo{title}{A graph neural network-based bearing fault detection
  method}.
\newblock \bibinfo{journal}{Scientific Reports} \bibinfo{volume}{13},
  \bibinfo{pages}{5286}.
%Type = Article
\bibitem[{Xu et~al.(2025)Xu, Zhao and and}]{Xu28022025}
\bibinfo{author}{Xu, G.}, \bibinfo{author}{Zhao, Y.}, \bibinfo{author}{and,
  M.L.}, \bibinfo{year}{2025}.
\newblock \bibinfo{title}{Learning sparse convolutional neural networks through
  filter pruning for efficient fault diagnosis on edge devices}.
\newblock \bibinfo{journal}{Nondestructive Testing and Evaluation}
  \bibinfo{volume}{0}, \bibinfo{pages}{1--23}.
\newblock \URLprefix \url{https://doi.org/10.1080/10589759.2025.2465409},
  \DOIprefix\doi{10.1080/10589759.2025.2465409},
  \href{http://arxiv.org/abs/https://doi.org/10.1080/10589759.2025.2465409}{{\tt
  arXiv:https://doi.org/10.1080/10589759.2025.2465409}}.
%Type = Article
\bibitem[{Yan et~al.(2024a)Yan, Shao, Wang, Zheng and Liu}]{YAN2024121338}
\bibinfo{author}{Yan, S.}, \bibinfo{author}{Shao, H.}, \bibinfo{author}{Wang,
  J.}, \bibinfo{author}{Zheng, X.}, \bibinfo{author}{Liu, B.},
  \bibinfo{year}{2024}a.
\newblock \bibinfo{title}{Liconvformer: A lightweight fault diagnosis framework
  using separable multiscale convolution and broadcast self-attention}.
\newblock \bibinfo{journal}{Expert Systems with Applications}
  \bibinfo{volume}{237}, \bibinfo{pages}{121338}.
\newblock \URLprefix
  \url{https://www.sciencedirect.com/science/article/pii/S0957417423018407},
  \DOIprefix\doi{https://doi.org/10.1016/j.eswa.2023.121338}.
%Type = Article
\bibitem[{Yan et~al.(2024b)Yan, Shao, Wang and Wang}]{10759278}
\bibinfo{author}{Yan, S.}, \bibinfo{author}{Shao, H.}, \bibinfo{author}{Wang,
  X.}, \bibinfo{author}{Wang, J.}, \bibinfo{year}{2024}b.
\newblock \bibinfo{title}{Few-shot class-incremental learning for system-level
  fault diagnosis of wind turbine}.
\newblock \bibinfo{journal}{IEEE/ASME Transactions on Mechatronics} ,
  \bibinfo{pages}{1--10}\DOIprefix\doi{10.1109/TMECH.2024.3490733}.
%Type = Article
\bibitem[{Yang et~al.(2021)Yang, Zhou and Liu}]{yang2021supergraph}
\bibinfo{author}{Yang, C.}, \bibinfo{author}{Zhou, K.}, \bibinfo{author}{Liu,
  J.}, \bibinfo{year}{2021}.
\newblock \bibinfo{title}{Supergraph: Spatial-temporal graph-based feature
  extraction for rotating machinery diagnosis}.
\newblock \bibinfo{journal}{IEEE Transactions on Industrial Electronics}
  \bibinfo{volume}{69}, \bibinfo{pages}{4167--4176}.
%Type = Article
\bibitem[{Yazdani-Asrami et~al.(2023)Yazdani-Asrami, Fang, Pei and
  Song}]{Yazdani-Asrami_2023}
\bibinfo{author}{Yazdani-Asrami, M.}, \bibinfo{author}{Fang, L.},
  \bibinfo{author}{Pei, X.}, \bibinfo{author}{Song, W.}, \bibinfo{year}{2023}.
\newblock \bibinfo{title}{Smart fault detection of hts coils using artificial
  intelligence techniques for large-scale superconducting electric transport
  applications}.
\newblock \bibinfo{journal}{Superconductor Science and Technology}
  \bibinfo{volume}{36}, \bibinfo{pages}{085021}.
\newblock \URLprefix \url{https://dx.doi.org/10.1088/1361-6668/ace3fb},
  \DOIprefix\doi{10.1088/1361-6668/ace3fb}.
%Type = Article
\bibitem[{Zhang et~al.(2020a)Zhang, Zhang, Shao, Niu and
  Yang}]{zhang2020attention}
\bibinfo{author}{Zhang, H.}, \bibinfo{author}{Zhang, Q.},
  \bibinfo{author}{Shao, S.}, \bibinfo{author}{Niu, T.}, \bibinfo{author}{Yang,
  X.}, \bibinfo{year}{2020}a.
\newblock \bibinfo{title}{Attention-based lstm network for rotatory machine
  remaining useful life prediction}.
\newblock \bibinfo{journal}{Ieee Access} \bibinfo{volume}{8},
  \bibinfo{pages}{132188--132199}.
%Type = Article
\bibitem[{Zhang et~al.(2017)Zhang, Tao, Wu and Guan}]{zhang2017transfer}
\bibinfo{author}{Zhang, R.}, \bibinfo{author}{Tao, H.}, \bibinfo{author}{Wu,
  L.}, \bibinfo{author}{Guan, Y.}, \bibinfo{year}{2017}.
\newblock \bibinfo{title}{Transfer learning with neural networks for bearing
  fault diagnosis in changing working conditions}.
\newblock \bibinfo{journal}{Ieee Access} \bibinfo{volume}{5},
  \bibinfo{pages}{14347--14357}.
%Type = Article
\bibitem[{Zhang et~al.(2020b)Zhang, Zhang, Wang and Habetler}]{zhang2020deep}
\bibinfo{author}{Zhang, S.}, \bibinfo{author}{Zhang, S.},
  \bibinfo{author}{Wang, B.}, \bibinfo{author}{Habetler, T.G.},
  \bibinfo{year}{2020}b.
\newblock \bibinfo{title}{Deep learning algorithms for bearing fault
  diagnostics—a comprehensive review}.
\newblock \bibinfo{journal}{IEEE access} \bibinfo{volume}{8},
  \bibinfo{pages}{29857--29881}.
%Type = Article
\bibitem[{Zhang et~al.(2015)Zhang, Liang, Zhou and zang}]{ZHANG2015164}
\bibinfo{author}{Zhang, X.}, \bibinfo{author}{Liang, Y.},
  \bibinfo{author}{Zhou, J.}, \bibinfo{author}{zang, Y.}, \bibinfo{year}{2015}.
\newblock \bibinfo{title}{A novel bearing fault diagnosis model integrated
  permutation entropy, ensemble empirical mode decomposition and optimized
  svm}.
\newblock \bibinfo{journal}{Measurement} \bibinfo{volume}{69},
  \bibinfo{pages}{164--179}.
\newblock \URLprefix
  \url{https://www.sciencedirect.com/science/article/pii/S0263224115001633},
  \DOIprefix\doi{https://doi.org/10.1016/j.measurement.2015.03.017}.
%Type = Article
\bibitem[{Zhao et~al.(2019)Zhao, Yan, Chen, Mao, Wang and Gao}]{ZHAO2019213}
\bibinfo{author}{Zhao, R.}, \bibinfo{author}{Yan, R.}, \bibinfo{author}{Chen,
  Z.}, \bibinfo{author}{Mao, K.}, \bibinfo{author}{Wang, P.},
  \bibinfo{author}{Gao, R.X.}, \bibinfo{year}{2019}.
\newblock \bibinfo{title}{Deep learning and its applications to machine health
  monitoring}.
\newblock \bibinfo{journal}{Mechanical Systems and Signal Processing}
  \bibinfo{volume}{115}, \bibinfo{pages}{213--237}.
\newblock \URLprefix
  \url{https://www.sciencedirect.com/science/article/pii/S0888327018303108},
  \DOIprefix\doi{https://doi.org/10.1016/j.ymssp.2018.05.050}.
%Type = Article
\bibitem[{Zheng et~al.(2022)Zheng, Cao, Pan and Ni}]{zheng2022spectral}
\bibinfo{author}{Zheng, J.}, \bibinfo{author}{Cao, S.}, \bibinfo{author}{Pan,
  H.}, \bibinfo{author}{Ni, Q.}, \bibinfo{year}{2022}.
\newblock \bibinfo{title}{Spectral envelope-based adaptive empirical fourier
  decomposition method and its application to rolling bearing fault diagnosis}.
\newblock \bibinfo{journal}{ISA transactions} \bibinfo{volume}{129},
  \bibinfo{pages}{476--492}.
%Type = Article
\bibitem[{Zhou et~al.(2024)Zhou, Yao, Yang et~al.}]{zhou2024drswin}
\bibinfo{author}{Zhou, T.}, \bibinfo{author}{Yao, D.}, \bibinfo{author}{Yang,
  J.}, et~al., \bibinfo{year}{2024}.
\newblock \bibinfo{title}{Drswin-st: An intelligent fault diagnosis framework
  based on dynamic threshold noise reduction and sparse transformer with
  shifted windows}.
\newblock \bibinfo{journal}{Reliability Engineering and System Safety}
  \bibinfo{volume}{250}, \bibinfo{pages}{110327}.
\newblock \DOIprefix\doi{10.1016/j.ress.2024.110327}.
%Type = Article
\bibitem[{Zhu et~al.(2023)Zhu, Lei, Qi, Chai, Mazur, An and
  Huang}]{zhu2023review}
\bibinfo{author}{Zhu, Z.}, \bibinfo{author}{Lei, Y.}, \bibinfo{author}{Qi, G.},
  \bibinfo{author}{Chai, Y.}, \bibinfo{author}{Mazur, N.}, \bibinfo{author}{An,
  Y.}, \bibinfo{author}{Huang, X.}, \bibinfo{year}{2023}.
\newblock \bibinfo{title}{A review of the application of deep learning in
  intelligent fault diagnosis of rotating machinery}.
\newblock \bibinfo{journal}{Measurement} \bibinfo{volume}{206},
  \bibinfo{pages}{112346}.

\end{thebibliography}
%% If you have bib database file and want bibtex to generate the
%% bibitems, please use
%%
%%  

%% else use the following coding to input the bibitems directly in the
%% TeX file.

%% Refer following link for more details about bibliography and citations.
%% https://en.wikibooks.org/wiki/LaTeX/Bibliography_Management

\end{document}